\documentclass[10pt,twocolumn,letterpaper]{article}

\usepackage{cvpr}              

\definecolor{cvprblue}{rgb}{0.21,0.49,0.74}
\usepackage[pagebackref,breaklinks,colorlinks,allcolors=cvprblue]{hyperref}

\usepackage[accsupp]{axessibility} 

\usepackage{multirow} 
\usepackage{arydshln} 
\usepackage{makecell}
\usepackage{pgfplots}
\pgfplotsset{compat=1.6}
\usepackage[export]{adjustbox} 

\usepackage{pgfplotstable}

\usepackage{pifont}
\newcommand{\xmark}{\ding{55}} 

\def\paperID{11} 
\def\confName{CVPR}
\def\confYear{2026}

\title{Video Patch Pruning: Efficient Video Instance Segmentation via Early Token Reduction}

\author{Patrick Glandorf, Thomas Norrenbrock, Bodo Rosenhahn\\
Institute for Information Processing (tnt)\\
L3S - Leibniz University Hannover, Germany\\
{\tt\small \{glandorf, norrenbr, rosenhahn\}@tnt.uni-hannover.de}
}

\begin{document}
\maketitle

\begin{abstract}

Vision Transformers (ViTs) have demonstrated state-of-the-art performance in several benchmarks, yet their high computational costs hinders their practical deployment.
Patch Pruning offers significant savings, but existing approaches restrict token reduction to deeper layers, leaving early-stage compression unexplored.
This limits their potential for holistic efficiency.
In this work, we present a novel Video Patch Pruning framework (VPP) that integrates temporal prior knowledge to enable efficient sparsity within early ViT layers. 
Our approach is motivated by the observation that prior features extracted from deeper layers exhibit strong foreground selectivity.
Therefore we propose a fully differentiable module for temporal mapping to accurately select the most relevant patches in early network stages.
Notably, the proposed method enables a patch reduction of up to 60\% in dense prediction tasks, exceeding the capabilities of conventional image-based patch pruning, which typically operate around a 30\% patch sparsity.
VPP excels the high-sparsity regime, sustaining remarkable performance even when patch usage is reduced below 55\%.
Specifically, it preserves stable results with a maximal performance drop of 0.6\% on the Youtube-VIS 2021 dataset.
\textbf{Code is available} \href{https://github.com/PatGlan/Video-Patch-Pruning.git}{\textbf{here}}~\footnote{\href{https://github.com/PatGlan/Video-Patch-Pruning.git}{https://github.com/PatGlan/Video-Patch-Pruning.git}\label{githubLink}}

\end{abstract}

\section{Introduction}

\begin{figure}
    \centering
    \vspace{-0pt}
    \includegraphics[width=1\linewidth]{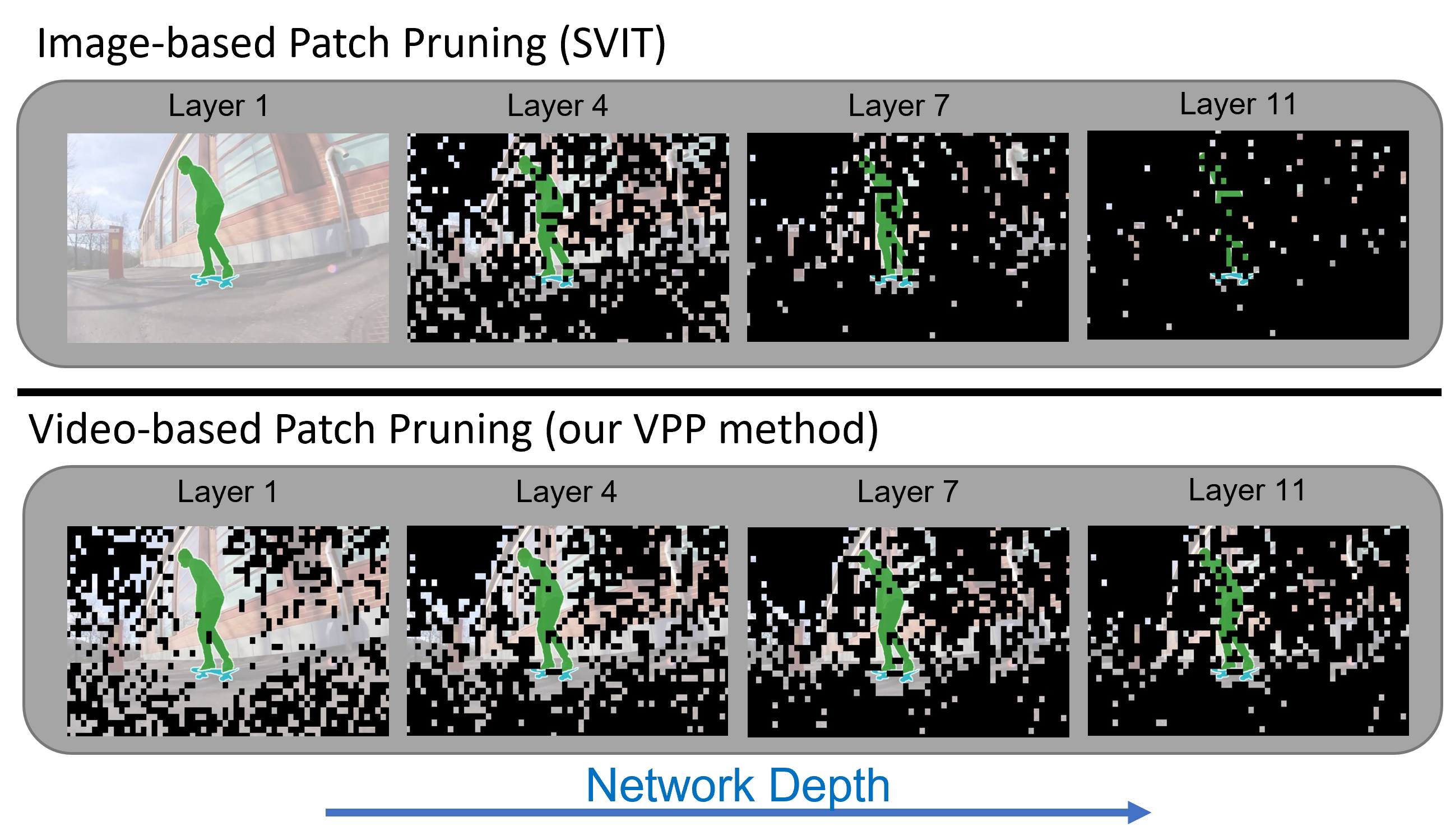}
    \vspace{-15pt}
    \caption{
    Pruning masks for image-based (SViT) and video-based (VPP) pruning at 60\% avg. patch reduction.
    For instance segmentation, only the highlighted objects must be segmented, while the background can be ignored.
    \textit{VPP} uses temporal information for early-stage pruning, effectively preserving the foreground.
    }
    \label{fig:img_vs_video_pp_masks}
    \vspace{-10pt}
\end{figure}

The task of Video Instance Segmentation (VIS) integrates the simultaneous detection, segmentation and tracking of object instances across video sequences.
This multimodal task is fundamental for safety-critical systems, such as autonomous driving and medical imaging~\cite{Siam2021VideoCA, Ma_2024}.
However, the substantial computational overhead of VIS makes it difficult to achieve the low-latency performance required for execution on edge devices.
For segmentation tasks, recent works increasingly favor Vision Transformer (ViTs) architectures, reflecting a broader shift toward attention-based modeling in computer vision~\cite{Yang2019vis, Zhang_2023_ICCV}.
Nonetheless, the self-attention mechanism of ViTs scales quadratically with the number of patches, leading to intensive memory and processing demands that often exceed the capabilities of resource-constrained edge devices~\cite{Fournier2023SurveyTransformers, SAHA2025130417}.
While weight pruning methods improve efficiency by reducing model parameters~\cite{HyperSparse2023, glandorf_2025_ICCVW, rosenh_optimSpConstNN_2023,norrenbrock2024q}, patch pruning aims to enhance efficiency by removing less informative features~\cite{tang2024surveytransformercompression}.
In VIS, the primary training objective is conditioned on instance related foreground information~\cite{ren15fasterrcnn, carion_2020_e2eODT, Cheng_2022_CVPR}.
Therefore the background is less critical to the learning process, as prioritizing foreground features enhances robustness against background variations~\cite{xiao2021noisesignalroleimage}.
Recent patch pruning methods typically operate through image-wise execution, treating each frame as an independent entity for processing~\cite{liang2022evit, Liu_2024_WACV, KIM2024105239}. 
A critical limitation of these methods is that feature sparsity is only enforced in deeper layers, while earlier features are kept fully dense.
According to Tang et al.~\cite{Tang_2022_CVPR}, this is caused by the increasing feature redundancy within deeper layers.
However, this late-stage reduction strategy of image-based patch pruning methods lowers the overall efficiency of the network.

To address this problem, we introduce \textit{Video Patch Pruning} (\textit{VPP}), a novel pruning strategy that overcomes the limitations of late-stage pruning by enabling effective patch reduction within the network's earlier layers.
This pruning strategy utilizes high level features from the previous frame in order to accurately identify task-relevant patches in early network stages.
Thus, \textit{VPP} is able to remove up to 40\% of patches after the first layer, by selectively removing background features while preserving the high-fidelity foreground patches necessary for accurate instance segmentation.
However, to ensure the identification of new object instances in the video context, background features are not fully removed, but instead are sparsely activated via Gumbel-distributed noise~\cite{jang2017gumbelsoftmax}.
Patch pruning methods often rely on the classification token to guide pruning decisions~\cite{fayyaz2022adaptivetokensamplingefficient, xu2021evovitslowfasttokenevolution, zhang2024synergistic}.
Our approach directly optimizes the spatial feature masking, thereby eliminating the dependency on classification tokens for dense prediction tasks.
Furthermore, \textit{VPP} reduces parameter count by over $1.8\times$, compared to existing methods such as SVIT, DynamicViT and TPS, as summarized in Tab.~\ref{tab:categorical_overview}.
By pruning patches dynamically, \textit{VPP} allows to scale the size of the pruning mask for larger object instances, while utilizing sparse features for smaller objects.
This adaptive mask scaling minimizes computational overhead by tailoring resource allocation to spatial demands.
In summary, \textit{VPP} enables early-stage pruning and dynamic mask-scaling, while operating directly on patch features rather than a classification token.
All these properties are essential for efficient \textit{VIS}, where the framework must simultaneously preserve high-fidelity features for detection, segmentation, and temporal tracking.

Furthermore this work analyzes the property of Foreground Selectivity across layers.
We show that early layers fail to correctly identify patches belonging to an object of interest.
Consequently, pruning at this stage results in arbitrarily sampled masks, explaining why image-based patch pruning fails when applied to these layers.
To overcome this problem, \textit{VPP} maps high-level features from previous frames onto the current one, in order to ensure high foreground selectivity in early pruning stages.
Specifically, we introduce a lightweight mapping module, which propagates the spatio-temporal motion of patches across frames.
Guided by these mapped features, less important patches are removed, effectively filtering out redundant background features.

Our proposed \textit{VPP} removes up to 40\% of patches after the very first layer, without sacrificing critical foreground information belonging to object instances.
This allows \textit{VPP} to achieve excellent results, while saving up to 60\% of the computed patches.
In contrast, image-based patch pruning methods usually operate at half the sparsity ($\approx\!30\%$ average reduction) over all layers~\cite{Rao_2021_Neurips, liang2022evit, Liu_2024_WACV}.
We show that \textit{VPP} effectively handles abrupt context shifts by successfully recapturing the entire foreground within two timesteps.

In summary, this work makes the following key contributions:
\begin{itemize}
    \item We introduce \textit{Video Patch Pruning} (\textit{VPP}), an online pruning method for early patch reduction. 
    \textit{VPP} sets a clear state-of-the-art in Patch Pruning for Video Instance Segmentation, achieving an average patch usage of down to 40\% with minimal loss in performance.

    \item We show that image-based patch pruning fails when applied to early-stage layers, by analyzing the property of Foreground Selectivity.
    \textit{VPP} utilizes the video context to propagate late-stage patch information into the next frame using our proposed \textit{Mapping Selective Module}.

    \item Extensive experiments demonstrate that \textit{VPP} shows superior foreground sensitivity throughout all layers,
    retaining at least 8\% more foreground patches compared to prior art.

\end{itemize}

\begin{table}[t]
    \centering
    \resizebox{1.0\linewidth}{!}{
    \begin{tabular}{c|cccc}
    \toprule
         Method & \makecell{pruning before \\ layer 3} & \makecell{pruning- \\stages} & \makecell{additional \\ prune param}& \makecell{dynamic \\ pruning} \\
         \midrule
         DynamicViT~\cite{Rao_2021_Neurips} & \xmark & 3 & 0.723 M & \xmark \\
         TPS~\cite{wei2023tps} & \xmark & 3 & 0.723 M & \xmark\\
         SViT~\cite{Liu_2024_WACV} & \xmark & 9 & 0.341 M & \checkmark \\
         \cdashline{1-5}
         \addlinespace[1pt]
         VPP (ours)  & \checkmark & 4 & 0.189 M & \checkmark\\
         \bottomrule
    \end{tabular}
    }
    \vspace{-0pt}
    \caption{Categorical overview Patch Pruning methods. 
    VPP leverages video context for early feature reduction, ensuring minimal computational overhead. 
    The method also enables adaptive mask sizing to better adapt the mask to varying instance sizes.
    }
    \vspace{-5pt}
    \label{tab:categorical_overview}
\end{table}

\section{Related Work}

\paragraph{Patch Pruning} has emerged as a key strategy for mitigating the quadratic computational complexity problem of ViTs, by removing less important features~\cite{khan2022TransfVisionSurvey}.
Most existing methods rely on classification-based metrics to identify prunable patches~\cite{fayyaz2022adaptivetokensamplingefficient, xu2021evovitslowfasttokenevolution, Tang_2022_CVPR, liang2022evit}.
For instance, Evo-ViT~\cite{xu2021evovitslowfasttokenevolution} and STAR~\cite{zhang2024synergistic} utilize the  classification-token as patch importance. 
Rather than using the attention mechanism, DynamicViT~\cite{Rao_2021_Neurips} introduces dedicated token-selection layers that determine patch importance in a learnable, end-to-end manner.
However, all these methods rely on the classification token for masking, which limits the application in dense predictions tasks, such as segmentation.
Therefore, Tang et al.~\cite{tang2023dynamictokenpruningplain} proposed an early-exit strategy by pruning patches once they reach a certain segmentation confidence.
The work of Liu et al.~\cite{Liu_2024_WACV} introduces a mechanism for dynamic patch reactivation that preserves features for later use.
In addition to the classic patch reduction, recent methods have introduced token fusion~\cite{Kim_2024_WACV, lee2024multi, liang2022evit} to compress multiple features into a smaller set.
One such approach is TPS, which fuses pruned patches into their most similar counterparts~\cite{wei2023tps}. 
Nevertheless, all mentioned pruning strategies primarily focus on sparsifying deeper layers, while keeping the earlier ones dense.

In contrast, PaPr~\cite{mahmud2024papr} uses a constant patch sparsity throughout all layers.
Therefore, PaPr applies an auxiliary model to generate a single pruning mask, which is used for the entire model.
However, the resulting computational overhead outweighs the savings achieved through token reduction.
RGTP~\cite{dinai_2025_ICCVW} recently utilized video context for training-free patch pruning, its application remains limited to object detection. 
Our work demonstrates that for more complex tasks like \textit{VIS}, it is critical to maintain a non-constant patch sparsity throughout the network's depth to preserve high segmentation performance.

\paragraph{Video Instance Segmentation (VIS)} requires the simultaneous detection, segmentation, and tracking of object instances~\cite{Yang2019vis}.
Many VIS frameworks utilize Mask R-CNN~\cite{he2018maskrcnn} for instance-level predictions, with works like SeqFormer~\cite{wu2021seqformer} adding sequential transformers to capture inter-frame context.
More recently, Mask2Former~\cite{Cheng_2022_CVPR} has established itself as the new standard for segmentation tasks.
This framework leverages masked attention to refine instance-level predictions across diverse segmentation tasks.
Building upon this work, several methods have integrated temporal tracking mechanisms into the Mask2Former architecture to extend its capabilities to the VIS domain~\cite{ying2023ctvis, Zhang_2023_ICCV, huang2022minvis, zhan2022rovis, zheng2024syncvis}.
For instance, MinVis~\cite{huang2022minvis} utilizes query matching for online tracking.
DVIS~\cite{Zhang_2023_ICCV} introduces a multi-stage tracking framework that by refining the temporal instance associations in an offline stage to ensure long-term consistency.
Instead, ROVIS~\cite{zhan2022rovis} introduces lightweight mechanism for robust online tracking, by utilizing a resource-efficient frame-to-frame training strategy.

\section{From Image to Video Patch Pruning}
\label{sec:from_img2vid_pp}

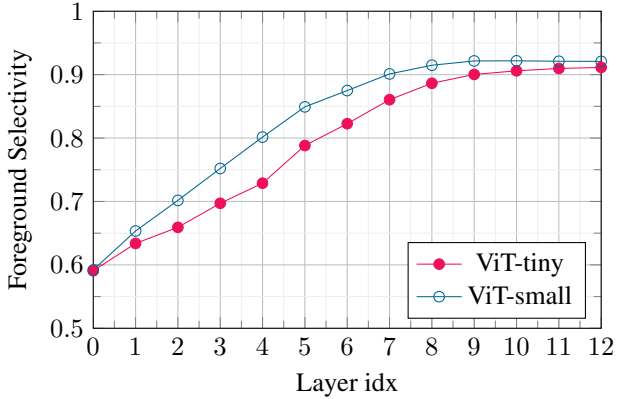
\begin{figure}[t]
\centering
\begin{tikzpicture}
\begin{axis}[
    xmin = 0, xmax = 12,
    ymin = 0.5, ymax = 1.0,
    xtick distance = 1,
    ytick distance = 0.10,
    grid = both,
    minor tick num = 1,
    major grid style = {lightgray},
    minor grid style = {lightgray!25},
    width = \linewidth,
    height = 5.8cm,
    legend pos = south east,
    ylabel=Foreground Selectivity,
    xlabel=Layer idx,
]


\addplot[OrangeRed, mark = *] table [x=idx, y expr=\thisrow{acc_tiny_ytvis21}/100, col sep=comma] {plots/table_acc_per_layer.csv};
\addlegendentry{ViT-tiny}
\addplot[MidnightBlue, mark = o] table [x=idx, y expr=\thisrow{acc_small_ytvis21}/100, col sep=comma] {plots/table_acc_per_layer.csv};
\addlegendentry{ViT-small}

\end{axis}
\end{tikzpicture}
\vspace{-15pt}
\caption{Foreground Selectivity (FGS) across layers. 
This plot shows the ability of feature \textit{x} to identify patches belonging to an object instance.
Initial features (idx: 0) and early blocks (idx: 1-3) significantly lack in Foreground Selectivity.}
\vspace{-5pt}
\label{fig:fgs}
\end{figure}

Current patch pruning methods~\cite{Wei_2023_CVPR, liang2022evit, Liu_2024_WACV, tang2023dynamictokenpruningplain} require the first layers to be dense, which limits their overall feature sparsity.
To further understand why these approaches fail when patch pruning is applied within the first layers, we analyze the ability of each intermediate layer to retain foreground-selective information.
Specifically for instance segmentation tasks, background elements must not be segmented, making them prunable candidates.
To this end, a dedicated foreground classifier is trained for each intermediate feature $x_i$ of the dense model, to distinguish between foreground and background patches.
The foreground probability is denoted as $p_{fg}(x_i)$.
As the background inherently holds a higher prior, we utilize a weighted Cross-Entropy Loss~\cite{9319440} to ensure class balance.
A more detailed description to this experiment is show in Appendix~\ref{supl:foreground_selectivity}.
The Foreground Selectivity (\textit{FGS}) is defined as the Accuracy over all patches within $x_i$, which measures the ratio of correctly identified patches.

We evaluate the \textit{FGS} of the ViT-Adapter model \cite{chen2022vitadapter} trained on the Youtube-VIS 2021 dataset~\cite{Yang2019vis}.
As shown in Figure \ref{fig:fgs}, \textit{FGS} increases almost linearly with network depth.
Starting at $\approx\!60\%$ at layer 0 (prior to the first processing block), the FGS-score reaches $92\%$ in the deepest layers of both ViT-Tiny and Small. 
Note that the lower bound for \textit{FGS} is $50\%$, which is equivalent to a randomly sampled mask.
Layer $0$ surpasses this baseline, indicating that raw image features inherently provide a $10\%$ gain in foreground information before processing begins.
Overall, foreground patches are poorly identified in the first half of the network, limiting the effectiveness of patch pruning.
Consequently, employing features from early layers (0-3) results in almost randomly sampled pruning masks, as these features are not capable to identify relevant foreground patches.
This leads to a loss of object related information for the instance segmentation task.

The results of this ablation study demonstrate that mandatory features for reliable foreground identification are only given in the deeper layers (specifically, from layer 6 onward). 
Existing patch pruning frameworks, including DynamicViT~\cite{Rao_2021_Neurips}, EViT~\cite{liang2022evit}, and TPS~\cite{Wei_2023_CVPR}, typically perform feature reduction at fixed stages of layers 3, 6 and 9.
Our analysis substantiates the incremental sparsification steps common in these architectures, as the discriminative capacity remains low throughout the first half of the network before reaching saturation at layer 9.
Nonetheless, this observed lack of foreground selectivity in the initial layers explains the failure of image-based patch pruning methods when applied to earlier features. 
To mitigate this problem, we make use of temporal prior knowledge derived from the video sequence to guide the early selection of foreground-relevant patches.

\begin{figure*}[t]
    \vspace{-0pt}
    \centering
    \includegraphics[width=1.\linewidth]{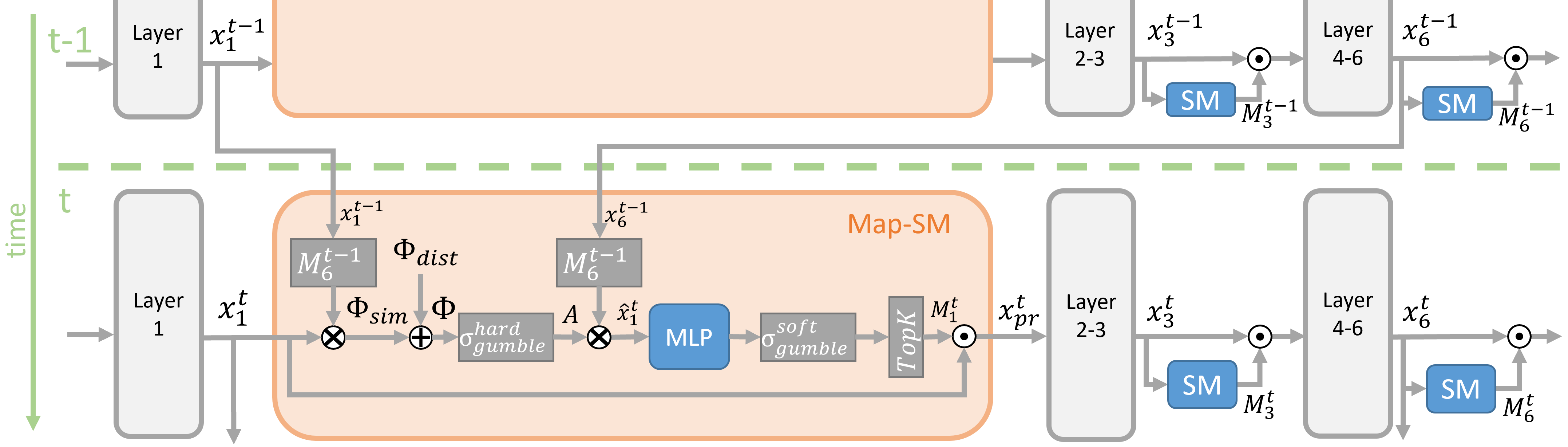}
    \vspace{-10pt}
    \caption{Mapping-Selective Module (Map-SM).   
    Using the features from preceding frame ($x^{t-1}_{1}$), \textit{Map-SM} calculates an association matrix $A$ to establish patch correspondences in order to map high-level features $x^{t-1}_{6}$ onto the current frame.
    Mask $M_6^{t-1}$ ensures that only the remaining patches from the preceding frame are considered for the mask selection process.
    }
    \vspace{-5pt}
    \label{fig:map_sm}
\end{figure*}

\section{Video Patch Pruning (VPP)}
\label{sec:patch_pruning}

\subsection{Preliminaries}
\label{sec:preliminaries}

The Vision Transformer (ViT) architecture processes an input image $\mathcal{I}\in^{3\times H\times W}$ by decomposing it into a sequence of $N$ non-overlapping patches~\cite{Vaswani2017AttnIsAllYouNeed}. 
Each patch is flattened and linearly projected to form the feature sequence $x\in^{NxE}$, where $E$ is the embedding dimension. 
This sequence is then processed through a series of $L$ transformer layers. 
Within each layer, the features are updated according to the relation $x_i = f_i(x_{i-1})$, where $f_i$ denotes the combination of the Multi-Head Self-Attention (MHSA) and Multi-Layer Perceptron (MLP) block.

\subsection{Mapping Selective Module}
\label{sec:map_selec_mod}

To address the in Sec.~\ref{sec:from_img2vid_pp} explained problem of insufficient Foreground Selectivity in earlier layers, we introduce the \textit{Mapping-Selective Module} (\textit{Map-SM}), a fully differentiable patch pruning strategy for earlier layers.
Map-SM temporarily aligns previous, high-level features to the current frame in order to estimate a binary pruning Mask $M_1^t$.

The mapping process is detailed in Figure~\ref{fig:map_sm}. 
First, the similarity scores $\Phi\in\mathbb{R}^{N_{t} \times N'_{t-1}}$ are calculated between features $x_1^t$ and $x_1^{t-1}$, where $N_{t}$ denotes the number of dense patches at timestep $t$, while $N'_{t-1}$ represents the number of previous patches retained after pruning:
\begin{equation}
    \Phi = (x_1^{t} \cdot (x_1^{t-1} \odot M_6^{t-1})^\text{T}) + \Phi_{dist}.
\end{equation}
The pruning mask $M_6^{t-1}$ of layer $6$ is applied by Hadamard product $\odot$ and reduces the mappable features to the remaining ones from the previous frame.
Constant $\Phi_{dist}$ is added to penalize patch associations with high spatial distance, prioritizing local patch associations.
Based on the similarity score $\Phi$, an association matrix $A$ is computed to map all remaining, previous patches onto the current frame. 
Matrix $A$ is conditioned on the temperature $\tau$ and computed via:
\begin{equation}
    A = \sigma^{hard}_{gumble}(\Phi, \tau).
\end{equation}
Note that $A \in \{0,1\}^{N_t \times N'_{t-1}}$ can be non-quadratic, as $N'_{t-1}$ denotes the remaining patches after pruning by mask $M_6^{t-1}$.
$\sigma^{hard}_{gumble}$ represents the Gumbel-Softmax function~\cite{jang2017gumbelsoftmax}, providing a differentiable relaxation of discrete categorical distributions.
Specifically, $\sigma^{hard}_{gumble}$ generates a one-hot-encoded matrix to map each patch from $x^{t-1}$ to it's successor in $x^{t}$, whereas its soft counterpart, $\sigma^{soft}_{gumble}$, produces continuous probability distributions normalized to~$1$.
During training, the Gumbel-noise is used to encourage exploration of similar patch associations, ensuring the selection process does not collapse into repetitive patterns.
Gumbel-Softmax ensures the association mask to be differentiable, allowing the input feature to acquire adjustments to improve the feature alignment during training.
Overall, the association matrix $A$ facilitates the temporal mapping of highly foreground-selective features $x_6^{t-1}$ from the previous video feature onto the current one $\hat{x}^t_1$.
The feature mapping is calculated by:
\begin{equation}
    \hat{x}^t_1 = A \cdot (x_6^{t-1} \odot M_6^{t-1}).
\end{equation}

Until this stage, the \textit{MAP-SM} module operates without any additional parameters.
To estimate the patch probability 
\begin{equation}
    p^t(\hat{x}^t_1) = \sigma^{soft}_{gumble}(MLP(\hat{x}^t_1), \tau)
\end{equation} 
for the current step, we apply a compact, 2-layer Multilayer Perceptron (MLP).
This MLP is designed for minimal computation, utilizing a hidden size of $e/2$ and an output size of $2$. 
In contrast to the discrete masks produced by $\sigma^{hard}_{gumble}$, the Soft Gumbel-Softmax $\sigma^{soft}_{gumble}$ generates a continuous probability map, further explained in Appendix~\ref{sec:gumbleSm}.
Masking the foreground ideally would lead to an incorrect mapping, as the background features in the successor frame would be fully assigned by foreground patches.
To mitigate this problem, we use Gumbel-noise controlled by a temperature $\tau$ to sparsely activate background patches.
This sparse background activation also enables the detection of new object instances, as few foreground-patches from the previous frame can be mapped onto a larger receptive area, allowing complete feature coverage.
The final mask $M_1^t$ is then generated by selecting the top-k elements of $p^t$.
While the introduced Map-SM-Module is only applied once after layer $1$, we further add 3 pruning stages after layers 3, 6 and 9 to reduce the computed features in network-depth.
According to~\cite{Liu_2024_WACV}, this 3 stages are pruned via the Selective Module $SM$:
\begin{equation}
M_l^t = SM(x)=\sigma^{hard}_{gumble}(MLP(x), \tau).
\end{equation}

In summary, \textit{MAP-SM} addresses two required properties for video-temporal feature alignment:

\begin{enumerate}
    \item Motion compensation: 
    The mapping ensures accurate spatial shifts of patches, compensating for the movement of foreground objects across consecutive frames.
    
    \item Sparse background activation: \textit{Map-SM} sparsely activates background patches to mitigate feature misalignment and to ensure the detection of new object instances.

\end{enumerate}

\subsection{Sparse Video Instance Segmentation}

In order to reach the target feature sparsity we use an additional sparsification loss for regularization.
For the Map-SM this loss is defined as:
\begin{equation}
    \mathcal{L}_{sp}^{Map} (p^t) = \left(\left(\frac{1}{|p^t|} \sum p^t\right) - \kappa_{init}\right)^2.
    \label{eq:l_sp_map}
\end{equation}
The module employs a top-k selection strategy using patch probabilities $p^t\in[0,1]^{N_t}$. 
To ensure $p^t$ effectively distinguishes feature importance rather than directly representing patch sparsity, we optimize it toward a mean value of $\kappa_{init}=0.5$.
In contrast, the SM in layer $l\in\{3,6,9\}$ is optimized toward target keep ratio $\kappa_l$, directly applied on binary mask $M_l\in\{0,1\}$:
\begin{equation}
    \mathcal{L}_{sp}^{SM} (M_l) = \left(\left(\frac{1}{|M_l|} \sum M_l\right) - \kappa_l\right)^2.
    \label{eq:l_sp_sm}
\end{equation}

This loss enforces a mean density constraint, whereas performing normalization at the batch level mitigates a penalty for deviating mask sizes, as long as the average density is satisfied.
Consequently, \textit{SM} enables dynamic scaling of the mask size if necessary.
A more detailed explanation to the setting is explained in Appendix~\ref{app:train_settings}.

\section{Experiments}

We evaluate our pruning method on an end-to-end VIS-framework designed for joint segmentation and tracking.
Therefore, we use ViT-Adapter~\cite{chen2022vitadapter} as backbone, Mask2Former~\cite{Cheng_2022_CVPR} to generate segmentation masks and ROVIS~\cite{zhan2022rovis} as tracking head.
ROVIS is particularly advantageous as it requires only two consecutive timesteps for training, achieving high-quality temporal results.
Our training protocol follows the official procedure of ROVIS, using the in Eq.~\ref{eq:l_sp_map} and ~\ref{eq:l_sp_sm} defined sparsification loss for regularization.

All models are trained for 6 epochs on Youtube-VIS~2019 and 2021 dataset~\cite{Yang2019vis}, following the ROVIS training protocol.
This includes a learning-rate of $2.5 \times 10^{-5}$ and a lr-decay of $0.1$ after $4$ epochs.
Note that all implementations are integrated within the mmDetection framework~\cite{mmdetection}.

\subsection{Impact of Patch Pruning}

\begin{figure}
\centering
\begin{tikzpicture}
\begin{axis}[
    width=\linewidth,
    height=5.3cm,
    xlabel={$\leftarrow$ {\scriptsize{Sparse}} | {\normalsize{ Patch Keep Ratio (\%)}} | {\scriptsize{Dense}} $\rightarrow$},
    ylabel={$\leftarrow$ {\scriptsize{High}} | {\normalsize{\makecell{AP loss}}} | {\scriptsize{Low}} $\rightarrow$},
    xmin=30, xmax=105,
    ymin=-1, ymax=15,
    y dir=reverse,
    grid=major,
    ytick distance=2,
    xtick distance=10,
    mark size=2pt,
    line width=1pt,
]

\addplot[mark=square, black, mark size=3pt, legend image post style={mark size=2pt}] coordinates {(100, 0)};
\label{pgfplots:c1r1}
\addplot[only marks, mark=*, black, mark size=1pt, legend image post style={mark size=2pt}] coordinates {(100, 0)};
\label{pgfplots:c2r1}

\addplot[mark=square, line width=1pt, Red] table [x expr={100-\thisrow{PruneRate}}, y=ErrorRate, col sep=comma] {plots/error_rate/svit_ytvis19_vitSmall.csv};
\label{pgfplots:c1r2}

\addplot[mark=*, line width=1pt, Red] table [x expr={100-\thisrow{PruneRate}}, y=ErrorRate, col sep=comma] {plots/error_rate/svit_ytvis21_vitSmall.csv};
\label{pgfplots:c2r2}

\addplot[mark=square, line width=1pt, Orange] table [x expr={100-\thisrow{PruneRate}}, y=ErrorRate, col sep=comma] {plots/error_rate/dynvit_ytvis19_vitSmall.csv};
\label{pgfplots:c1r4}

\addplot[mark=*, line width=1pt, Orange] table [x expr={100-\thisrow{PruneRate}}, y=ErrorRate, col sep=comma] {plots/error_rate/dynvit_ytvis21_vitSmall.csv};
\label{pgfplots:c2r4}


\addplot[mark=square, line width=1pt, Blue] table [x expr={100-\thisrow{PruneRate}}, y=ErrorRate, col sep=comma] {plots/error_rate/vpp_ytvis19_vitSmall.csv};
\label{pgfplots:c1r3}

\addplot[mark=*, line width=1pt, Blue] table [x expr={100-\thisrow{PruneRate}}, y=ErrorRate, col sep=comma] {plots/error_rate/vpp_ytvis21_vitSmall.csv};
\label{pgfplots:c2r3}

\node[draw,fill=white,inner sep=0pt,above left=0.5em] at (rel axis cs:1, 0) {\small
    \begin{tabular}{rcc}
    YTVIS & 2019 & 2021 \\
    Vit-Ada & \ref{pgfplots:c1r1} & \ref{pgfplots:c2r1} \\
    DynVit & \ref{pgfplots:c1r4} & \ref{pgfplots:c2r4} \\
    SViT & \ref{pgfplots:c1r2} & \ref{pgfplots:c2r2} \\
    VPP (ours) & \ref{pgfplots:c1r3} & \ref{pgfplots:c2r3} \\
    \end{tabular}};

\end{axis}

\end{tikzpicture}
\vspace{-5pt}
\caption{Performance Loss in AP vs. Pruned Patches on Youtube-VIS 2019 and 2021, using model size small. 
}
\vspace{-10pt}
\label{fig:error_rate}
\end{figure}
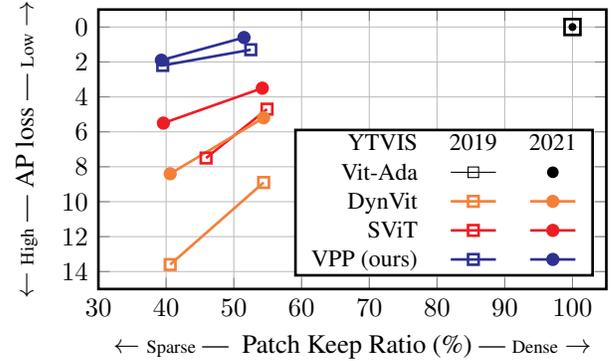 
\begin{table*}
    \vspace{0.0cm}
    \centering
    \setlength{\tabcolsep}{4pt}
    \resizebox{1.0\linewidth}{!}{
    \begin{tabular}{ccc|ccccc|ccccc}
        \toprule
        

        \multicolumn{3}{c|}{model} & \multicolumn{5}{c|}{tiny} & \multicolumn{5}{c}{small} \\
        \cline{1-13}
        \addlinespace[1pt]
        
        dataset & \makecell{goal \\PKR (\%)} & method 
        & PKR (\%) $\downarrow$ & $mAP_{bbox}\uparrow$ & $mAP_{segm} \uparrow$ & $AP \uparrow$ & $\Delta AP$
        & PKR (\%) $\downarrow$ & $mAP_{bbox}\uparrow$ & $mAP_{segm} \uparrow$ & $AP \uparrow$ & $\Delta$ AP \\

        \cline{1-13}
        \addlinespace[1pt]
        \multirow{7}{*}{\rotatebox{90}{YTVIS 21}}
        & 100 & ViT-Adapter~\cite{chen2022vitadapter} 
        & 100 & 49.1 & 44.8 & 42.2 & - 
        & 100 & 57.3 & 52.3 & 50.4 & - 
        \\
        \cdashline{2-13}
        \addlinespace[1pt]
        & \multirow{4}{*}{55} & DynViT~\cite{Rao_2021_Neurips} 
        & 54.4 & 44.1 & 40.4 & 36.9 & -5.3 
        & 54.4 & 53.8 & 48.8 & 45.2 & -5.2 
        \\
        & & TPS~\cite{wei2023tps} 
        & 54.4 & 46.9 & 42.8 & 38.9 & -3,3 
        & 54.4 & 54.9 & 50.2 & 46.2 & -4.2 
        \\
        & & SViT~\cite{Liu_2024_WACV} 
        & 54.0 & 46.7 & 42.9 & 39.2 & -3.0 
        & 54.2 & 55.8 & 51.1 & 46.9 & -3.5 
        \\
        & & \textbf{VPP (ours)}  
        & 52.7 & \textbf{49.9} & \textbf{45.7} & \textbf{42.3} & \textbf{+0.1} 
        & 51.5 & \textbf{56.1} & \textbf{51.3} & \textbf{49.8} & \textbf{-0.6} 
        \\
        \cdashline{2-13}
        \addlinespace[1pt]
        & \multirow{4}{*}{40} & DynViT~\cite{Rao_2021_Neurips} 
        & 40.6 & 29.9 & 27.8 & 24.0 & -18.2 
        & 40.6 & 48.9 & 44.2 & 42.0 & -8.4 
        \\
        & & TPS~\cite{wei2023tps} 
        & 40.6 & 41.4 & 37.8 & 35.0 & -7.2 
        & 40.6 & 50.7 & 46.3 & 43.3 & -7,1 
        \\
        & & SViT~\cite{Liu_2024_WACV} 
        & 42.0 & 46.5 & 42.9 & 39.3 & -2.9 
        & 39.6 & 52.7 & 48.1 & 44.9 & -5.5 
        \\
        & & \textbf{VPP (ours)}  
        & 40.9 & \textbf{47.6} & \textbf{43.6} & \textbf{40.9} & \textbf{-1.3} 
        & 39.3 & \textbf{54.3} & \textbf{49.9} & \textbf{48.5} & \textbf{-1.9} 
        \\

        \cline{1-13}
        \addlinespace[1pt]
        \multirow{7}{*}{\rotatebox{90}{YTVIS 19}} 
        & 100 & ViT-Adapter~\cite{chen2022vitadapter}
        & 100 & 49.0 & 45.7 & 44.6 & - 
        & 100 & 58.3 & 54.3 & 52.4 & - 
        \\
        \cdashline{2-13}
        \addlinespace[1pt]
        & \multirow{4}{*}{55} & DynViT~\cite{Rao_2021_Neurips} 
        & 54.4 & 44.9 & 41.8 & 39.5 & -5.1 
        & 54.4 & 49.7 & 46.7 & 43.5 & -8.9 
        \\
        & & TPS~\cite{wei2023tps} 
        & 54.4 & 45.6 & 41.9 & 39.5 & -5.1 
        & 54.4 & 51.9 & 48.9 & 45.6 & -6,8 
        \\
        & & SViT~\cite{Liu_2024_WACV} 
        & 55.5 & 44.9 & 42.0 & 40.2 & -4.4 
        & 54.9 & 52.7 & 49.4 & 47.7 & -4.7 
        \\
        & & \textbf{VPP (ours)} 
        & 54.5 & \textbf{47.9} & \textbf{44.4} & \textbf{42.2} & \textbf{-2.4} 
        & 52.5 & \textbf{55.9} & \textbf{52.5} & \textbf{51.1} & \textbf{-1.3} 
        \\
        \cdashline{2-13}
        \addlinespace[1pt]
        & \multirow{4}{*}{40} & DynViT~\cite{Rao_2021_Neurips} 
        & 40.6 & 28.1 & 25.7 & 22.9 & -21.7 
        & 40.6 & 44.5 & 41.1 & 38.8 & -13.6 
        \\
        & & TPS~\cite{wei2023tps} 
        & 40.6 & 38.7 & 35.5 & 31.8 & -12.8 
        & 40.6 & 46.2 & 43.1 & 40.8 & -11.6 
        \\
        & & SViT~\cite{Liu_2024_WACV} 
        & 42.4 & 42.8 & 39.8 & 37.1 & -7.5 
        & 45.9 & 50.9 & 47.7 & 44.9 & -7.5 
        \\
        & & \textbf{VPP (ours)} 
        & 41.9 & \textbf{46.7 }& \textbf{42.6} & \textbf{41.0} & \textbf{-3.6} 
        & 39.5 & \textbf{55.3} & \textbf{52.2} & \textbf{50.2} & \textbf{-2.2} 
        \\

        \bottomrule
    \end{tabular}
    }
    \vspace{-5pt}
    \caption{Average Precision on YTVIS 19/21 Dataset.
    The results are grouped into similar sparsity levels, measured by PKR.
    Instance segmentation metrics $mAP$ (bbox and segm) are reported, independent of tracking ID association.
    The best scores are highlighted as bold.
    }
    \vspace{-10pt}
    \label{tab:performance}
\end{table*}

This section analyses the performance of different patch pruning methods at comparable sparsity levels.
As the target keep ratio in dynamic pruning methods deviates from the actual patch utilization~\cite{Liu_2024_WACV}, we measure sparsity via the Patch Keep Ratio (PKR), defined as the mean patch activation across L layers:
\begin{equation}
    PKR = \frac{1}{L}\sum^L_{l=0} \frac{|N^l_{sparse}|}{|N^l_{dense}|}.
\end{equation}
$N^l_{sparse}$ and $N^l_{dense}$ represent the number of remaining and original patches, respectively.

\paragraph{Tracking Performance}

We compare the performance of our \textit{VPP} with SOTA pruning methods.
To quantify the segmentation and tracking performance, we measure the Average Precision (AP), also used in Video Instance Segmentation~\cite{Yang2019vis}.

Tab.~\ref{tab:performance}, shows the performance results for ViT-Tiny and Small on Youtube-VIS~2019 and 2021 datasets~\cite{Yang2019vis}.
The sparsity levels are grouped into target PKR of 55\% and 40\%.
Corresponding to Tab.~\ref{tab:performance}, Fig.~\ref{fig:error_rate} visualizes the performance loss to the dense model, over different feature sparsities.

\textit{VPP} consistently demonstrates superior performance results across all evaluated model sizes and datasets.
Notably, on the YouTube-Vis 2021 dataset, \textit{VPP} achieves substantial computational efficiency with minimal loss in performance. 
For model size small, \textit{VPP} successfully reduces input patches to 51.5\% and 39.3\% PKR, while only loosing 0.6\% and 1.9\% in AP, respectively.
Furthermore, \textit{VPP} operates with only 52.7\% of the patches on the tiny model, yet it still outperforms the dense baseline by +0.1\% AP.
In comparison, ''classic'' image-based pruning methods, such as SViT~\cite{Liu_2024_WACV} and Dynamic-ViT~\cite{Rao_2021_Neurips}, usually operate on a patch density of 60-70\% PKR.
The observed PKR of \textit{VPP} and SViT slightly deviates from the target sparsity, as both frameworks dynamically scale the size of the pruning mask to account for varying object sizes.

\begin{table}
    \centering
    \vspace{5pt}
    \resizebox{1.0\linewidth}{!}{
    \begin{tabular}{cc|c|ccc}
        \toprule
         PKr (\%) & method & $IoI$ & $IoI_{\underline{S}}$ & $IoI_{\underline{M}}$ & $IoI_{\underline{L}}$ \\
         
         \midrule
         \multirow{2}{*}{55\%} 
         & SViT & 73.7\% & 77.2\% & 68.8\% & 65.8\% \\ 
         & VPP (ours)  & \textbf{82.3\%} & \textbf{87.3\%} & \textbf{76.2\%} & \textbf{72.2\%} \\ 
         
         \cdashline{1-6}
         \addlinespace[1pt]
         
         \multirow{2}{*}{40\%} 
         & SViT & 62.7\% & 67.5\% & 56.0\% & 52.1\% \\ 
         & VPP (ours)  & \textbf{73.4\%} & \textbf{79.5\% }& \textbf{65.9\%} & \textbf{60.7\%} \\ 
         \bottomrule
    \end{tabular}
    }
    \vspace{-5pt}
    \caption{Intersection over Instance (IoI) of ViT-Small on YoutubeVIS 2021.
    Scores are separated by instance size: \underline{S}mall ($\leq 10\%$), \underline{M}edium $(10\%-20\%)$, and \underline{L}arge $(>20\%)$.
    }
    \vspace{-10pt}
    \label{tab:ioi}
\end{table}

\paragraph{Instance Segmentation Performance}
\label{sec:inst_seg_performance}

The tracking metric $AP$ strictly depends on consistent temporal ID-associations, that may not fully reflect the quality of individual frame-level detection and segmentation accuracy.
To provide a holistic performance assessment, we further evaluate the instance wise  $mAP$ metric~\cite{Lin2014MicrosoftCC} for bounding boxes (bbox) and segmention mask (segm), also shown in Table~\ref{tab:performance}.

Our proposed method \textit{VPP} achieves superior segmentation results on both datasets, notably achieving a $mAP_{bbox}$ of 49.9\% and $mAP_{segm}$ of 45.7\% on Youtube-VIS~2021, demonstrating a clear improvement over dense model ViT-Adapter (tiny).
This strong performance is reached despite using a significantly lower patch density of only 52.7\%.
Moreover, \textit{VPP} significantly surpasses the sparse SViT baseline in all settings, achieving an improved performance of up to 3.0\% and 2.4\% in $mAP_{bbox}$ and $mAP_{segm}$, respectively, on Youtube-VIS 2019 using model tiny.

In addition to this quantitative performance results, we provide qualitative examples of the initial pruning masks in Appendix~\ref{app:qualitativePruningResults}.
The initial masks, generated by Map-SM, highly focus on foreground objects and consistently prune background features, demonstrating the model's discriminative precision.

\subsection{Image vs. Video Patch Pruning}

Image-based patch pruning processes each frame independently, while in this work we use the video context to allow early feature sparsification.
To illustrate the operational advantage of our \textit{VPP}-approach over image-based counterparts, we adopt the top-performing \textit{SViT} as our primary baseline to provide a rigorous comparative analysis.
While the image-based method \textit{SViT} prioritizes foreground patch activation, it processes the initial layers almost densely.
Conversely, \textit{VPP} identifies and removes background tokens immediately after the first block, significantly reducing the computational load and allowing subsequent inference steps to focus on the crucial foreground patches. 

To further quantify how effectively our method captures the foreground region across a diverse dataset, we introduce the Intersection over Instance (IoI) metric, which is closely related to Intersection over Union (IoU)~\cite{Rezatofighi_2019_CVPR}.
\begin{equation}
    IoI(M) = \frac{1}{I L} \sum_{i=1}^I \sum_{l=1}^L \frac{|M_{pred}^{i,l} \cap M_{GT}^i|}{|M_{GT}^i|}
\end{equation}
This metric measures the patch activation averaged over all object instances $I$ and layer $L$ between pruning mask $M_{pred}^{i,l}$ and ground truth mask $M_{GT}^i$.
Note that a perfect score of 1 indicates that every patch belonging to an object instance has not been pruned away.
The resulting \textit{IoI} scores are shown in Tab.~\ref{tab:ioi}.
\textit{VPP} reaches an instance intersection of $82.3\%$ at $55\%$ PKR, which is an improvement of $8.6\%$, compared to SViT.
Notably, \textit{VPP} can reduce the PKR down to $40\%$ while ensuring $72.4\%$ of the foreground patches to be computed.
More results to that study are shown in Appendix~\ref{app:ioi}.
In all our experiments, we observe \textit{VPP} to have an increased patch coverage for small sized object instances. 
For example, it reaches an $IoI$ score of $87.3\%$ for instances containing less than $10\%$ foreground size.
This behavior is particularly valuable for segmentation tasks since
the higher resolution of larger objects contains more redundant information~\cite{choudhury2025accelvita}.
Consequently, larger objects can be represented using a lower patch density.

\subsection{Positioning of the Masking Modules}
\label{sec:param_eval}

\begin{table}
    \centering
    \small
    \begin{tabular}{ccccccc}
        \toprule
         No. & \makecell{Map-SM \\ index i} & \makecell{SM idx \\ index i} & PKR (\%) & AP & Note\\
         \midrule
         1 & - & 0,3,6,9 & 55.9 & 40.1 & \\
         2 & - & 1,3,6,9 & 52.4 & 40.3 & \\
         3 & 0 & 3,6,9   & 51.6 & 41.5 & \\
         4 & 1 & -        & 54.2 & 40.6 & const Kr\\
         5 & 1 & 3,6,9   & 52.7 & \textbf{42.3} & (default)\\       

        
         \bottomrule
    \end{tabular}
    \vspace{-5pt}
    \caption{Positioning ablation of Map-SM and SM on dataset YoutubeVIS 2021 at PKR=55\%.
    Early patch initialization with Map-SM (no.~5) improves AP by $+2.0\%$, compared to image-based patch selection (no.~1+2).
    }
    \vspace{-5pt}
    \label{tab:module_positioning}
\end{table}

The introduced \textit{VPP} facilitates early removal of unimportant patches via the Map-SM, effectively conserving feature sparsity throughout the network depth.
In this section we evaluate the performance impact of module positioning to demonstrate the practical benefits of the video-based \textit{Map-SM} across different architectural configurations.
We define the module index such that $i=0$ is processed before the first layer.
Conversely, index $i>0$ indicates a position following the i-th processing block.

The resulting AP performances are listed in Table~\ref{tab:module_positioning}.
Overall, the best performance of $42.3\%$ is reached, if \textit{Map-SM} is positioned after Layer $1$, combined with depth-decreasing patch density via \textit{SM} modules.
In comparison to setting $3$, where Map-SM-module is positioned before layer 1, the performance drops by $0.8\%$, showing the raw-features negatively impact the feature alignment.
Pruning Frameworks such as RGTP~\cite{dinai_2025_ICCVW} and PaPr~\cite{mahmud2024papr} remove patches only once, rather than applying multi-stage reduction.
Similarly we simulate a constant PKR in setting no.~4 by skipping the SM-layers from deeper layers.
Despite the slightly increased patch usage, the model performance drops by 1.7\% down to 40.6\%.
This suggests that the initial layers should maintain a higher density to preserve essential feature information.
Notably, pruning patches solely based on the current features, without incorporating video context (settings 1 and 2), drastically decreases the AP, demonstrating the critical role of the video context.

\begin{figure}[t]
    \centering

    \begin{tikzpicture}
    \begin{axis}[
        xmin = 1, xmax = 12,
        ymin = 0.0, ymax = 1.05,
        xtick distance = 1,
        ytick distance = 0.25,
        grid = both,
        minor tick num = 1,
        major grid style = {lightgray},
        minor grid style = {lightgray!25},
        width = \linewidth,
        height = 5.3cm,
        ylabel=Patch Density,
        xlabel=Layer idx,
        legend style={
            at={(1.0, 1.0)},
            anchor=north east,
            font=\scriptsize,
        }
    ]


    \addplot[Emerald, mark = *, fill=Emerald, fill opacity=0.15] table [x = {layer_idx}, y = {kr_vpp55}] {plots/table_img_vs_video_ytvis21.csv} \closedcycle;
    
    \addplot[blue, mark = *, fill=blue, fill opacity=0.15] table [x = {layer_idx}, y = {kr_vpp40}] {plots/table_img_vs_video_ytvis21.csv} \closedcycle;

    \addplot[red, mark = *, , fill=red, fill opacity=0.15] table [x ={layer_idx}, y = {kr_svit40}] {plots/table_img_vs_video_ytvis21.csv} \closedcycle;


    \legend{
        VPP 55\% (ours), 
        VPP 40\% (ours), 
        SVIT 40 \%,
    }
    \end{axis}
    \end{tikzpicture}
    \vspace{-20pt}
    \caption{Patch density per layer. 
    \textit{VPP} removes patches after layer 1, while maintaining high patch density within the deepest layers (7-12).
    In contrast, image-based SViT requires dense initial layers, leaving the deepest layers with insufficient patch density.
    }
    \vspace{-10pt}
    \label{fig:img_vs_video_kr}
\end{figure}
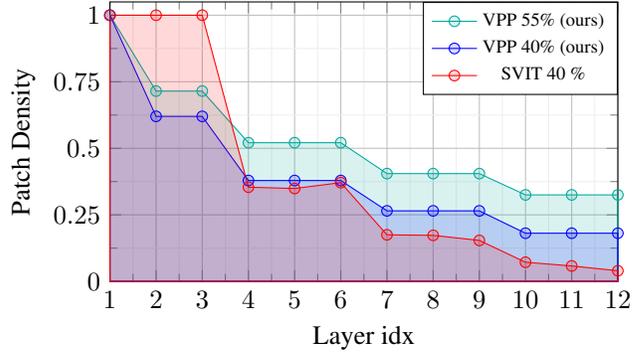


\subsection{Mask-Adaptation in Dynamic Scenes}
\label{sec:handling_dynamic_scenes}

\begin{figure*}
    \setlength{\tabcolsep}{1pt}
    \begin{tabular}{cccc|c}
        
        \includegraphics[width=0.195\linewidth, height=2.3cm, valign=m]{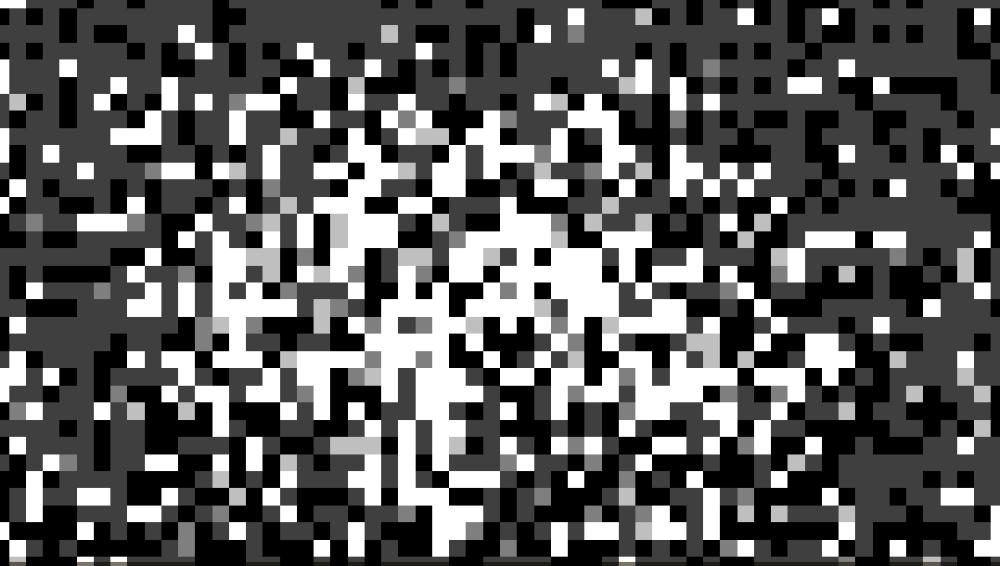} &
        \includegraphics[width=0.195\linewidth, height=2.3cm, valign=m]{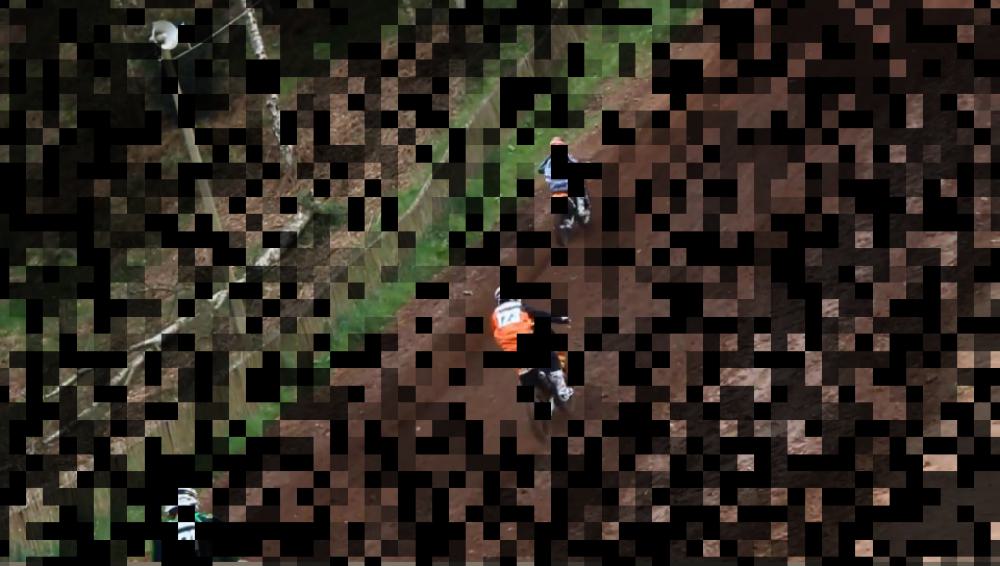} &
        \includegraphics[width=0.195\linewidth, height=2.3cm, valign=m]{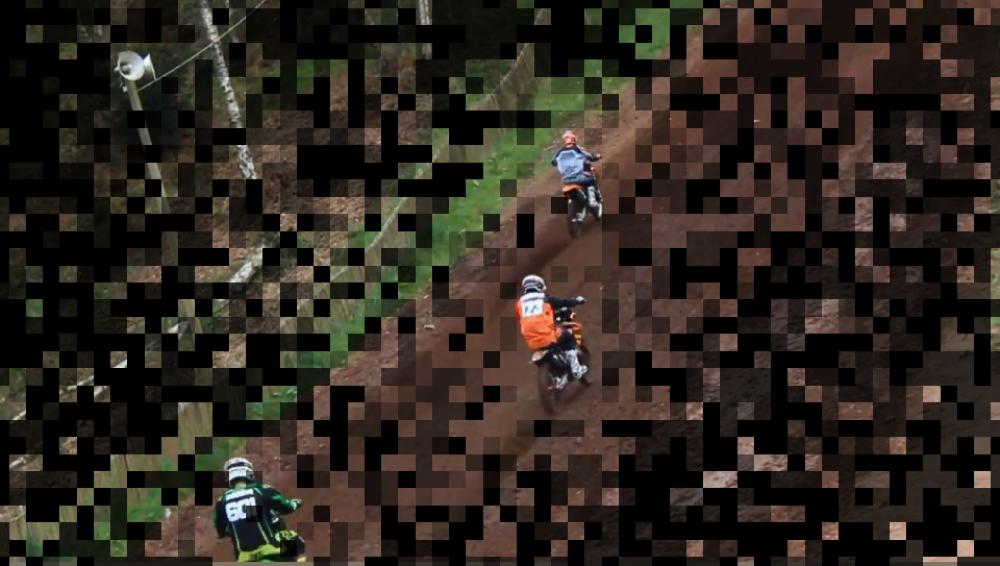} &
        \includegraphics[width=0.195\linewidth, height=2.3cm, valign=m]{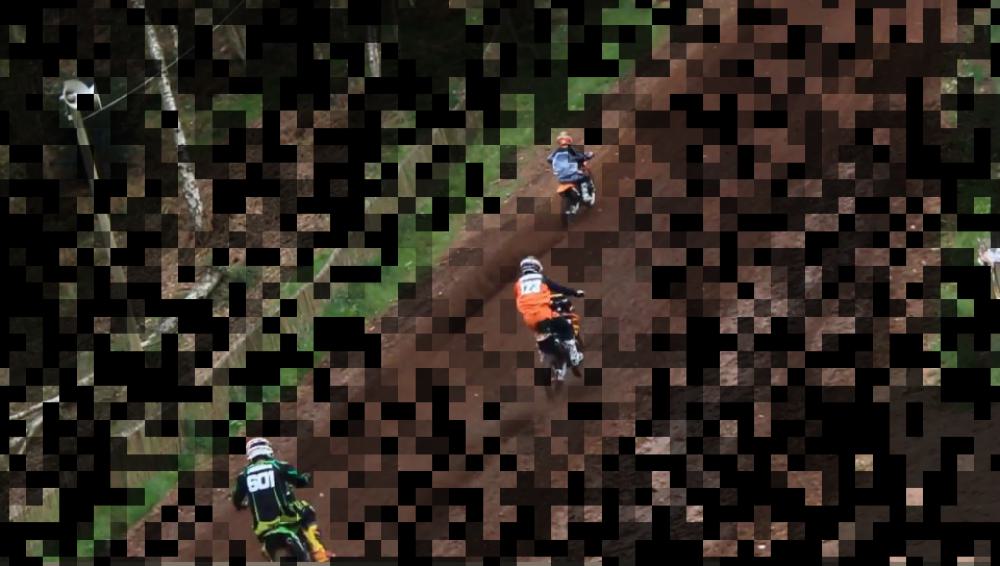} &
        \includegraphics[width=0.195\linewidth, height=2.3cm, valign=m]{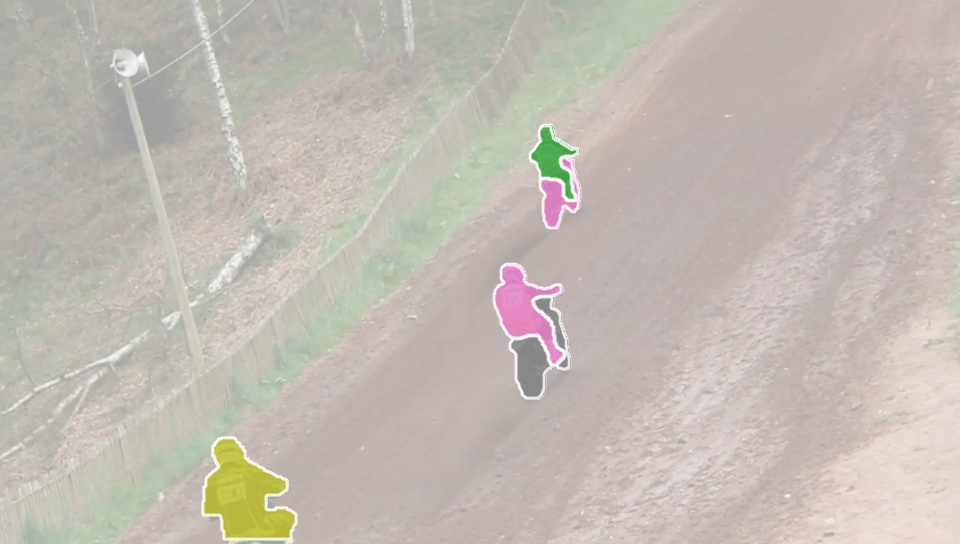} \\
        \addlinespace[2pt]

        \includegraphics[width=0.195\linewidth, height=2.3cm, valign=m]{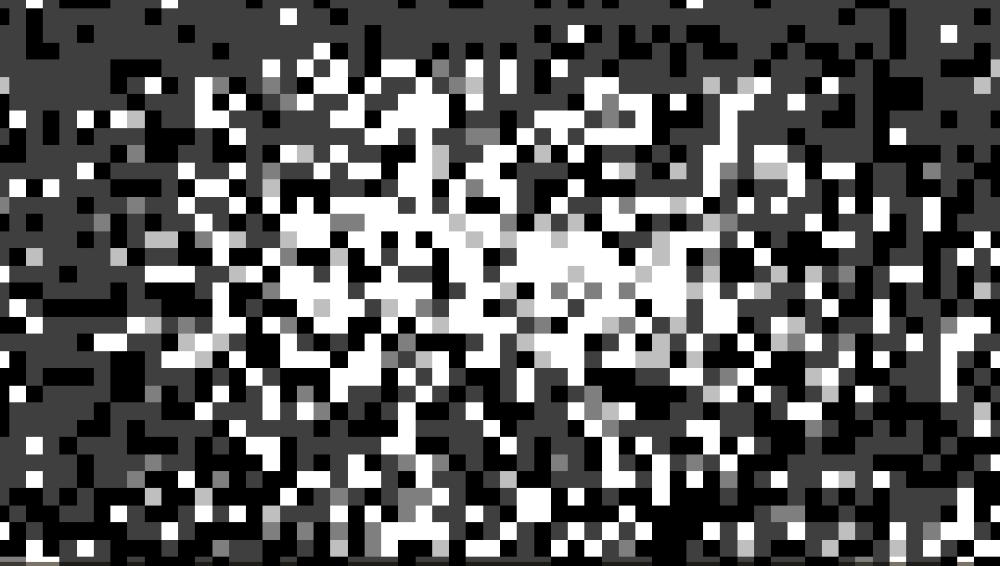} &
        \includegraphics[width=0.195\linewidth, height=2.3cm, valign=m]{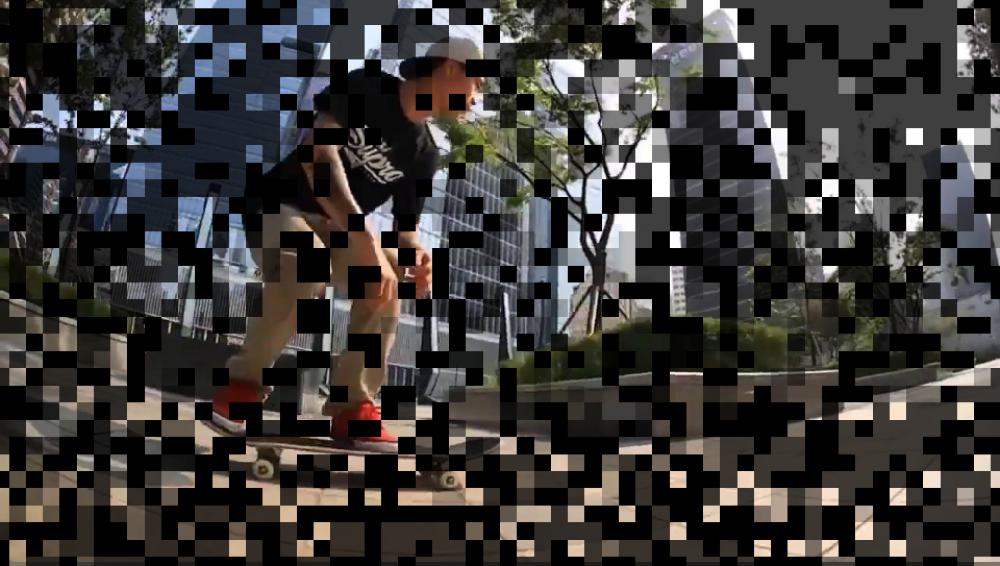} &
        \includegraphics[width=0.195\linewidth, height=2.3cm, valign=m]{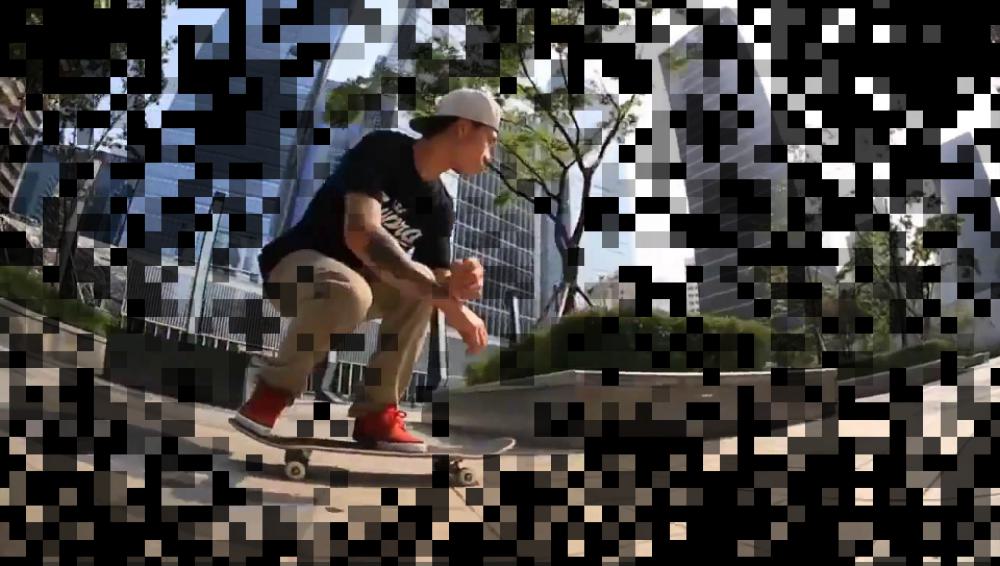} &
        \includegraphics[width=0.195\linewidth, height=2.3cm, valign=m]{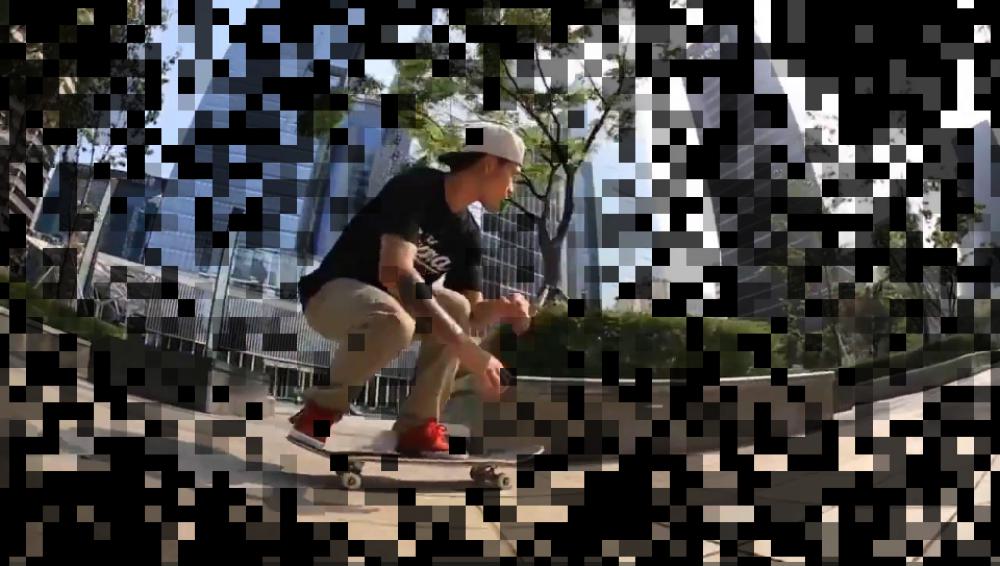} &
        \includegraphics[width=0.195\linewidth, height=2.3cm, valign=m]{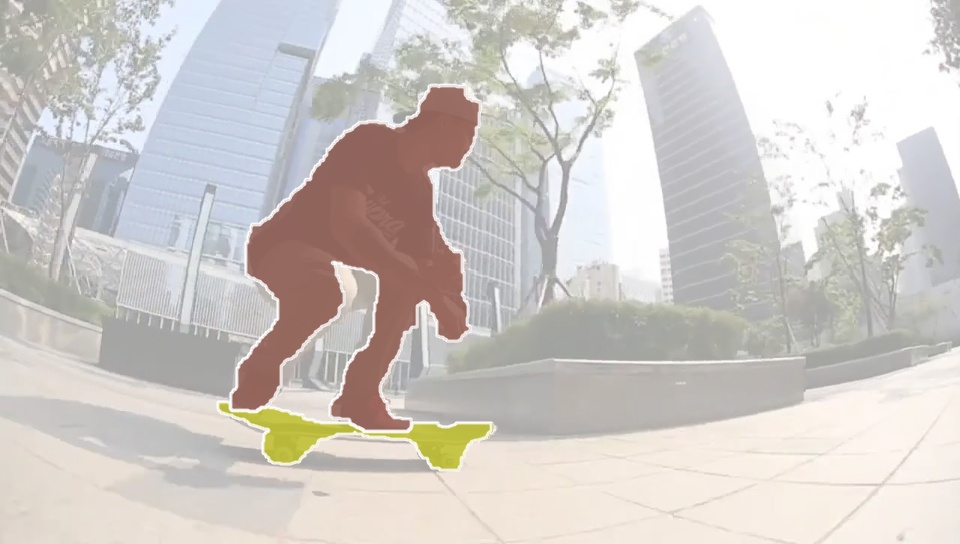} \\
        

        t=0 & t=1 & t=2 & t=3 & GT-mask at t=2 \\
    \end{tabular}
    \vspace{-5pt}
    \caption{
    Mask adaptation of \textit{VPP} at scene switches.
    Patch activity is visualized through shading: less activated patches are darker, and fully black patches are skipped.
    White, textureless inputs ($t\leq0$) led to a centrally sampled masks.
    After the video is enabled at $t=1$, \textit{VPP} is able to reallocate the foreground mask within a single timestep ($t=2$).
    The last column shows the GT-mask as reference.
    }
    \vspace{-0pt}
    \label{fig:image_sequence_inline}
\end{figure*}

The introduced \textit{VPP}-method estimates relevant foreground patches based on mapped features from the previous frame. 
This section focuses on analyzing the dynamic adaptation mechanism of the pruning mask.
Particularly, we evaluate the impact of temporal context when video content shifts abruptly, such as during a scene switch.
Therefore, \textit{VPP} is initially processed for three timesteps using a sequence of textureless, fully white images, followed by the actual video sequence.

Figure~\ref{fig:image_sequence_inline} demonstrates two examples for the dynamic adaption of \textit{VPPs} pruning mask.
Here the fully black patches are pruned in the very beginning of the model by Map-SM-layer and gray patches symbolize patches that are pruned in deeper layers using SM.
A series of blank initial frame ($t=0$) forces \textit{VPP} into centered patch sampling, highlighting its reliance on positional information for mask generation.
At $t=1$ the mask shows reduced foreground focus, as it is derived from a temporally inconsistent reference image and thus remains largely random with a slight positional bias.
However, all subsequent pruning masks in the video sequence $t\geq2$ show a high overlap with foreground objects, including the classes: person, skateboard and motorbike in this example.
Notably, in frames $t=2$ and $t=3$, \textit{VPP} fully identifies the person in the lower-left corner as a foreground object, irrespective of its off-center position.
In all videos to this experiment, \textit{VPP} was able to focus on the foreground objects after only one intermediate frame.
Consequently, even if the mask temporarily drifts toward irrelevant regions during long sequences, it successfully recovers the foreground within a single timestep.

To summarize the qualitative results in this experiment, \textit{VPP} confidently captures foreground objects 
throughout all pruning stages and ensures background patches to be activated sparsely.
This process achieves fast adaptation to unknown scene switches by only requiring one intermediate frame for mask alignment.

\subsection{Layerwise and Computational Efficiency}
\label{sec:feature_sparsity}

\begin{table}[]
    \centering
    \small
    \resizebox{1.0\linewidth}{!}{
    \begin{tabular}{c|cc|cc}
        \toprule
        method & AP $\uparrow$ & FLOPS $\downarrow$ & $FPS_{BB} \uparrow$ & $FPS_M \uparrow$\\
        \midrule
        ViT-Ada~\cite{chen2022vitadapter}  & 50.4 & 88.0 G & 17.14 & 6.58 \\
        SViT~\cite{Liu_2024_WACV}     & 46.9 & 59.6 G & 24.87 & 8.19 \\
        VPP (ours)     & 49.8 & 59.9 G & 25.94 & 8.13 \\ 
        \bottomrule
    \end{tabular}
    }
    \vspace{-5pt}
    \caption{FLOPS and Throughput, measured in Frames per Second (FPS).
    All experiments utilize the YTVIS-2021 dataset and involve pruning the ViT-Ada-small model with a 55\% PKR.
    }
    \vspace{-10pt}
    \label{tab:flops}
\end{table}

This section provides a comparative analysis of \textit{VPP}'s sparsity level as well as corresponding improvements in FLOPS and inference speed.

Therefore, Fig.~\ref{fig:img_vs_video_kr} demonstrates the patch density per layer, conditioned on the goal-PKR of 55\% and 40\%, respectively. 
The results are evaluated on Youtube-VIS~2021 dataset using ViT-Adapter-Small.
\textit{VPP} operates effectively at this high sparsity level, by removing unimportant background features in earlier layers, while still retaining 33\% in the deepest ones at a 55\% PKR.
With a PKR of 40\%, \textit{VPP} prunes the input features down to a 60\% patch density immediately following the first ViT layer. 
This early-stage selection effectively isolates foreground objects, a behavior further illustrated by the qualitative results in Appendix~\ref{app:qualitativePruningResults}.
In comparison SViT (shown in red) requires the first layers to be dense at the cost of a high feature reduction in the deepest layers.
This results in an insufficient patch density of 4\% at layer 12 given the 40\% PKR constraint.

To further quantify the sparsity gains, we evaluate the speedup of patch pruning methods.
Therefore we measure the FLOPs of the ViT-Adapter, as well as the models throughput in frames per second (FPS).
The FPS are measured on a \textit{Nvidia RTX2080TI} GPU using batch size 1.
Table~\ref{tab:flops} shows the results measured on Youtube-VIS~2021 at 55\% PKR.
\textit{VPP} reduces the FLOPS from 88.0~G down to 59.9~G while only loosing 0.6\% in AP.
In contrast, SViT looses 4\% in AP at an equivalent sparsity.
Moreover, \textit{VPP} achieves a significant speedup compared to the dense ViT-Adapter, demonstrating superior throughput for both the backbone (25.9 FPS vs. 17.1 FPS) and the full model (8.13 FPS vs. 6.58 FPS).

\section{Conclusion}

In this work, we introduced \textit{Video Patch Pruning} (\textit{VPP}), a novel pruning framework for early patch reduction in video domains. 
\textit{VPP} significantly surpasses the sparsity limitations of conventional image-based patch pruning methods, reaching an average patch reduction of up to 60\%.
On Video Instance Segmentation tasks, the proposed method demonstrates excellent performance results with negligible loss in Average Precision.
The early pruning strategy of \textit{VPP} involves a lightweight, temporal mapping of patches to account for object motion.
Specifically, foreground-selective features are projected from the previous frame to ensure accurate background removal in the current frame.
As a result, VPP sets clear state of the art for Video-Instance Segmentation, outperforming all competitors at a 55\% of keep ratio in Average Precision, even when only keeping 40\% itself.

\clearpage

\section{Acknowledgements}

This work was supported by the MWK of Lower Saxony within Hybrint (VWZN4219) and LCIS (VWZN4704), the Deutsche Forschungsgemeinschaft (DFG) under Germany’s Excellence Strategy within the Cluster of Excellence PhoenixD (EXC2122) and Quantum Frontiers (EXC2123), the European Union under grant agreement no. 101136006 – XTREME.

{
    \small
    \bibliographystyle{ieeenat_fullname}
    \bibliography{main}
}

\appendix
\clearpage
\setcounter{page}{1}
\maketitlesupplementary

\section{Training setting}
\label{app:train_settings}

This section provides a detailed specification of the training configurations employed in our Video Patch Pruning (\textit{VPP}) framework.

For the Video Instance Segmentation (\textit{VIS}) task we use ROVIS~\cite{zhan2022rovis} as tracking method.
It requires only two temporally corresponding frames during training, significantly lowering the memory consumption.
In all our experiments we stick to the training settings, reported by ROVIS.
Therefore we first train the dense model for 6 epochs using AdamW as optimizer with a weight decay of $0.05$ and a learning rate of $2.5\times10^{-5}$, with a lr-decay of $0.1$ after $4$ epochs.
For initialization, we use the model weights, pretrained on COCO dataset.
In the second step, the patch pruning method is applied to the dense model using the standard training procedure. For consistency, this same protocol is maintained across all evaluated pruning methods to ensure a fair comparison.

\textit{VPP}-specific hyperparameters are defined in Tab.~\ref{tab:hyperparameters_vpp}.
We utilize layer $6$ from the previous frame for mapping.
According to Fig.~\ref{fig:fgs}, $x_6$ shows almost maximal \textit{Foreground Selectivity (FGS)} while remaining comparably dense, as shown in Fig.~\ref{fig:img_vs_video_kr}.
Although deeper layers marginally improve \textit{FGS}, they provide fewer features for reference, preventing highly sparse regions from having adequate background representatives for mapping.
Moreover, we set the top-k selection of Map-SM module to $0.7$ and $0.6$ for Patch Keep Ratio (PKR) 55\% and 40\%, respectively.
According to Dynamic-Vit~\cite{tang2023dynamictokenpruningplain} and TPS~\cite{wei2023tps}, we define the target keep ratio $\kappa_l$ for each pruning stage as geometric sequence $[\rho, \rho^2, \rho^3]$.
We set $\rho$ to the value, where the goal PKR of 55\% and 40\% is reached.

\begin{table}[h]
    \centering
    \small
    \begin{tabular}{cc}
        \toprule
         name hyperparameter & value\\
         \midrule
         pos idx Map-SM & 1 \\
         ref-layer Map-SM & 6 \\
         pos idx SM     & 3,6,9 \\
         $\tau$         & 10 \\
         scale $\mathcal{L}^{Map}_{sp}$ & 10 \\
         scale $\mathcal{L}^{SM}_{sp}$  & 40 \\
         $\kappa_{init}$ & 0.5 \\
         \bottomrule
    \end{tabular}
    \vspace{-5pt}
    \caption{Hyperparameters used in Video Patch Pruning}
    \vspace{-0pt}
    \label{tab:hyperparameters_vpp}
\end{table}

In addition, we investigate the influence of the initial pruning threshold $\kappa_{init}$ in Table~\ref{tab:hp_kappa_init}. 
These experiments were conducted using ViT-Adapter Tiny on the YouTube-VIS 2021 dataset, optimized towards the goal-PKR of 55\%.
As defined in Eq.~\ref{eq:l_sp_map}, $\kappa_{init}$ specifies the intended mean patch probability $p^t \in \{0, 1\}$ to be reached by the \textit{Map-SM} during training.
Given that \textit{Map-SM} utilizes a top-k selection mechanism, the training objective $\kappa_{init}$ does not represent the sparsity level, but rather ensures that the estimated patch probabilities $p^t$ remain within a stable, non-saturated range.
The results show $\kappa_{init}=0.5$ performs best in our setting, reaching an AP score of $42.3$, which we therefore adopt as our default configuration.

\begin{table}[]
    \centering
    \begin{tabular}{ccc}
        \toprule
         $\kappa_{init}$ & PKR (\%) & AP$\uparrow$ \\
         \midrule
         0.3 & 51.4 & 41.8 \\
         \textbf{0.5} & \textbf{52.7} & \textbf{42.3} \\
         0.7 & 54.1 & 39.5 \\
         \bottomrule
    \end{tabular}
    \caption{Evaluation of model performance across different settings for $\kappa_{init}$. 
    ViT-Adapter Tiny is trained on Youtube-VIS 2021 at a goal-PKR of 55\%.
    $\kappa_{init}=0.5$ shows the best performance.}
    \label{tab:hp_kappa_init}
\end{table}

\section{Gumble Softmax function}
\label{sec:gumbleSm}

In this section we define the Gumble-Softmax function\cite{jang2017gumbelsoftmax}, used in our proposed Map-SM module.
For a set of class probabilities $\pi_{1}, \dots, \pi_{k}$ the soft Gumbel-probability is defined as
\begin{equation}
    \sigma^{soft}_{gumbel}(\pi_i, \tau) = \frac{exp(\frac{log(\pi_i)\cdot\tau + g_i}{\tau})}{\sum^k_{j=1}exp(\frac{log(\pi_j)\cdot\tau + g_j}{\tau})},
\end{equation}
where $g_i$ denotes Gumbel-distributed noise. 
In this formulation temperature $\tau$ scales the Gumbel-distributed noise $g_i$, such that higher values minimize its stochastic influence on the resulting soft probabilities.
Consequently, sampling the remaining patches with a higher $\tau$, results in a reduced activation of background patches, limiting the reference features required for consistent background mapping.
The addition of Gumbel-distributed noise $g_i$ serves as a reparameterization trick, establishing a differentiable approximation of the discrete $argmax$ operation. 
This continuous relaxation enables the model to optimize categorical selections through standard backpropagation during the training phase.
To bridge the gap between continuous probabilities, represented by $\sigma^{soft}_{gumbel}$, and discrete selections, the final mask is derived through a one-hot operation as straight-through estimator:
\begin{equation}
    \sigma^{hard}_{gumbel}(\pi_i, \tau) = \text{one\_hot}(\sigma^{soft}_{gumbel}(\pi_i, \tau)).
\end{equation}

In contrast to standard inference procedures~\cite{Rao_2021_Neurips, Liu_2024_WACV} that employ a deterministic $argmax$ to eliminate stochasticity, we retain Gumbel noise during inference for background activation. 
This noisy activation is essential for both identifying new objects in the video sequence and ensuring the background being sparsely activated for mapping.

\begin{figure*}[ht]
    \setlength{\tabcolsep}{1pt}
    \begin{tabular}{cccc}
        \includegraphics[width=0.24\linewidth]{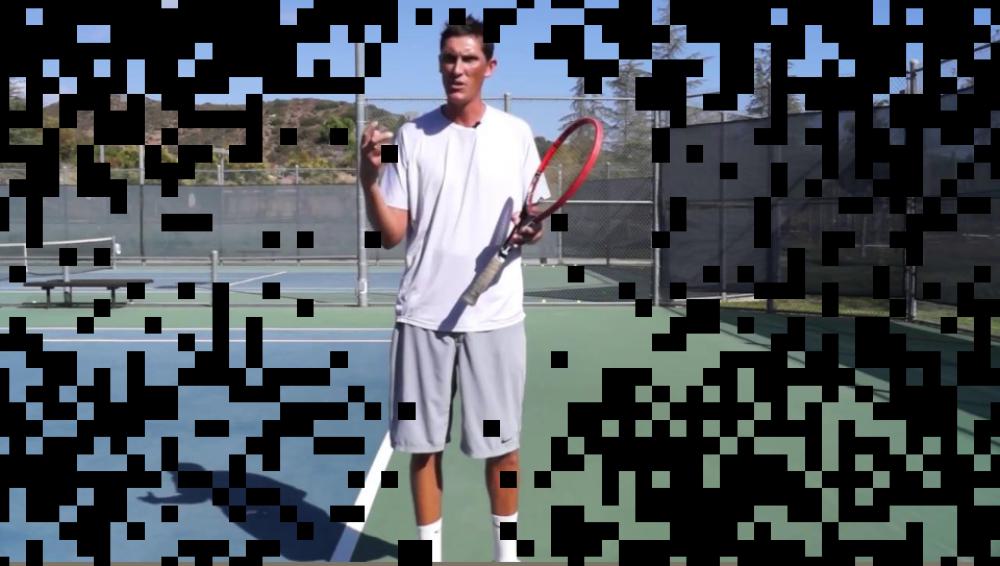} &
        \includegraphics[width=0.24\linewidth]{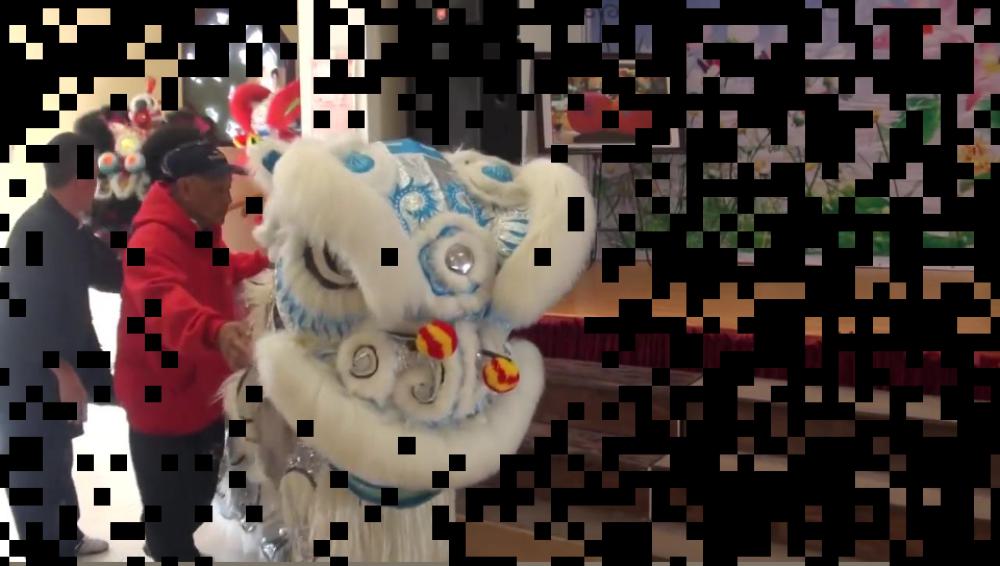} &
        \includegraphics[width=0.24\linewidth]{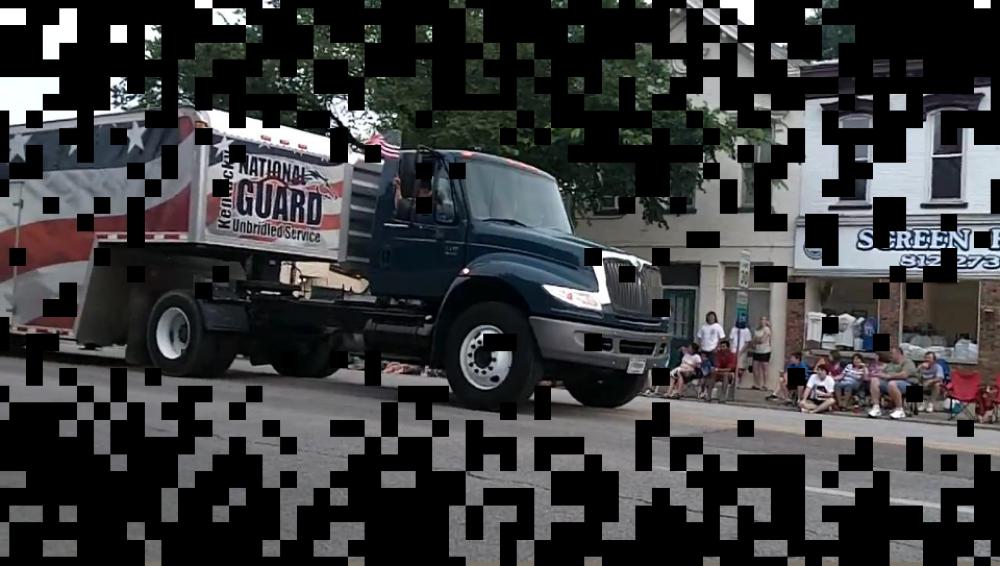} &
        \includegraphics[width=0.24\linewidth]{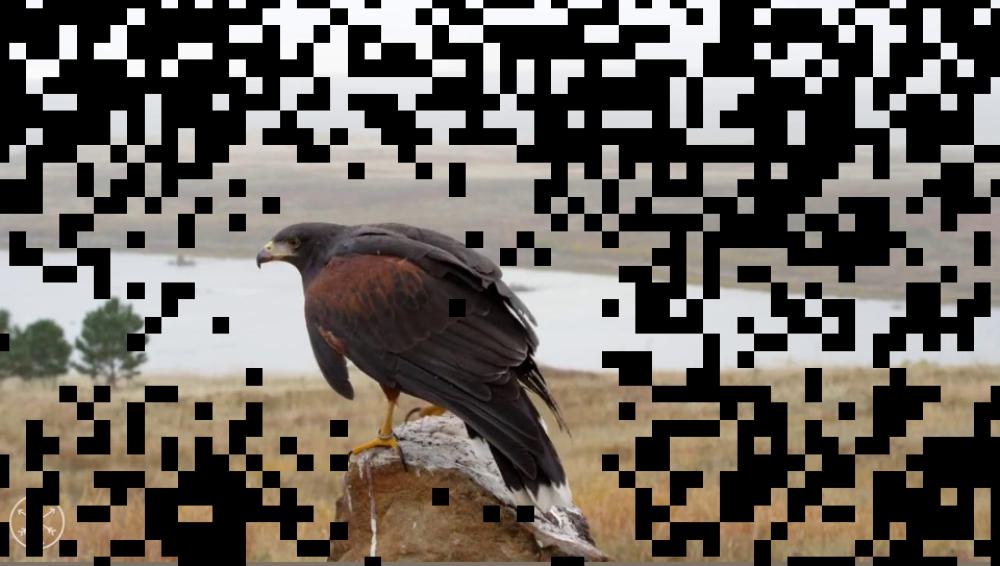} \\

        \includegraphics[width=0.24\linewidth]{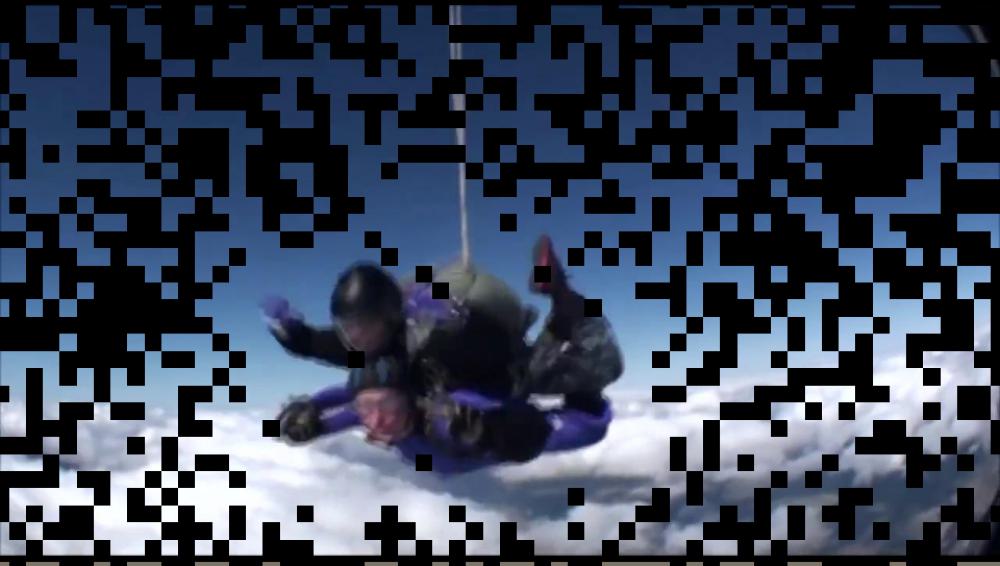} &
        \includegraphics[width=0.24\linewidth]{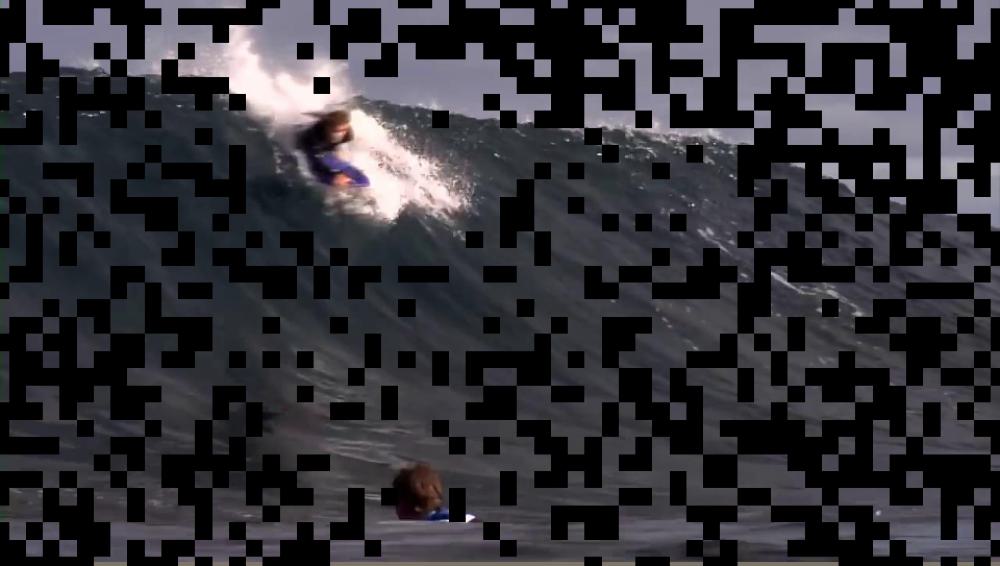} &
        \includegraphics[width=0.24\linewidth]{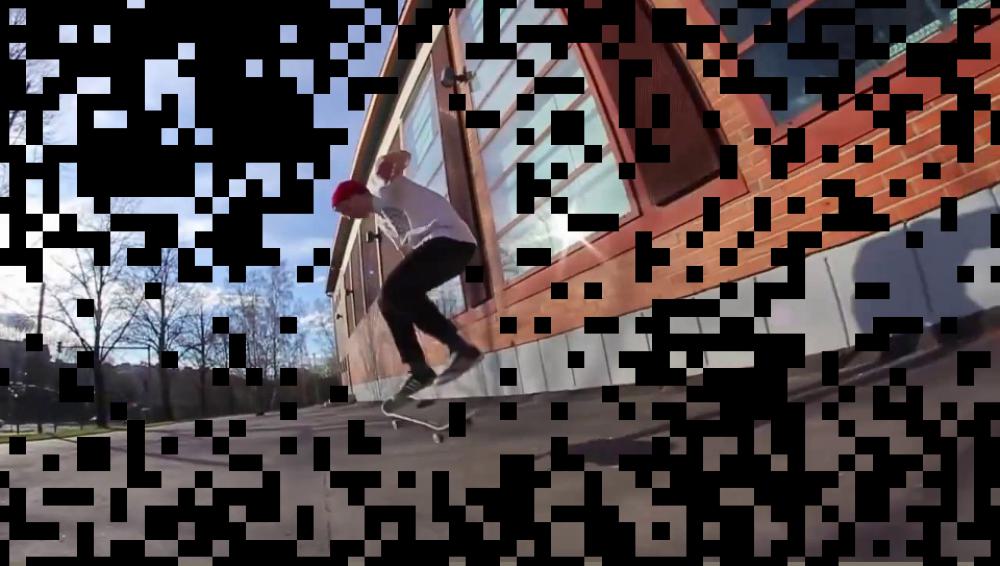} &
        \includegraphics[width=0.24\linewidth]{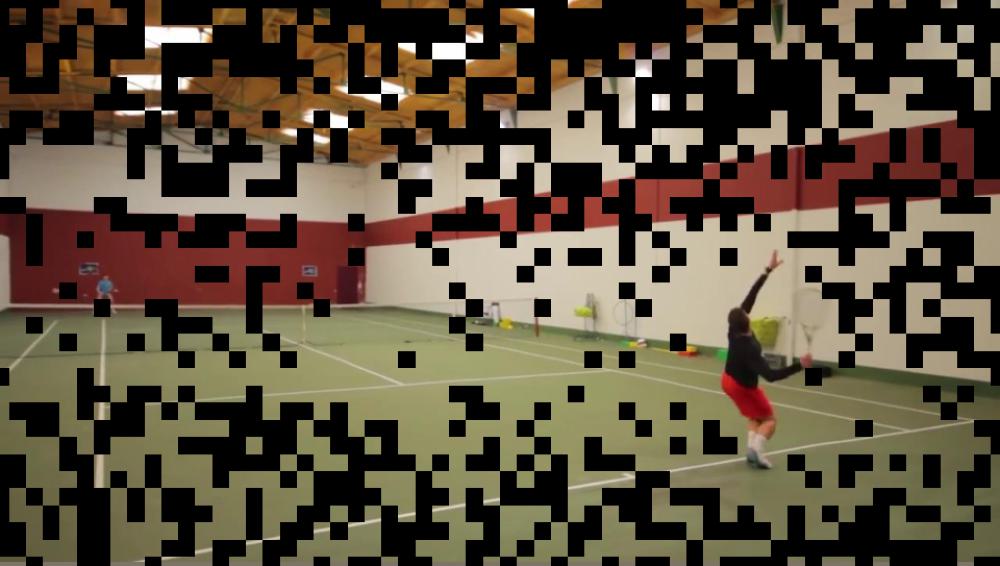} \\

    \end{tabular}
    \caption{Examples for the initial pruning mask $M_1^t$, generated by Mapping Selective Module (Map-SM).
    This mask removes 40\% of patches after the first transformer block.  
    All masks demonstrate that patches from foreground objects are sampled more densely, while background regions are effectively pruned during the initial mask generation.
    }
    \label{fig:init_mask_examples}
\end{figure*}

\section{Foreground Selectivity}
\label{supl:foreground_selectivity}

In this section, we provide further details on the experiments to Foreground Selectivity (FGS) from Sec.~\ref{sec:from_img2vid_pp}. 
This experiment evaluates the capacity of dense, intermediate features to accurately identify foreground patches. 
Such a property is essential for patch pruning, as during pruning the focus should be shifted towards the foreground patches that belong to object instances.
Background patches contain less relevant information and can even negatively effect the prediction~\cite{Tian_2025_PAFGO}.
Specifically, in instance segmentation tasks, background patches do not belong to target classes and should not be segmented.
Consequently, the ability to bypass these non-informative regions allows the model to reduce overhead.

We evaluate patch relevance by adding binary classification heads to each intermediate layer of the frozen model. 
This head is optimized to differentiate between patches that belong to object related foreground or background.
The ground truth label is extracted from the ground-truth segmentation mask.
To mitigate the effects of class imbalance, such as the disproportionate background-to-foreground ratio observed in the YouTube-VIS datasets, we utilize weighted CE loss~\cite{9319440} for optimization. 
Specifically, we monitor the ratio of foreground patches through a running mean, denoted as $\overline{r}_{fg}$. 
To balance the loss, we weight the background and foreground class for the classification loss by $w_{bg}=\overline{r}_{fg}$ and $w_{fg}=1-\overline{r}_{fg}$, respectively.
During training we freeze the whole model, except the intermediate classification heads.
The classifier are trained for $2$ epochs, following the settings form Sec.~\ref{app:train_settings}. 
The learning rate is decayed by a factor of $0.1$ after the initial epoch.

For the patch pruning task, specifically in instance segmentation tasks, the primary objective is to ensure that all foreground patches are retained during propagation.
Therefore we define Foreground Selectivity (FGS) as the binary classification accuracy between foreground and background patches within feature $x_i$, given layer index $i$.

The results in Fig.~\ref{fig:fgs} of the main paper show that the features almost linearly improve in depth in terms of Foreground Selectivity. 
Thereby, 0.5 indicates the lower bound classification, of a random patch sampling.
At index 0, an FGS score of 0.6 indicates that while low-level features possess some discriminative information, it is insufficient for robust foreground selection.
The FGS-score almost linearly increases in depth until layer 9.
Overall, these results indicate that robust foreground patch selection is only achievable in the deepest layers. 
Common patch pruning methods such as DynamicViT \cite{Rao_2021_Neurips}, EViT \cite{liang2022evit}, TPS~\cite{Wei_2023_CVPR} and DPS \cite{Tang_2022_CVPR} typically reduce features linearly at predefined intervals of layers 3, 6 and 9.
This aligns with our experimental results, suggesting that reliable foreground-background separation is a property of mature, deep-layer features, making early-stage pruning at layers 3 and 6 potentially suboptimal.

\section{Intersection over Instance}
\label{app:ioi}

\begin{table}[t]
    \centering
    \resizebox{1.0\linewidth}{!}{
    \begin{tabular}{ccc|c|ccc}
        \toprule
         PKR (\%) & \makecell{dataset \\ Youtube-VIS} & method & $IoI$ & $IoI_{\underline{S}}$ & $IoI_{\underline{M}}$ & $IoI_{\underline{L}}$ \\
         
         \midrule
         \multirow{4}{*}{55\%} 
         & \multirow{2}{*}{2021} 
         & SViT & 73.7\% & 77.2\% & 68.8\% & 65.8\% \\ 
         & & VPP (ours)  & \textbf{82.3\%} & \textbf{87.3\%} & \textbf{76.2\%} & \textbf{72.2\%} \\ 

         \cdashline{2-7}
         \addlinespace[1pt]
         
         & \multirow{2}{*}{2019} 
         & SViT & 63.1\% & 67.9\% & 61.0\% & 58.5\%\\ 
         & & VPP (ours)  & \textbf{77.3\%} & \textbf{84.4\%} & \textbf{75.0\%} & \textbf{71.4\%} \\
         
         \cdashline{1-7}
         \addlinespace[1pt]
         
         \multirow{4}{*}{40\%} 
         & \multirow{2}{*}{2021} 
         & SViT & 62.7\% & 67.5\% & 56.0\% & 52.1\% \\ 
         & & VPP (ours) & \textbf{73.4\%} & \textbf{79.5\% }& \textbf{65.9\%} & \textbf{60.7\%} \\ 

         \cdashline{2-7}
         \addlinespace[1pt]
         
         & \multirow{2}{*}{2019} 
         & SViT & 54.9\% & 60.6\% & 52.4\% & 49.4\% \\ 
         & & VPP (ours)  & \textbf{66.7\%} & \textbf{75.8\%} & \textbf{63.7\%} & \textbf{58.8\%} \\
         
         \bottomrule
    \end{tabular}
    \vspace{-10pt}
    }
    \caption{Intersection over Instance (IoI) over datasets Youtube-VIS 2019 and 2021.
    VPP shows superior patch coverage every setting, improving the IoI scores by $>10\%$ compared to SViT. 
    }
    \vspace{-0pt}
    \label{tab:ioi_full}
\end{table}

In the experiments to \textit{Intersection over Instance} (\textit{IoI}) in the main paper, we analyzed how many foreground patches are actually activated during inference.

In instance segmentation tasks, background patches must not be segmented and therefore are less relevant for te segmentation task. 
We aim to maximize the activation of patches belonging to an instance, while reducing the computed patches to a certain point.
Therefore we see \textit{IoI} as valuable metric, especially for instance segmentation tasks.

Tab.~\ref{tab:ioi_full} shows the IoI-scores for for Youtube-VIS 2019 and 2021, given a sparsity level of $55\%$ and $40\%$ PKR.
In addition, we report IoI-scores across three sized categories based on spatial coverage: small (\underline{S}), representing instances with a spatial size $\leq10\%$; medium (\underline{M}), ranging from $10\%$ to $20\%$; and large (\underline{L}), for instances exceeding $20\%$.
The results show, \textit{VPP} has a higher IoI-score on all settings, compared to SVIT. 
Notably, \textit{VPP} allocates a higher patch density to smaller objects.
This strategy is reasonable due to the fact that larger instances typically exhibit higher spatial redundancy~\cite{choudhury2025accelvita}, allowing for accurate predictions with a relatively lower patch sampling rate.

\begin{figure}[t]
    \setlength{\tabcolsep}{1pt}
    \centering
    \begin{tabular}{ccccc} 
        & Raw Image & SViT & VPP (ours) & \multirow{2}{*}{\includegraphics[width=0.08\linewidth, height=3.6cm]{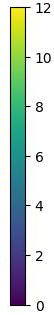}} \\
    
        (a) & \includegraphics[width=0.28\linewidth, valign=m]{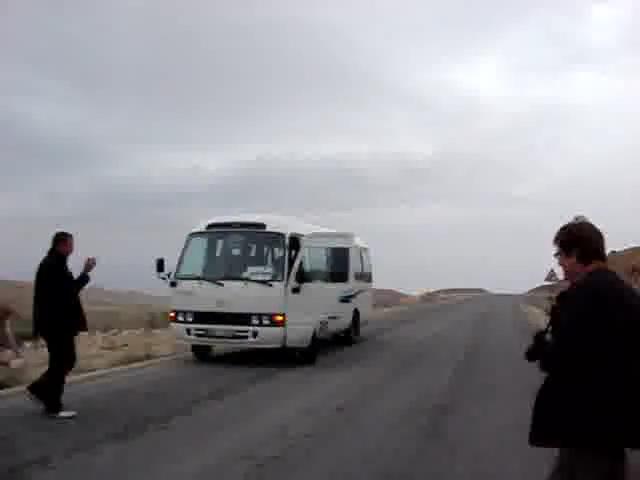} &
        \includegraphics[width=0.28\linewidth, valign=m]{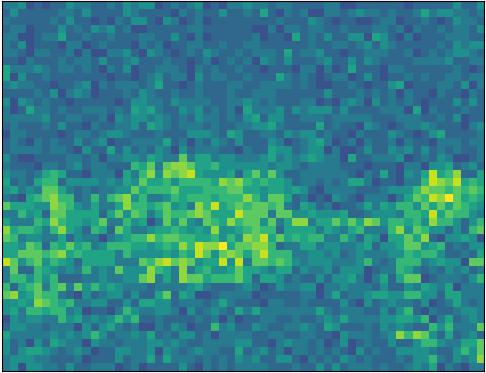} &
        \includegraphics[width=0.28\linewidth, valign=m]{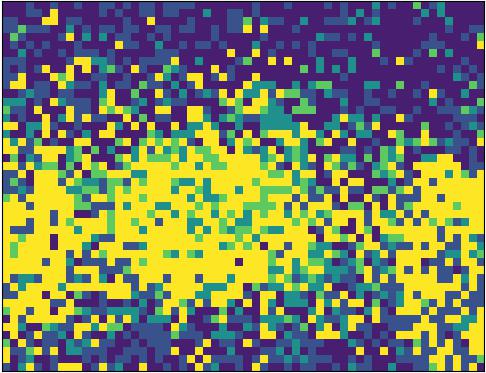}\\

        (b) & \includegraphics[width=0.28\linewidth, valign=m]{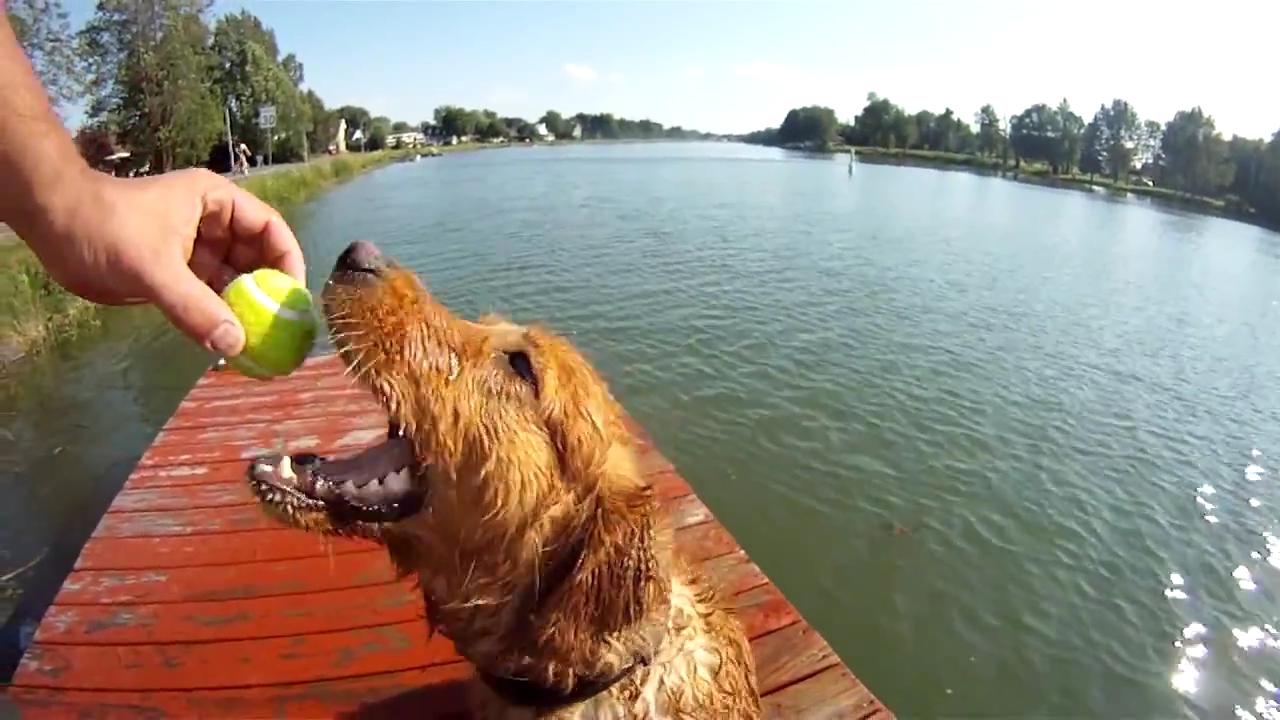} &
        \includegraphics[width=0.28\linewidth, valign=m]{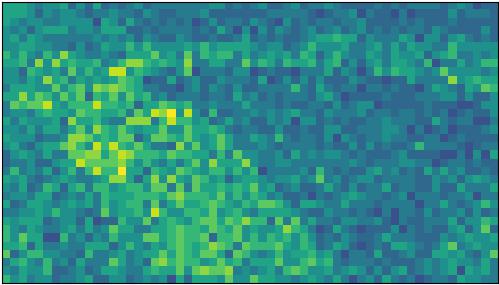} &
        \includegraphics[width=0.28\linewidth, valign=m]{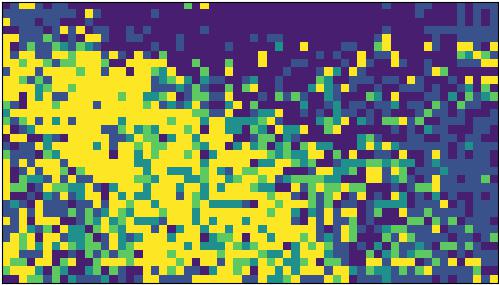}\\
    \end{tabular}
    \vspace{0pt}
    \caption{Comparison of token usage ($55\%$ PKR) for image-based (\textit{SViT}) and video-based (\textit{VPP}) Pruning.
    Unlike image-based patch pruning, \textit{VPP} computes foreground features almost fully dense, while sparsely activating background patches for context preservation.
    For Video Instance Segmentation, this is essential to ensure the model focuses on the instances to be segmented, thereby minimizing redundant background computation.
    }
    \vspace{-10pt}
    \label{fig:comp_depth_map}
\end{figure}

To qualitatively demonstrate the increased focus on object-related patches in \textit{VPP}, Fig.~\ref{fig:comp_depth_map} visualizes the patch activation patterns of both \textit{SVIT} and \textit{VPP}. 
The image-based pruning method \textit{SVIT} processes the first three layers as fully dense, only introducing sparsity in deepest layers. 
Consequently, irrelevant background patches are processed unnecessarily at least five times.
In contrast, \textit{VPP} identifies and removes background patches in earlier layers, effectively shifting the computational focus toward foreground regions. 
As illustrated in Fig.~\ref{fig:comp_depth_map}, foreground objects such as the person, dog, and hand exhibit nearly dense activation patterns.

\section{Qualitative Pruning Results}
\label{app:qualitativePruningResults}

In this section we show several examples for the applied pruning mask of the proposed \textit{VPP} method.
Therefore Fig.~\ref{fig:init_mask_examples} demonstrates several examples for the initial pruning mask, generated by Map-SM and applied to the early ViT layers.
Black patches are removed in the in layer 1 and are kept deactivated in all subsequent layers to save computational costs.
The pruning masks consistently maintain high patch density on the foreground while preferentially removing background features. 
Nevertheless, \textit{VPP} retains a sparse level of background activation, which is essential to identify new object instances.
Note that \textit{VPP} does not require auxiliary loss functions to enforce the retention of foreground patches during the pruning process.
The optimization of instance segmentation losses~\cite{zhan2022rovis} naturally biases the model toward foreground features, effectively prioritizing them over the background.

\begin{figure*}
    \setlength{\tabcolsep}{1pt}
    \begin{tabular}{cccccc}
        & & \multicolumn{4}{c}{time $\rightarrow$} \\

        \multirow{2}{*}{\rotatebox{90}{Video 1}} & \rotatebox{90}{\hspace{20pt} t=1-4} &
        \includegraphics[width=0.23\linewidth]{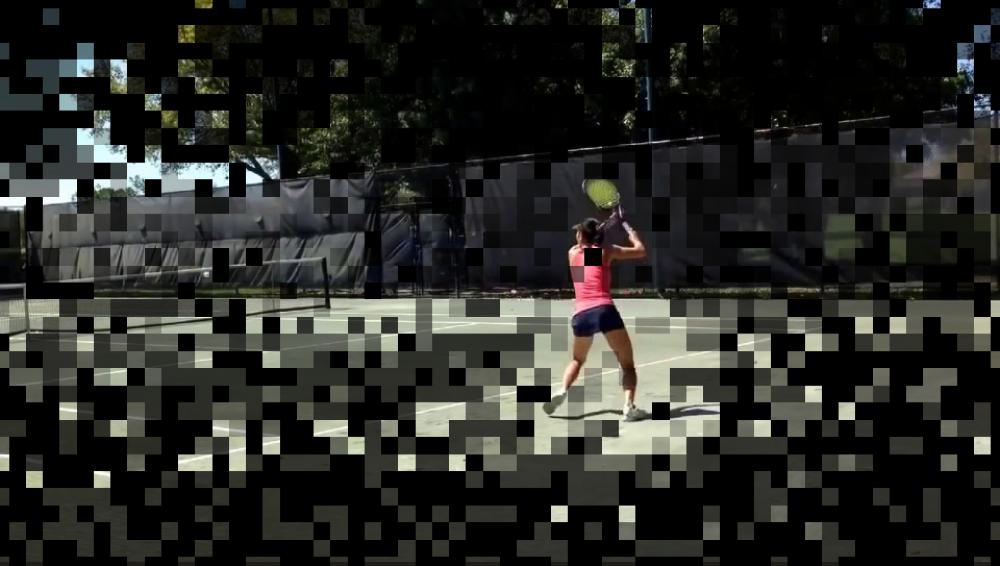} &
        \includegraphics[width=0.23\linewidth]{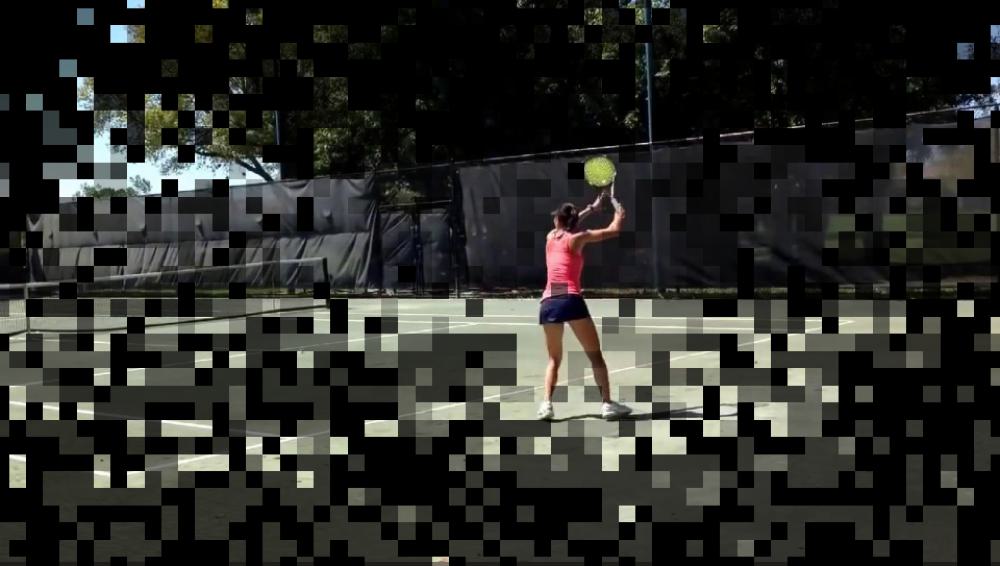} &
        \includegraphics[width=0.23\linewidth]{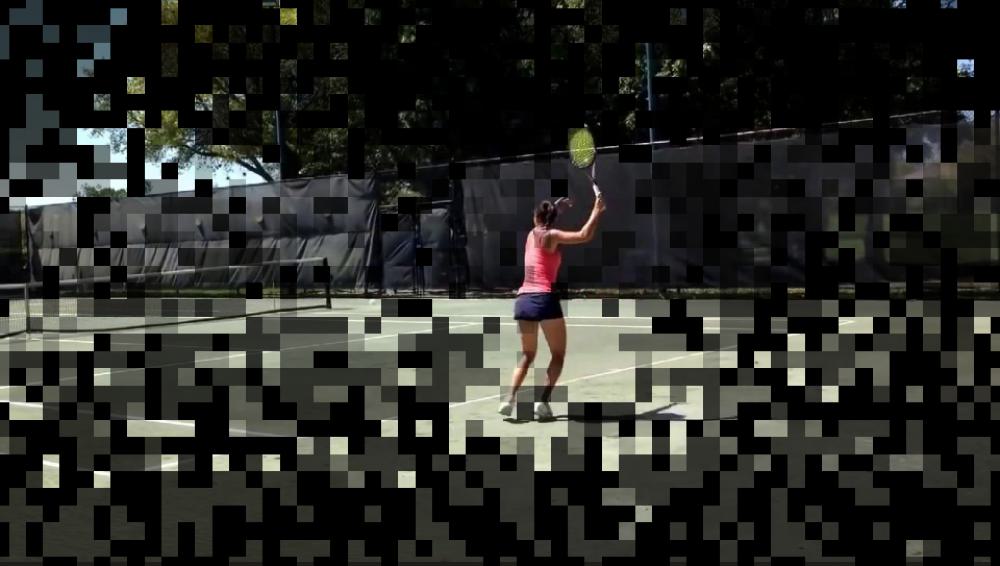} &
        \includegraphics[width=0.23\linewidth]{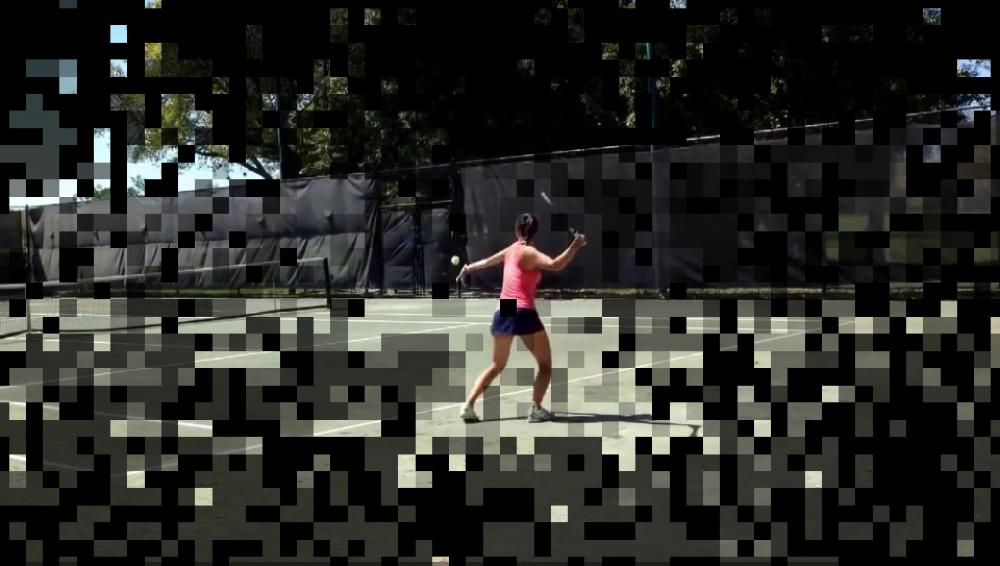} \\
        & \rotatebox{90}{\hspace{20pt}  t=5-8} &
        \includegraphics[width=0.23\linewidth]{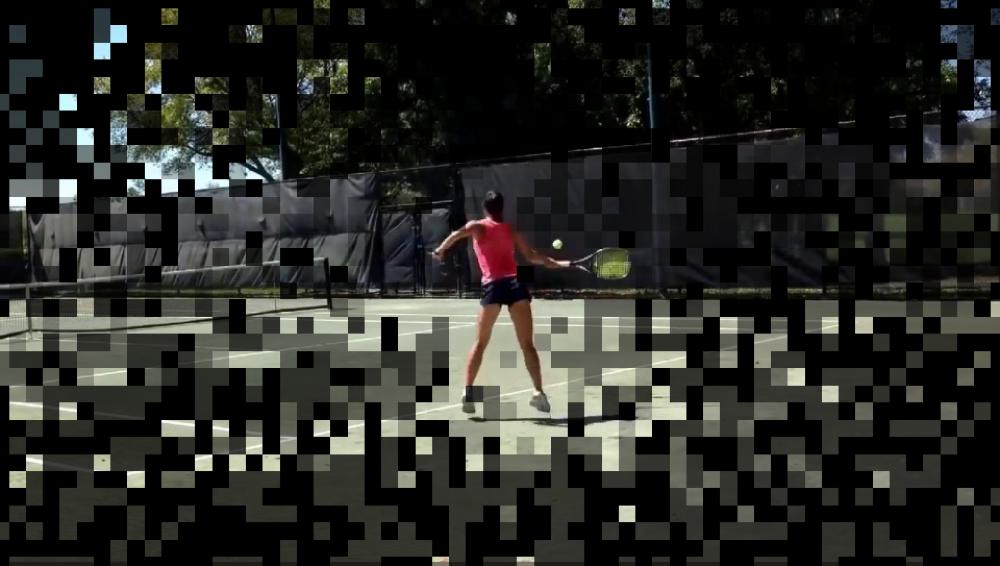} &
        \includegraphics[width=0.23\linewidth]{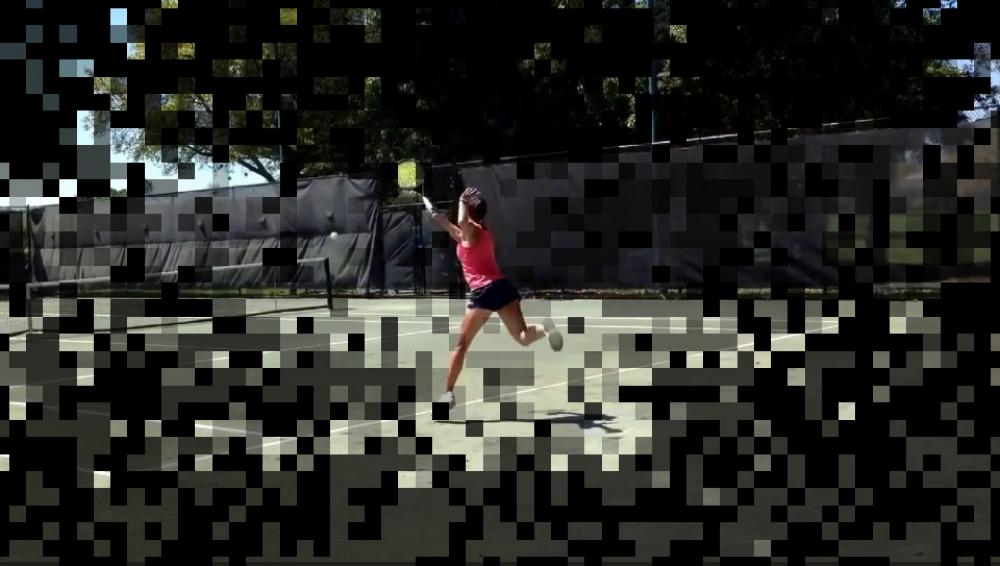} &
        \includegraphics[width=0.23\linewidth]{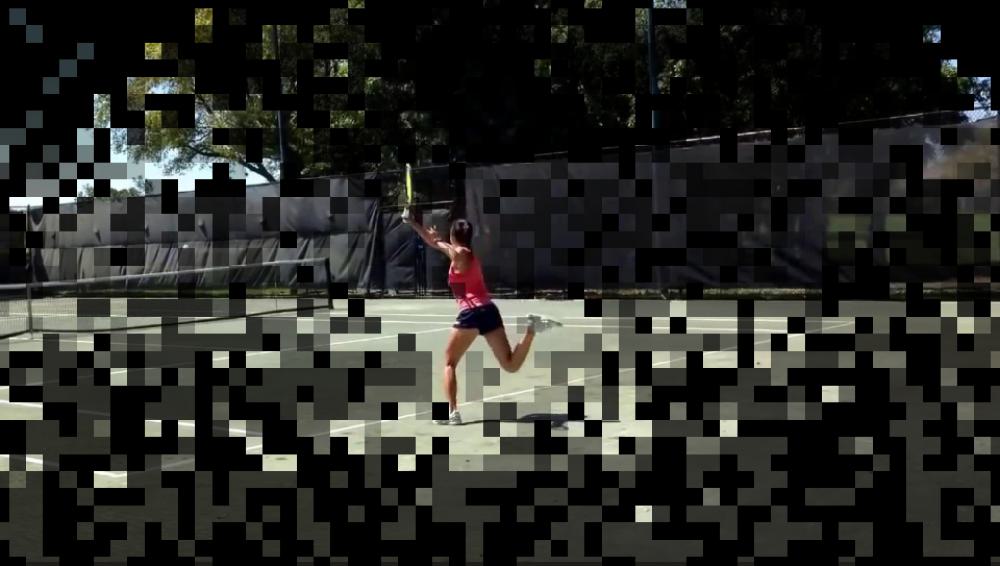} &
        \includegraphics[width=0.23\linewidth]{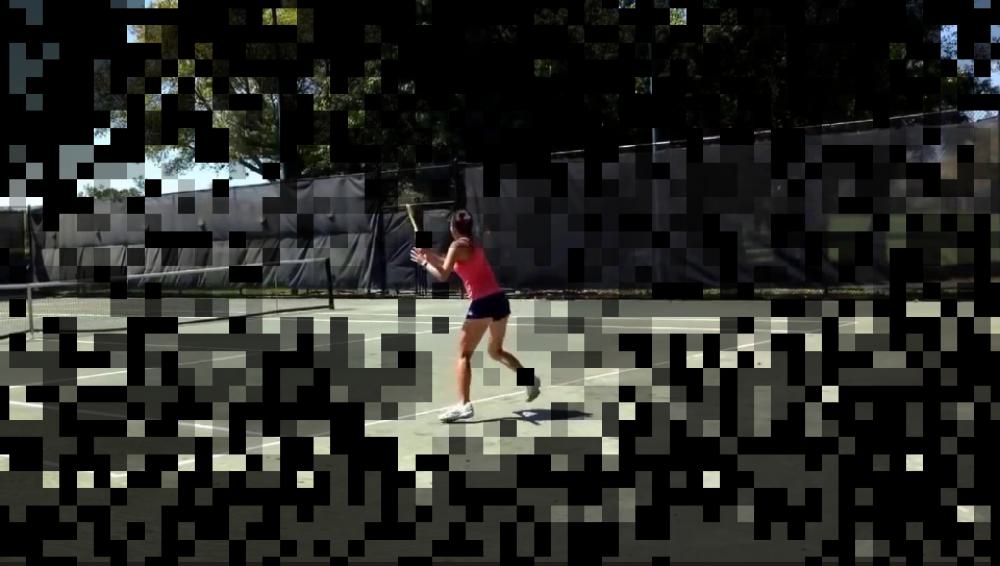} \\
        \midrule 
        
        \multirow{2}{*}{\rotatebox{90}{Video 2}} & \rotatebox{90}{\hspace{20pt} t=1-4} &
        \includegraphics[width=0.23\linewidth]{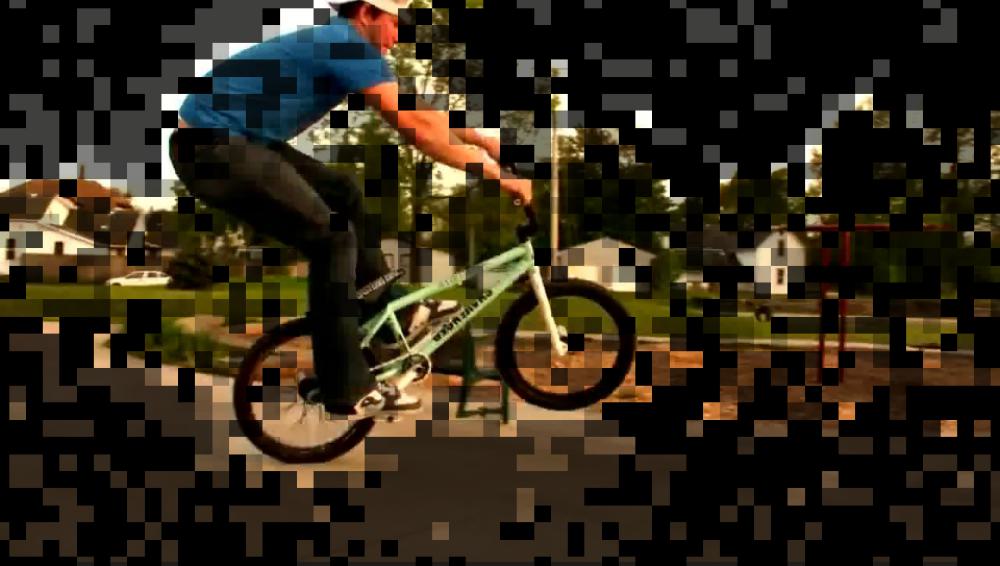} &
        \includegraphics[width=0.23\linewidth]{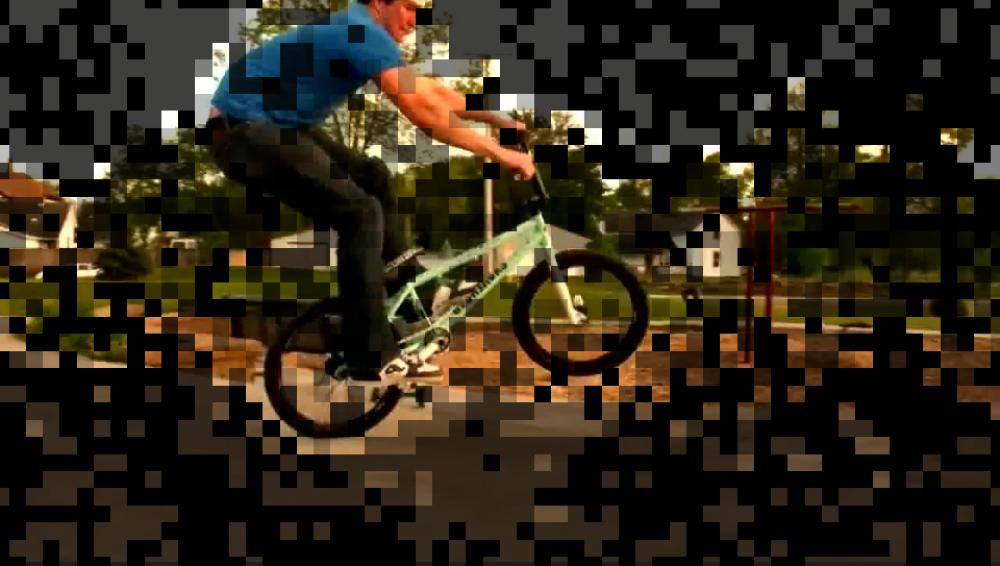} &
        \includegraphics[width=0.23\linewidth]{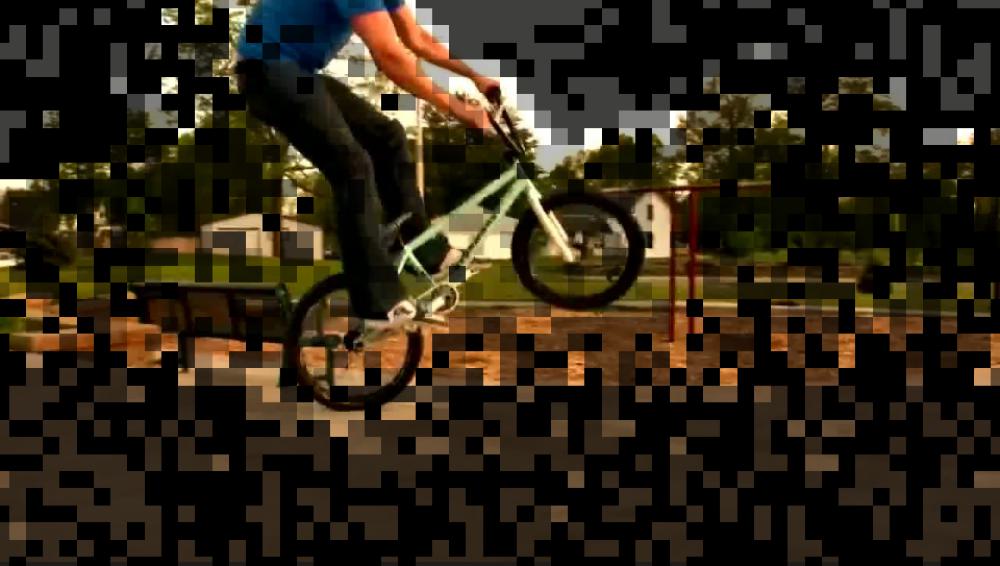} &
        \includegraphics[width=0.23\linewidth]{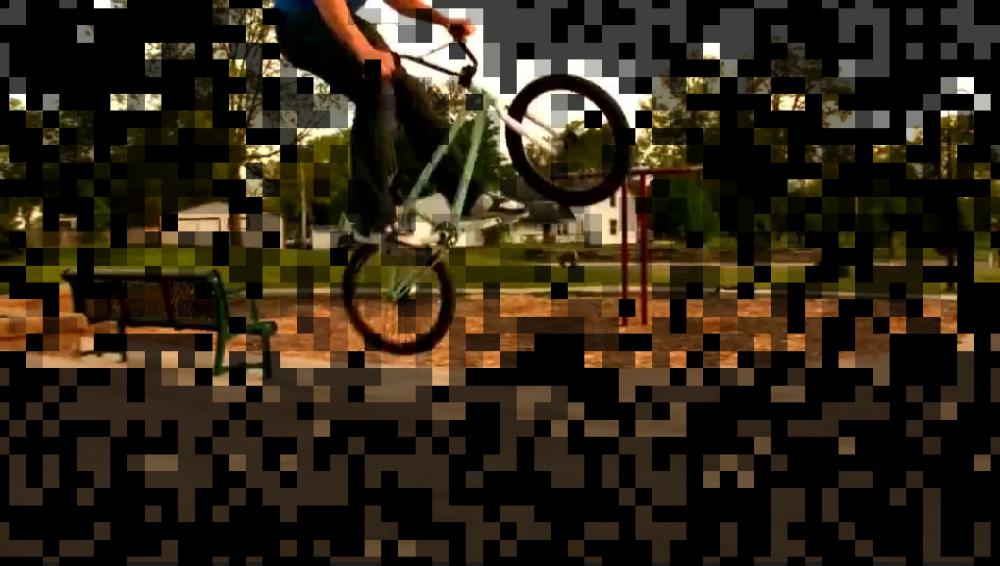} \\
        & \rotatebox{90}{\hspace{20pt}  t=5-8} &
        \includegraphics[width=0.23\linewidth]{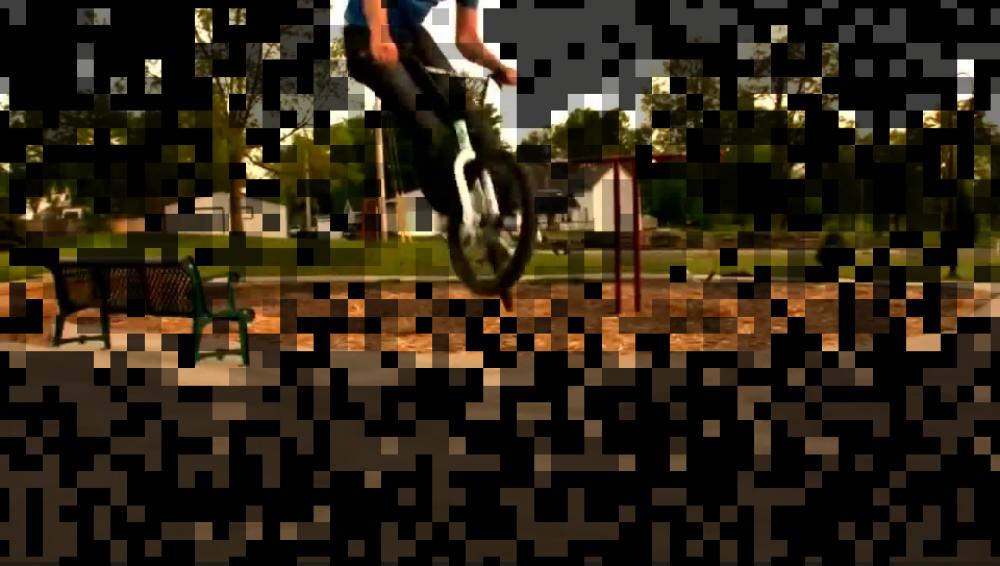} &
        \includegraphics[width=0.23\linewidth]{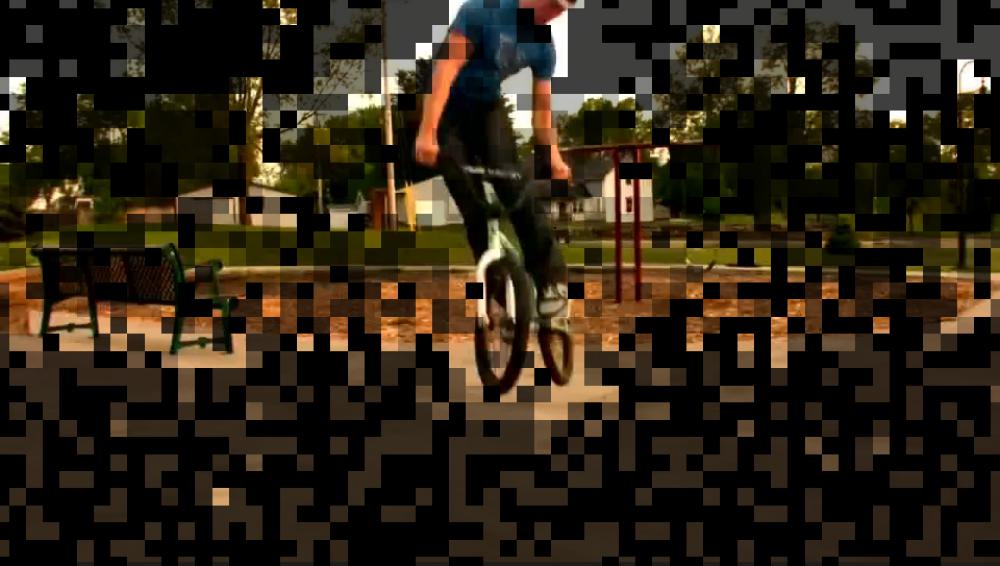} &
        \includegraphics[width=0.23\linewidth]{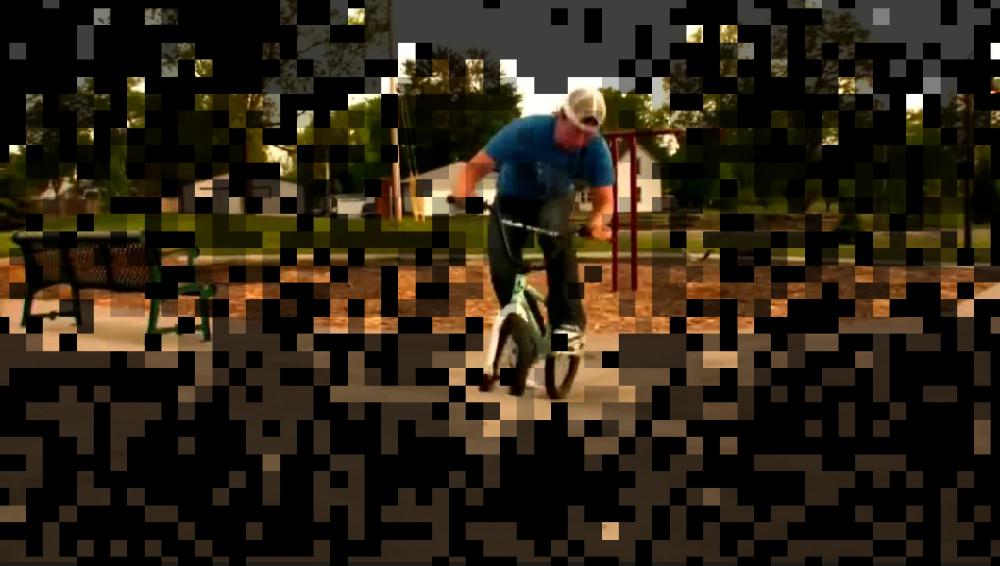} &
        \includegraphics[width=0.23\linewidth]{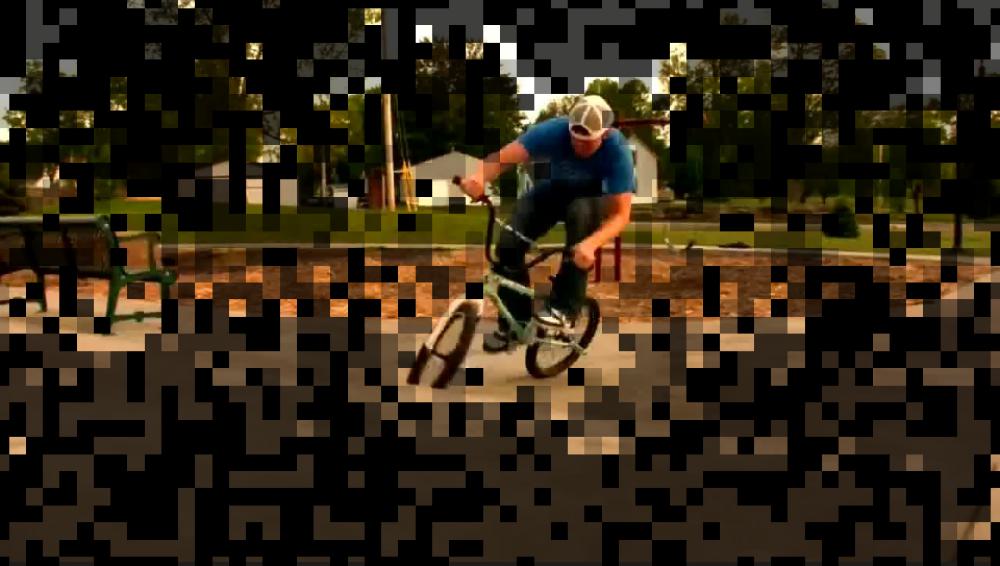} \\
        \midrule 

        \multirow{2}{*}{\rotatebox{90}{Video 3}} & \rotatebox{90}{\hspace{20pt} t=1-4} &
        \includegraphics[width=0.23\linewidth]{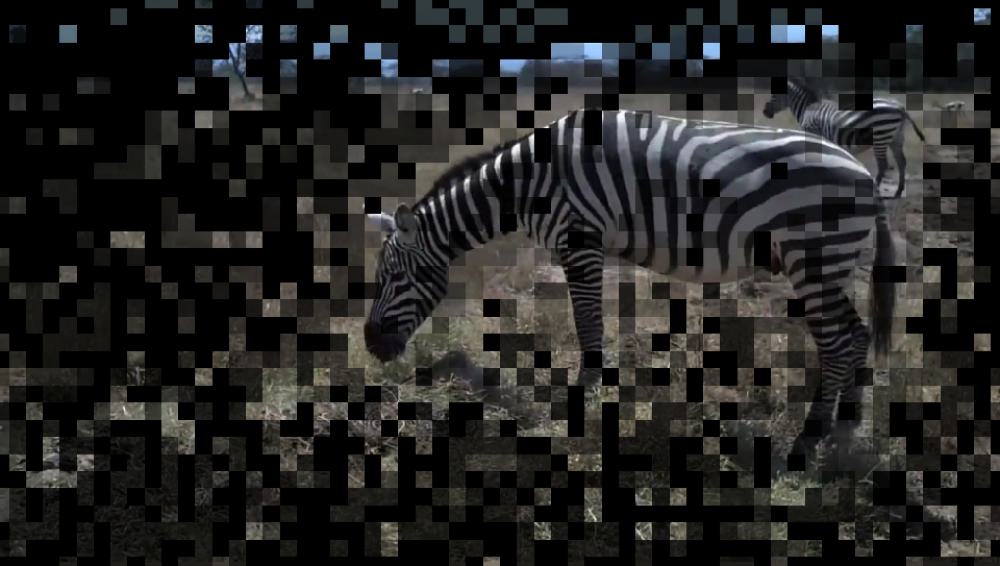} &
        \includegraphics[width=0.23\linewidth]{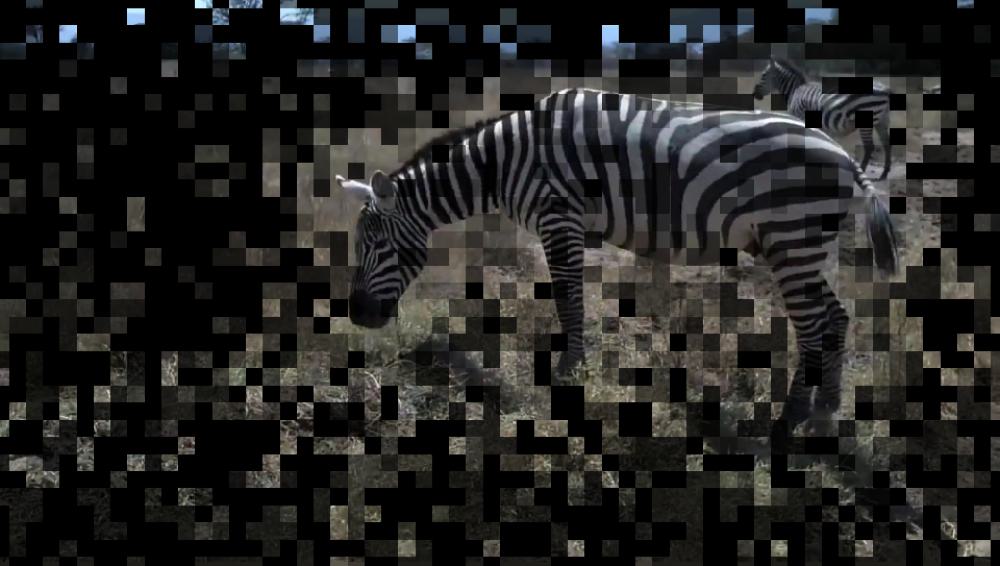} &
        \includegraphics[width=0.23\linewidth]{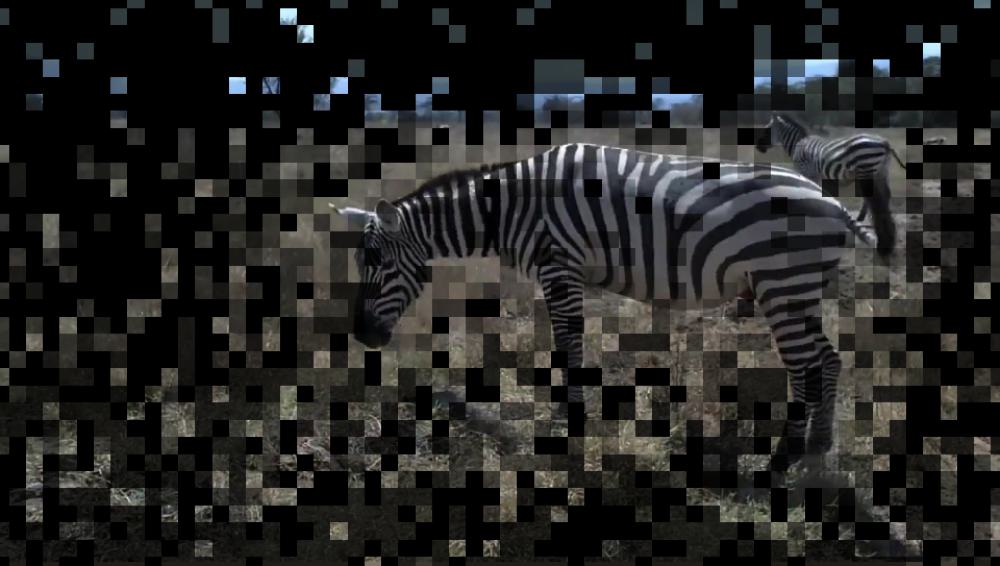} &
        \includegraphics[width=0.23\linewidth]{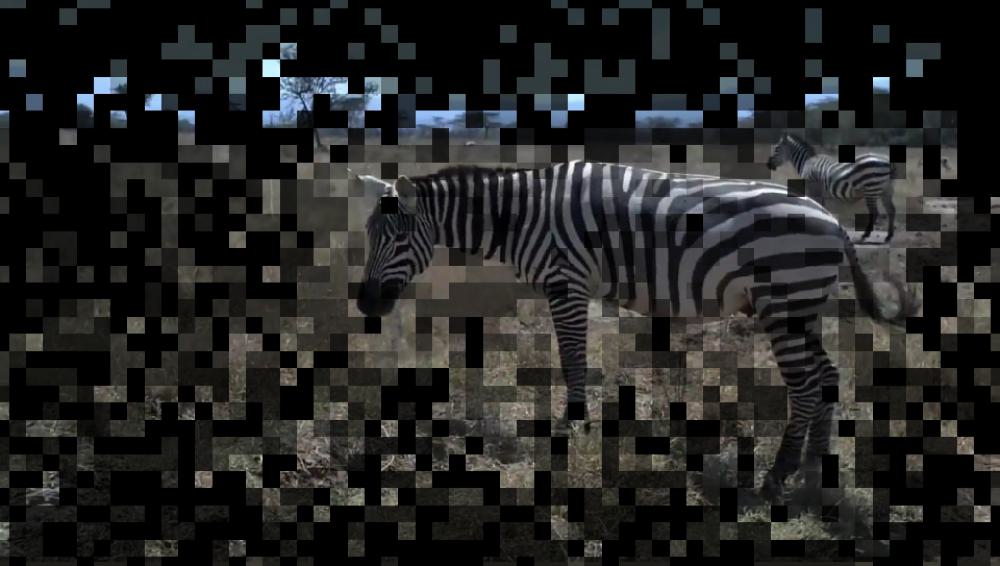} \\
        & \rotatebox{90}{\hspace{20pt}  t=5-8} &
        \includegraphics[width=0.23\linewidth]{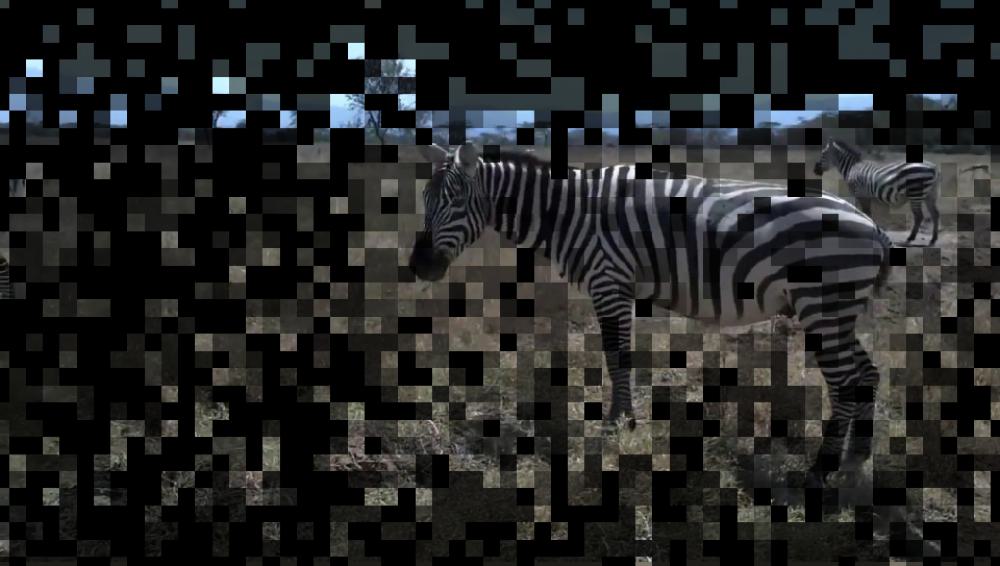} &
        \includegraphics[width=0.23\linewidth]{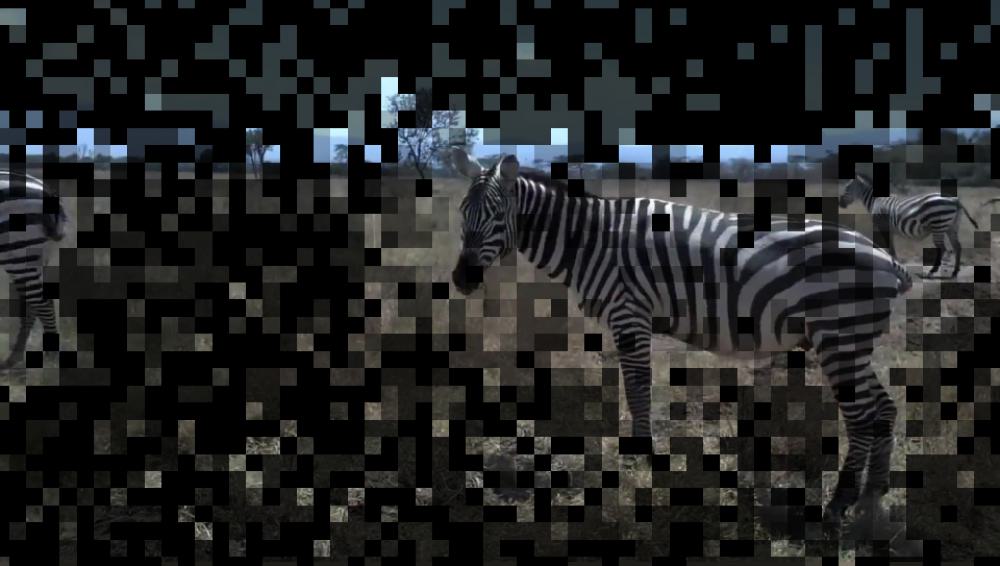} &
        \includegraphics[width=0.23\linewidth]{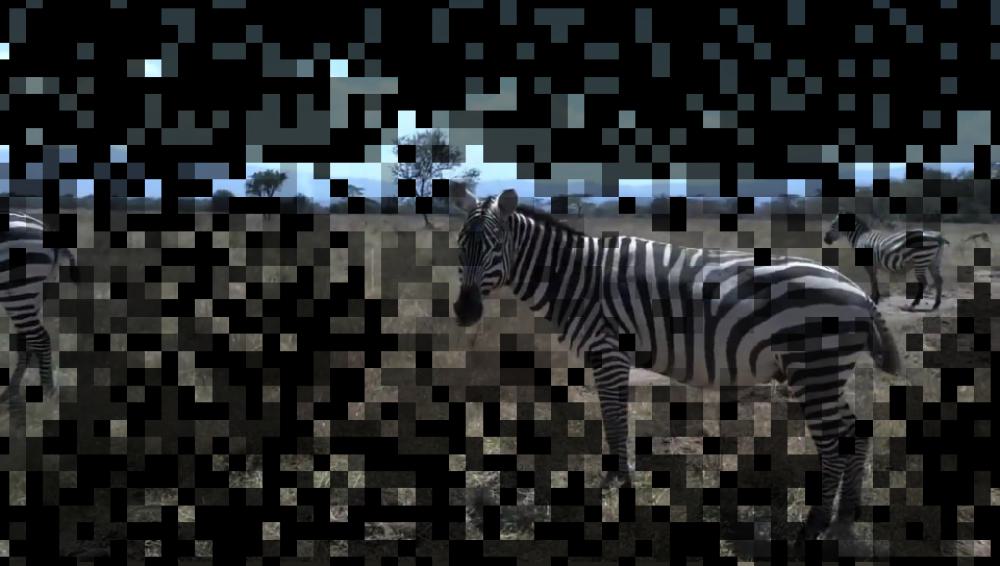} &
        \includegraphics[width=0.23\linewidth]{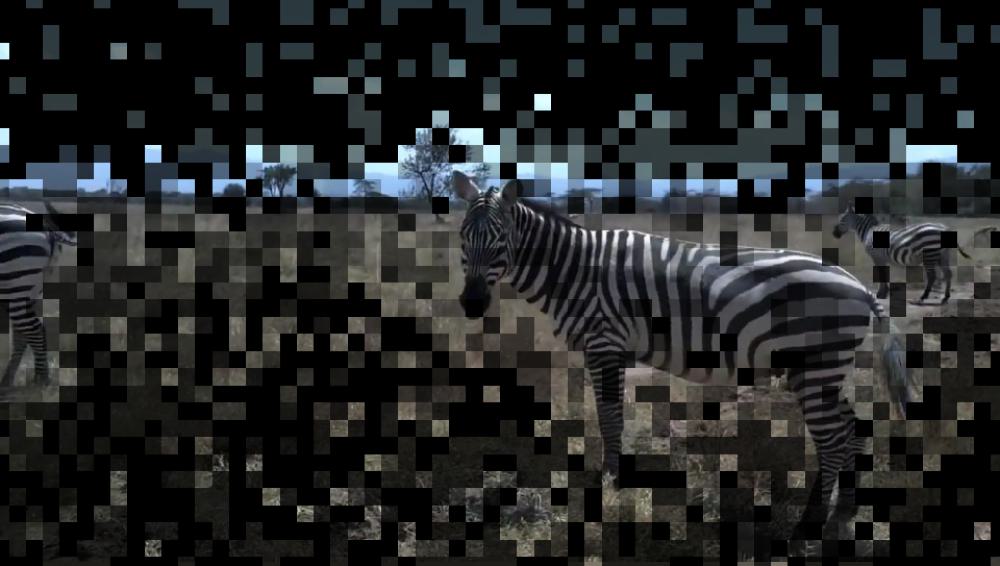} \\
        \midrule 
        
        \multirow{2}{*}{\rotatebox{90}{Video 4}} & \rotatebox{90}{\hspace{20pt} t=1-4} &
        \includegraphics[width=0.23\linewidth]{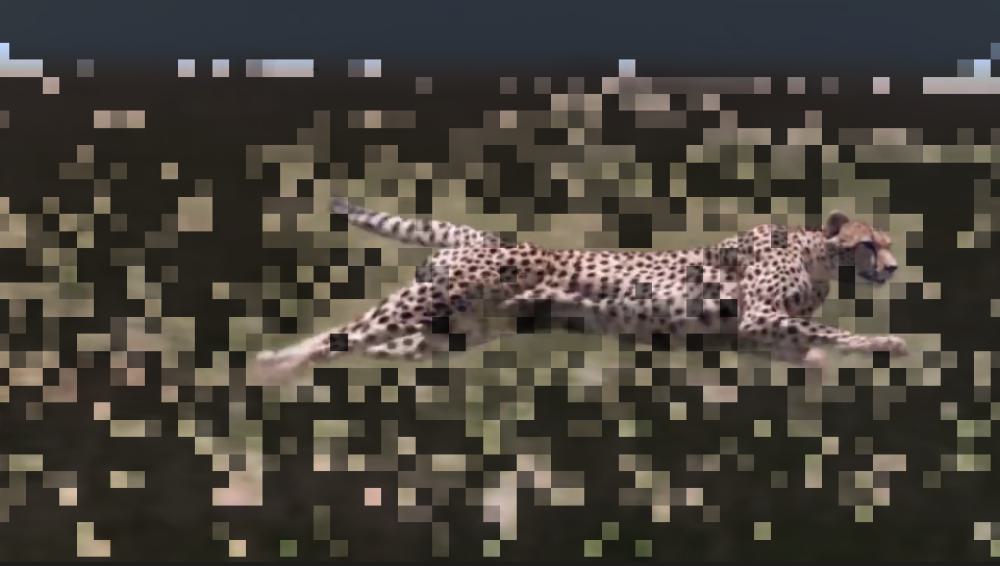} &
        \includegraphics[width=0.23\linewidth]{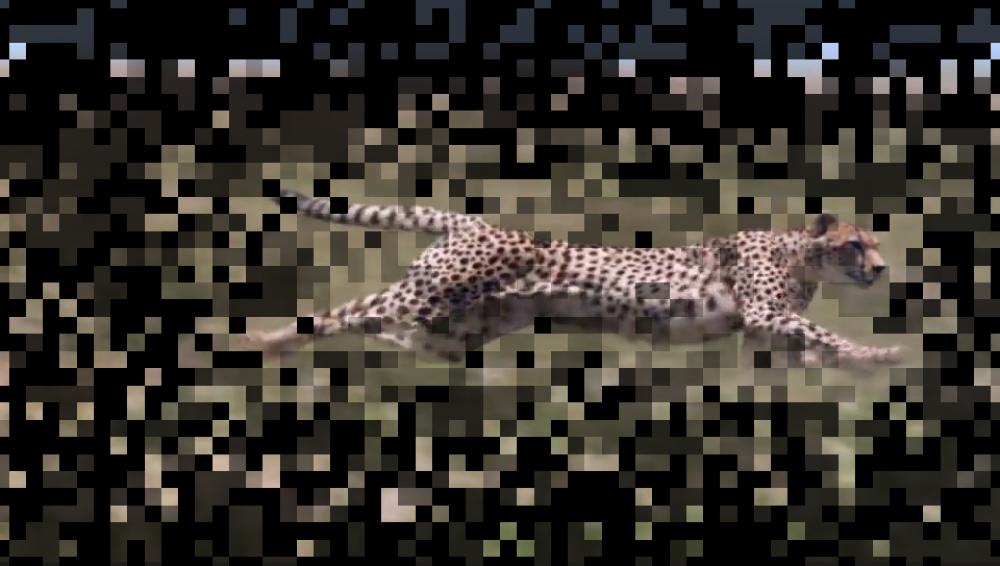} &
        \includegraphics[width=0.23\linewidth]{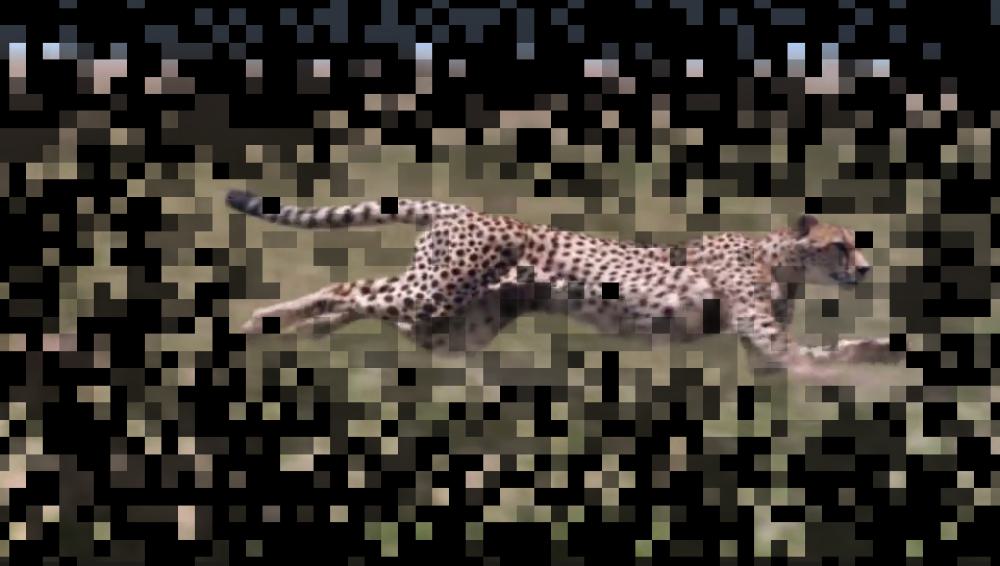} &
        \includegraphics[width=0.23\linewidth]{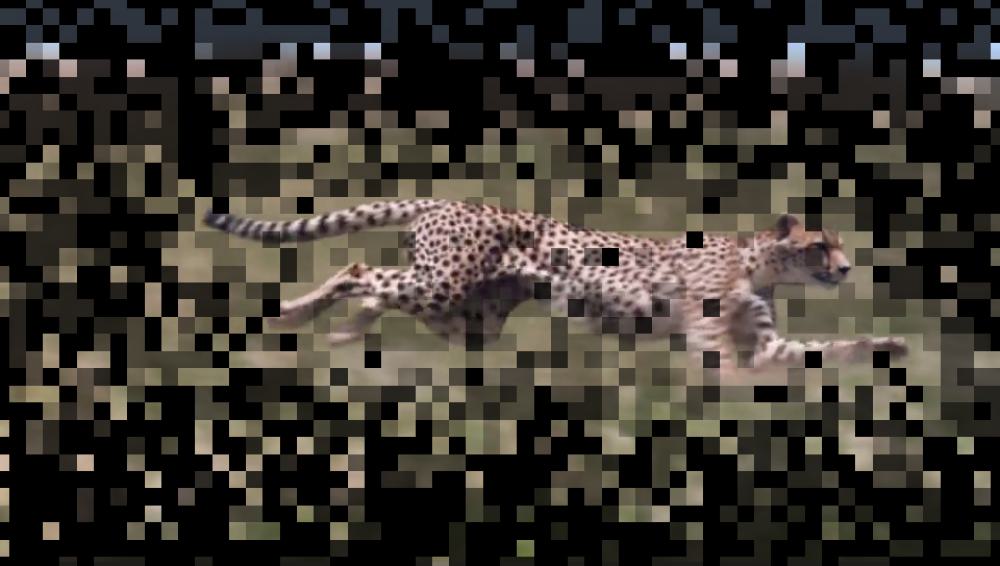} \\
        & \rotatebox{90}{\hspace{20pt}  t=5-8} &
        \includegraphics[width=0.23\linewidth]{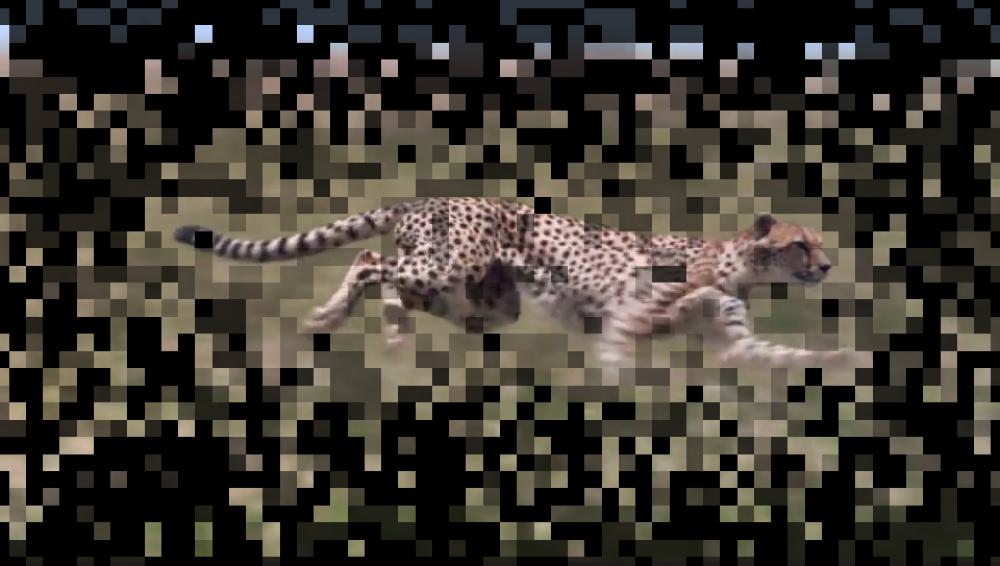} &
        \includegraphics[width=0.23\linewidth]{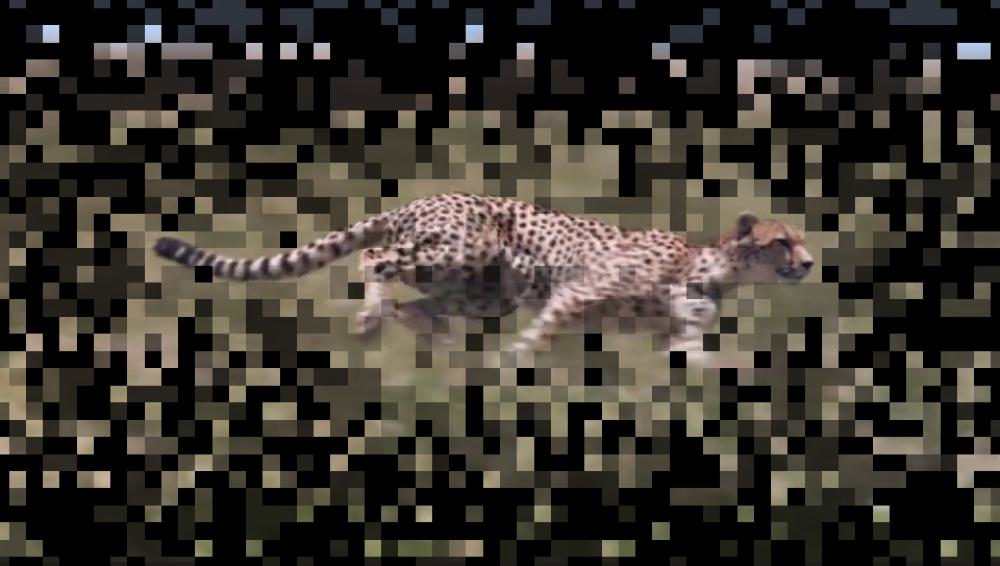} &
        \includegraphics[width=0.23\linewidth]{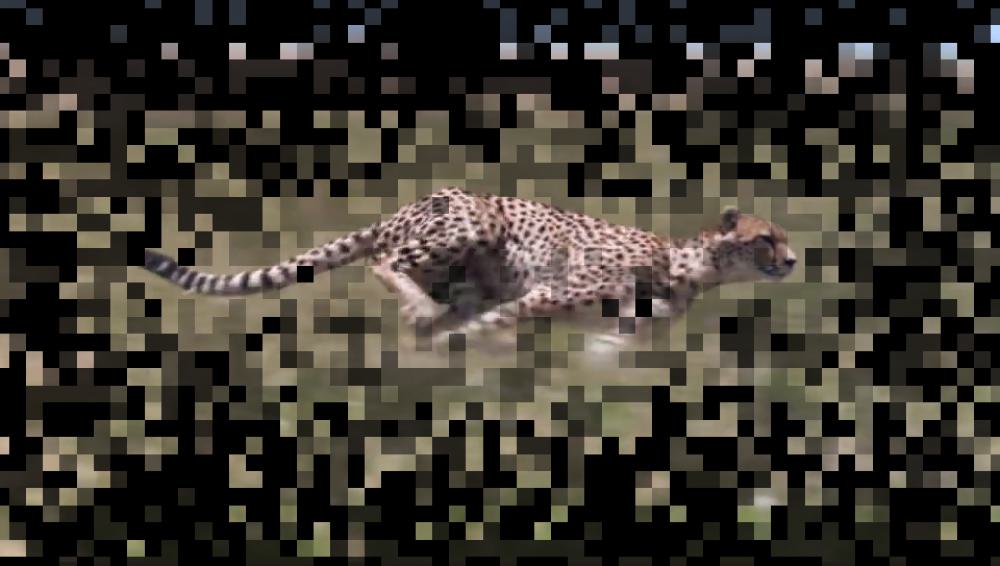} &
        \includegraphics[width=0.23\linewidth]{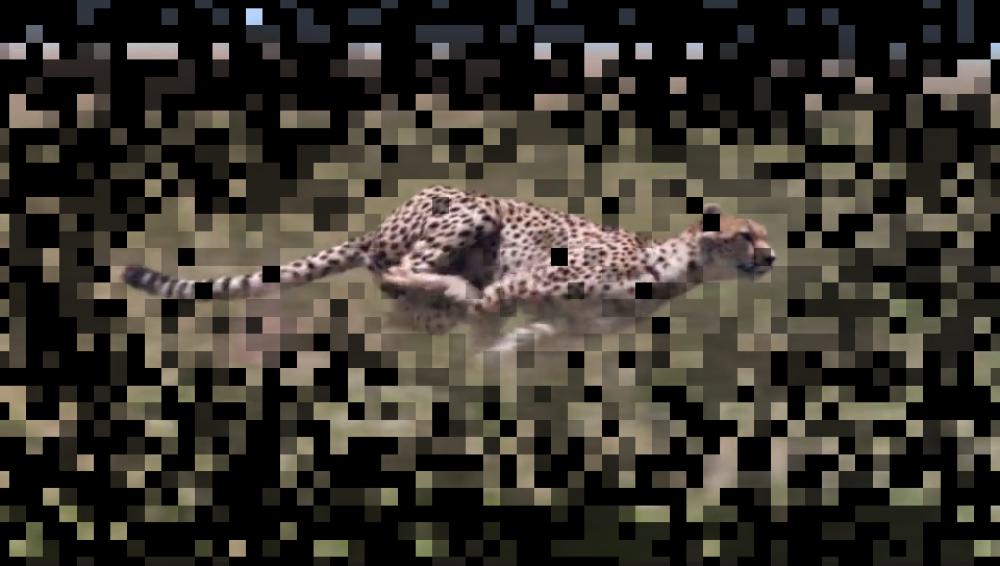} \\

    \end{tabular}
    \caption{Exemplary pruning masks for videos 1–4 at a 40\% Patch Keep Ratio (PKR). 
    Results are shown for timesteps $t\in[1,8]$.
    Patch activity is represented via grayscale shading, where darker patches are activated less.
    Fully black patches are removed by \textit{Map-SM} after the first layer.
    }
    \label{fig:video_examples01}
\end{figure*}

\begin{figure*}
    \setlength{\tabcolsep}{1pt}
    \begin{tabular}{cccccc}
        & & \multicolumn{4}{c}{time $\rightarrow$} \\
        
        \multirow{2}{*}{\rotatebox{90}{Video 5}} & \rotatebox{90}{\hspace{20pt} t=1-4} &
        \includegraphics[width=0.23\linewidth]{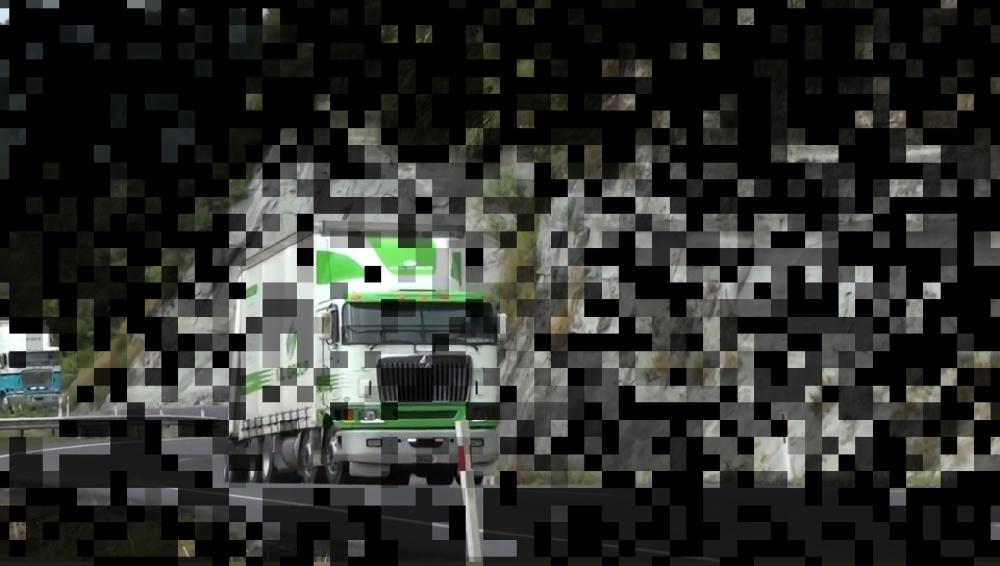} &
        \includegraphics[width=0.23\linewidth]{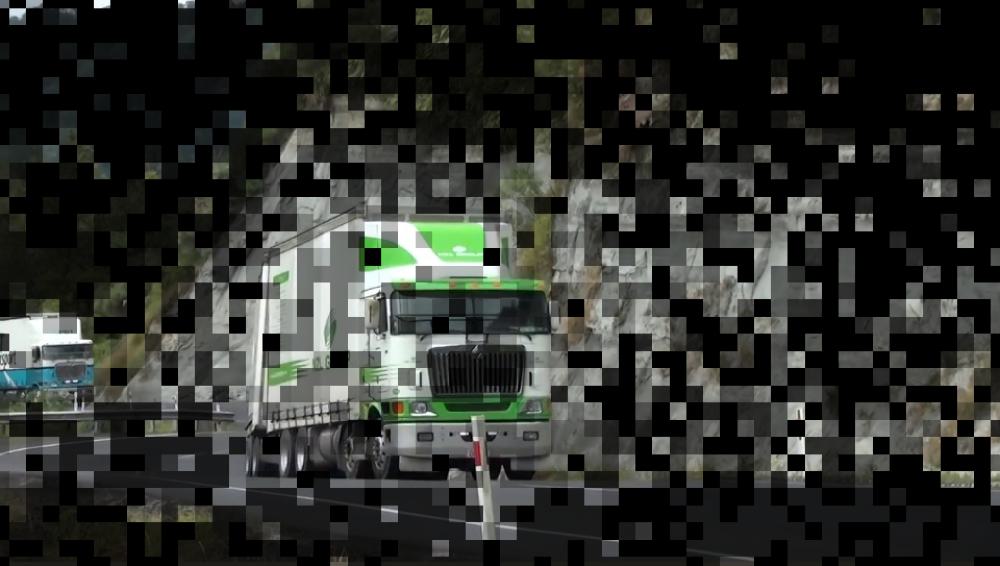} &
        \includegraphics[width=0.23\linewidth]{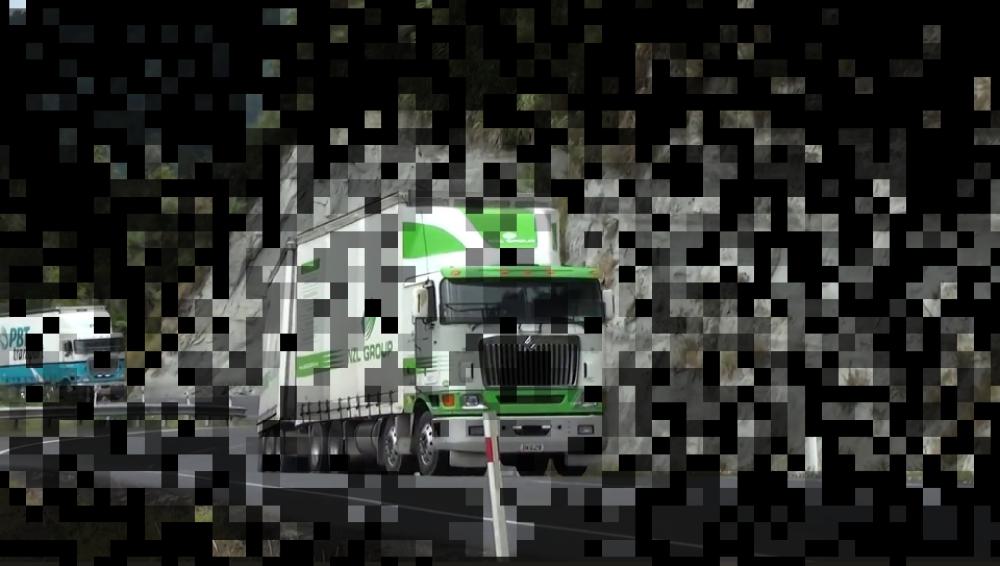} &
        \includegraphics[width=0.23\linewidth]{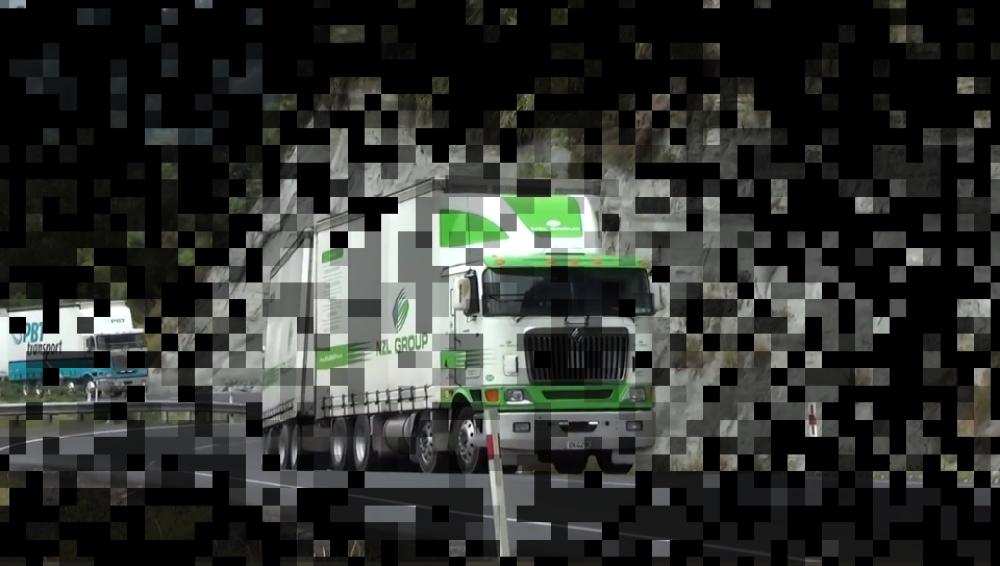} \\
        & \rotatebox{90}{\hspace{20pt}  t=5-8} &
        \includegraphics[width=0.23\linewidth]{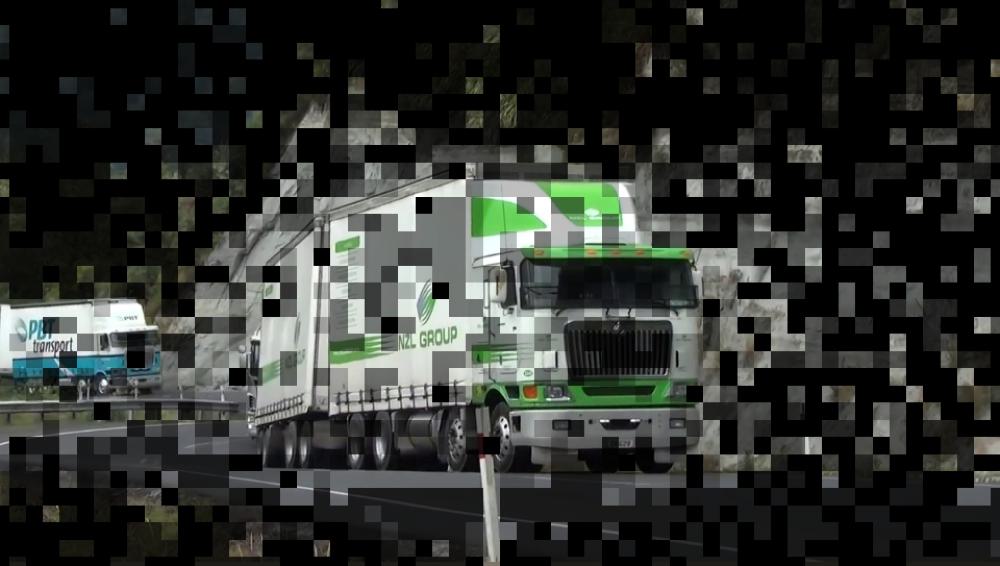} &
        \includegraphics[width=0.23\linewidth]{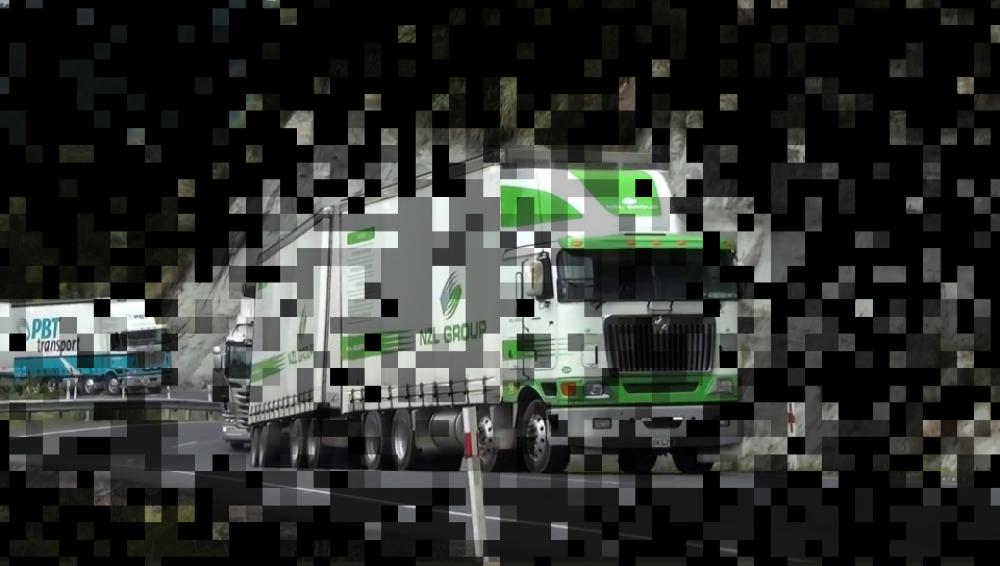} &
        \includegraphics[width=0.23\linewidth]{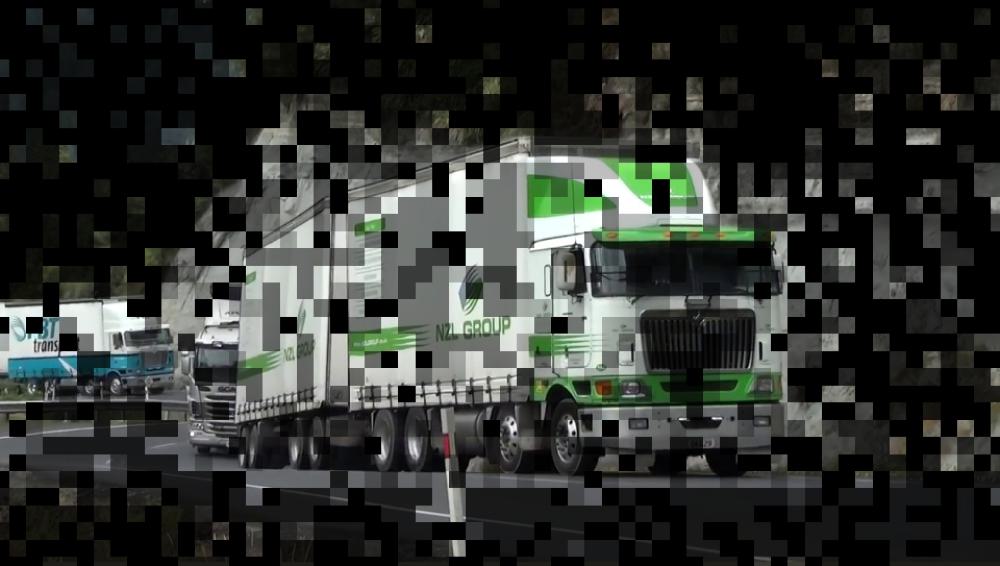} &
        \includegraphics[width=0.23\linewidth]{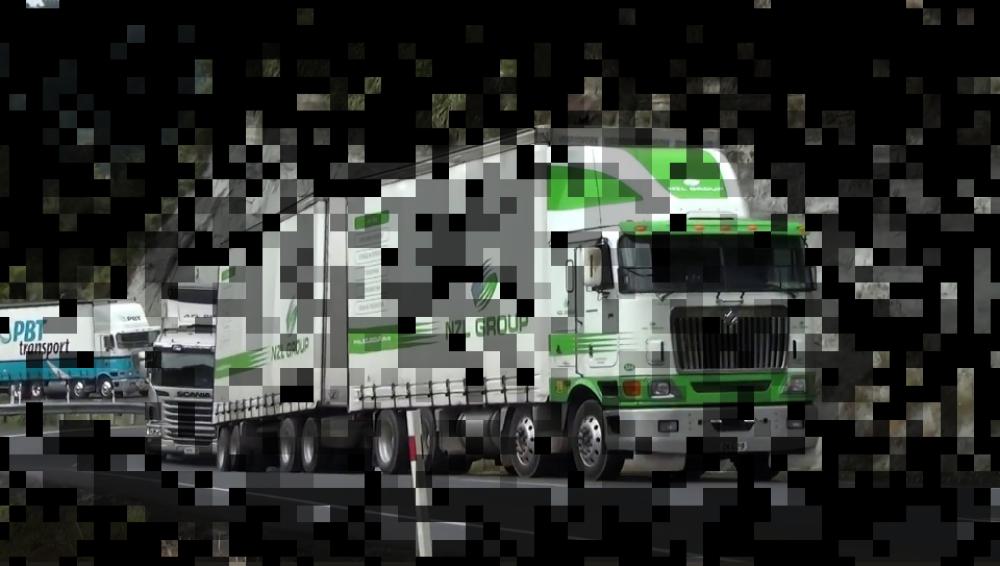} \\
        \midrule 
        
        \multirow{2}{*}{\rotatebox{90}{Video 6}} & \rotatebox{90}{\hspace{20pt} t=1-4} &
        \includegraphics[width=0.23\linewidth]{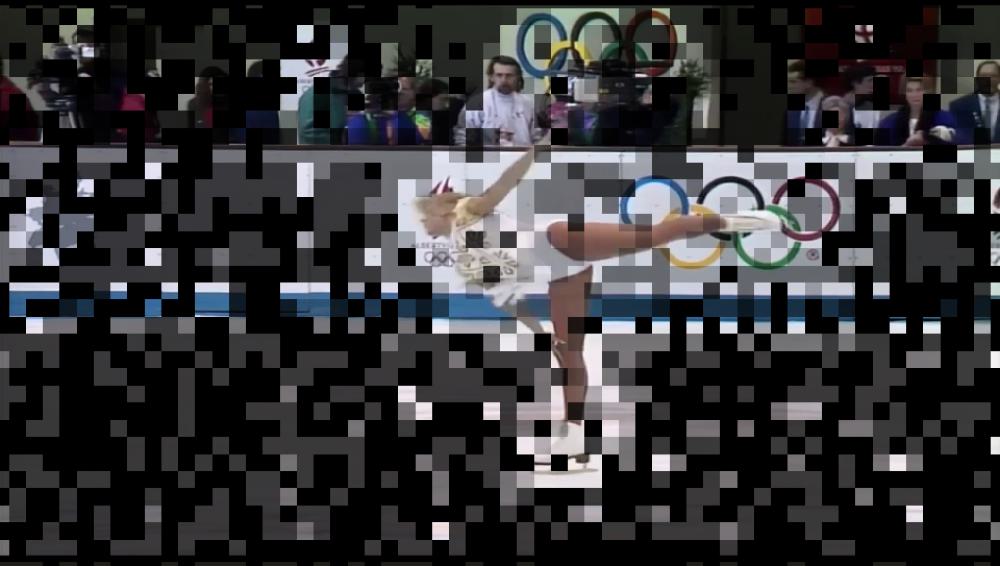} &
        \includegraphics[width=0.23\linewidth]{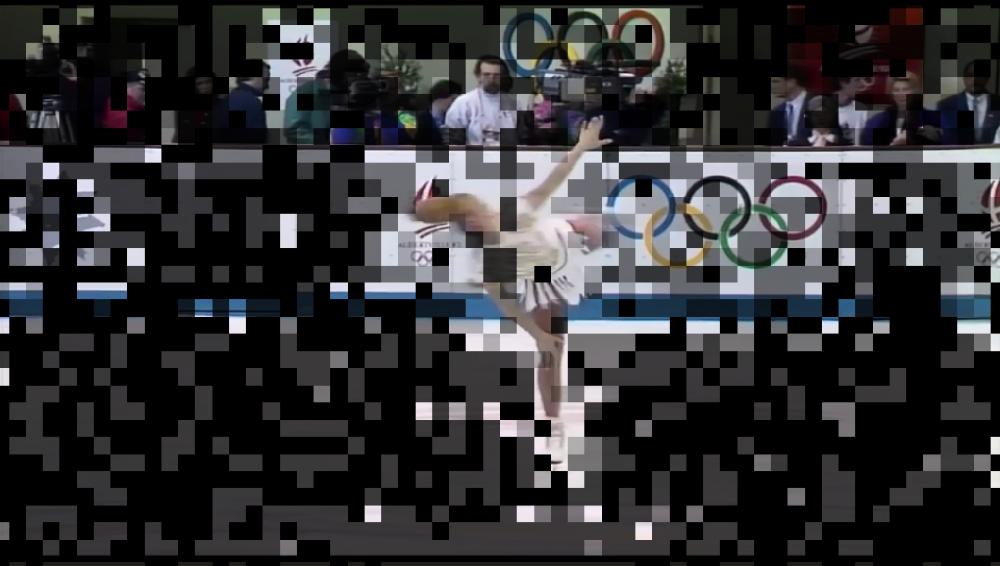} &
        \includegraphics[width=0.23\linewidth]{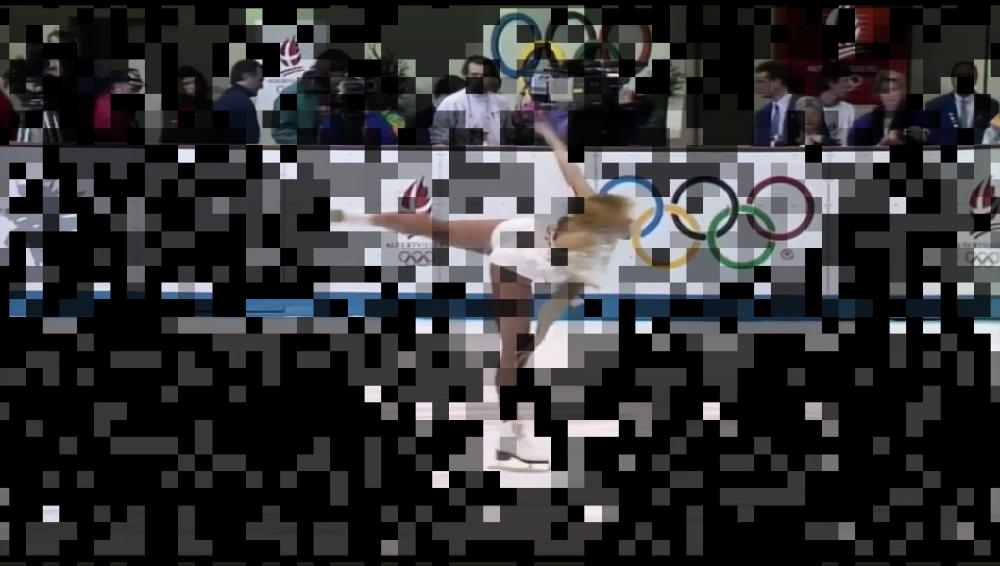} &
        \includegraphics[width=0.23\linewidth]{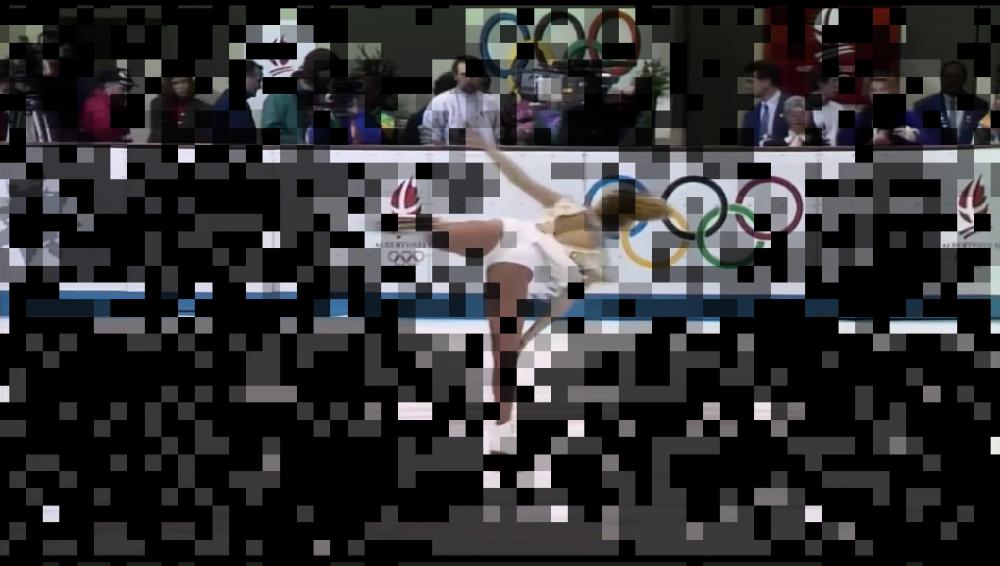} \\
        & \rotatebox{90}{\hspace{20pt}  t=5-8} &
        \includegraphics[width=0.23\linewidth]{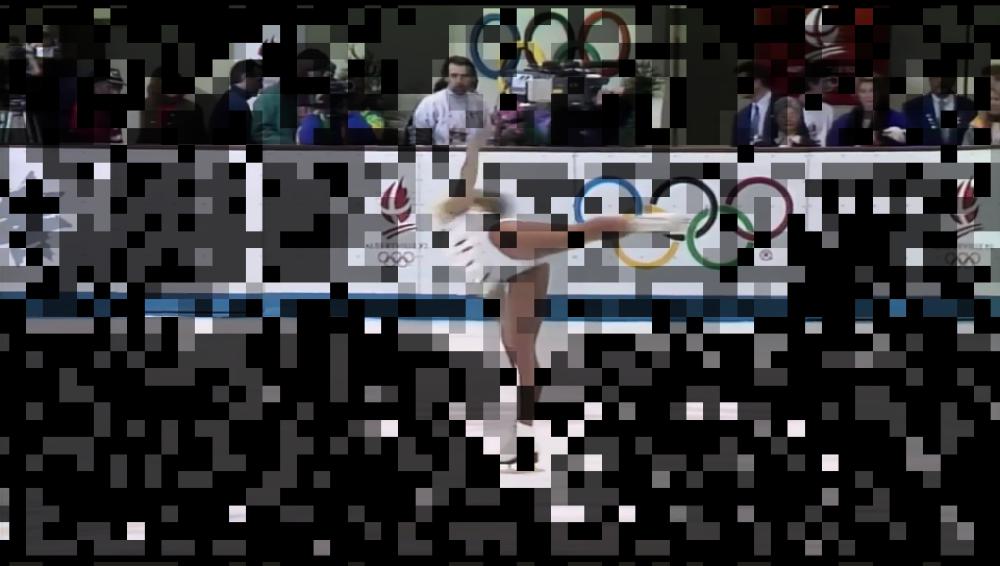} &
        \includegraphics[width=0.23\linewidth]{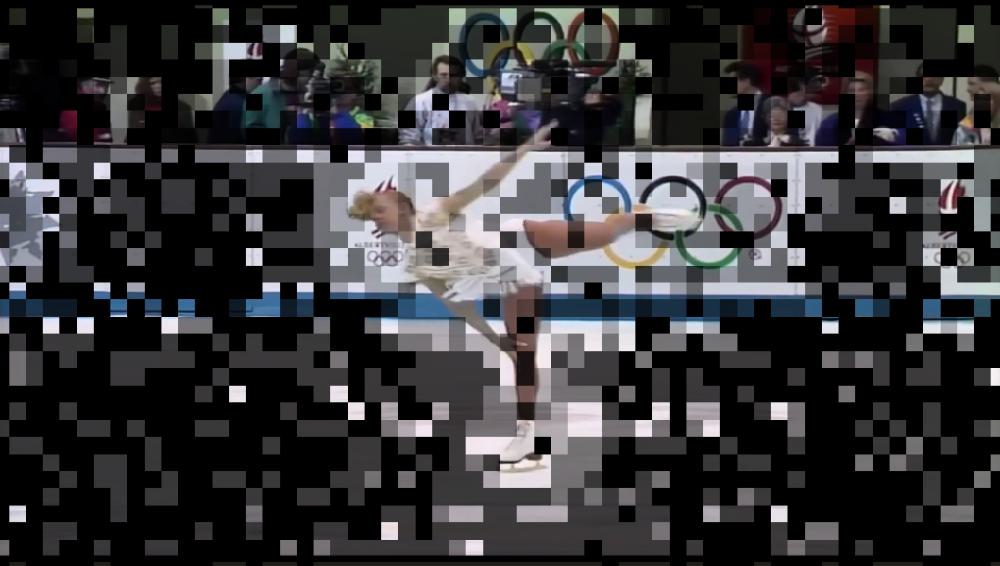} &
        \includegraphics[width=0.23\linewidth]{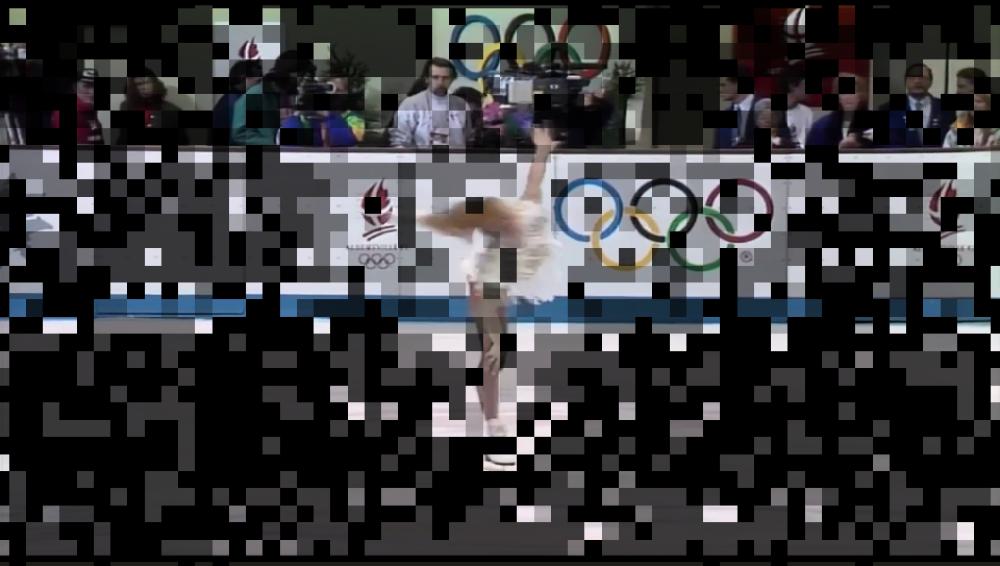} &
        \includegraphics[width=0.23\linewidth]{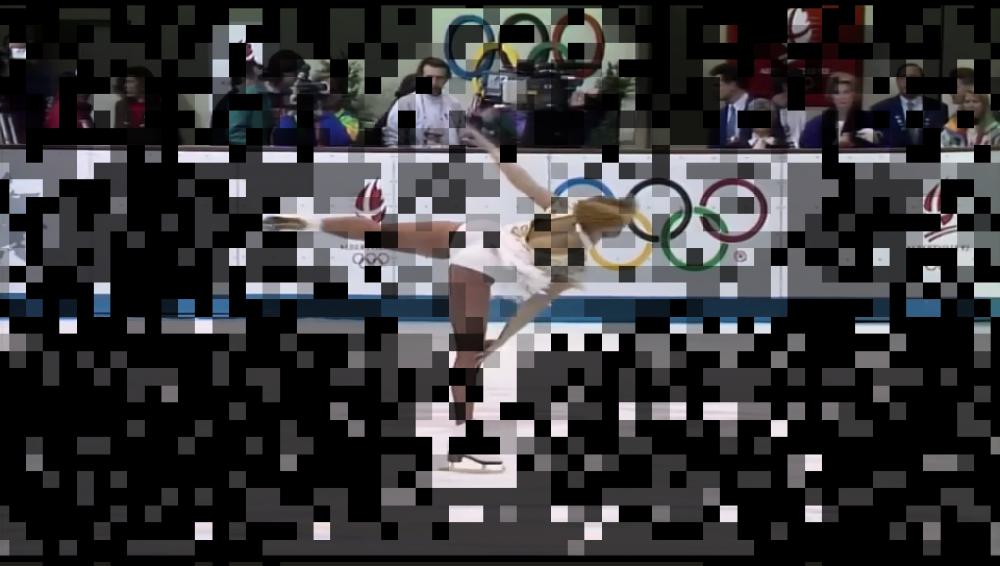} \\
        \midrule 

        \multirow{2}{*}{\rotatebox{90}{Video 7}} & \rotatebox{90}{\hspace{20pt} t=1-4} &
        \includegraphics[width=0.23\linewidth]{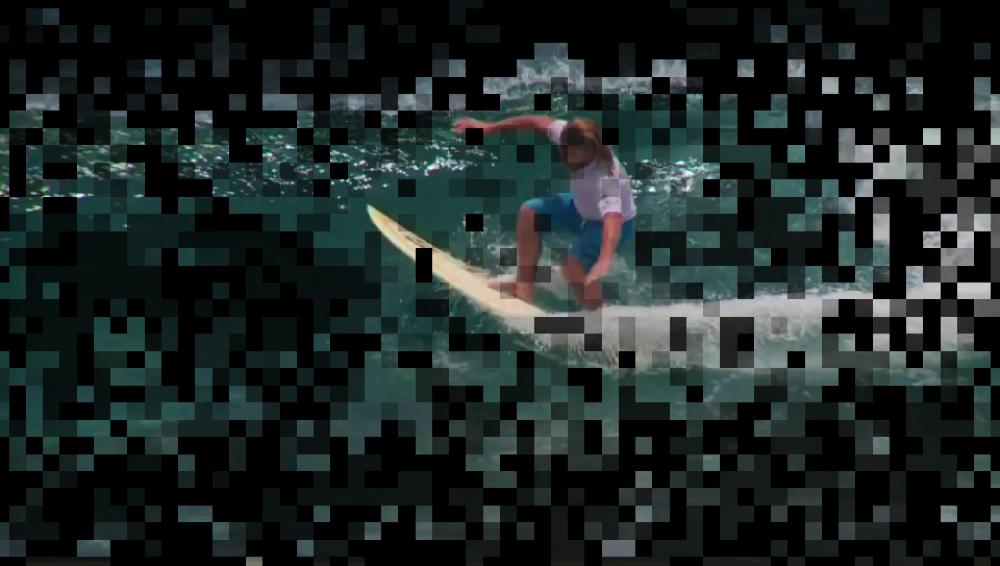} &
        \includegraphics[width=0.23\linewidth]{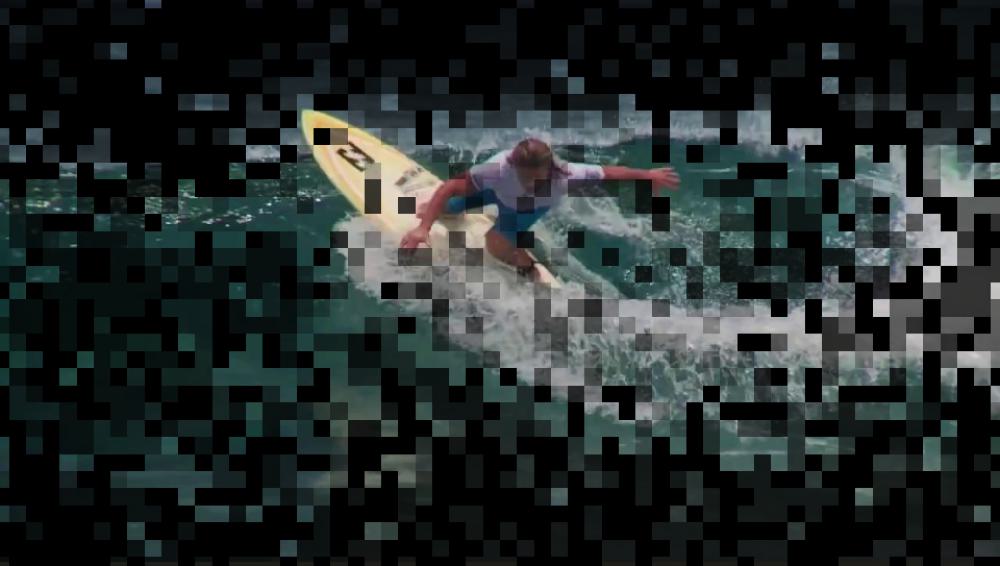} &
        \includegraphics[width=0.23\linewidth]{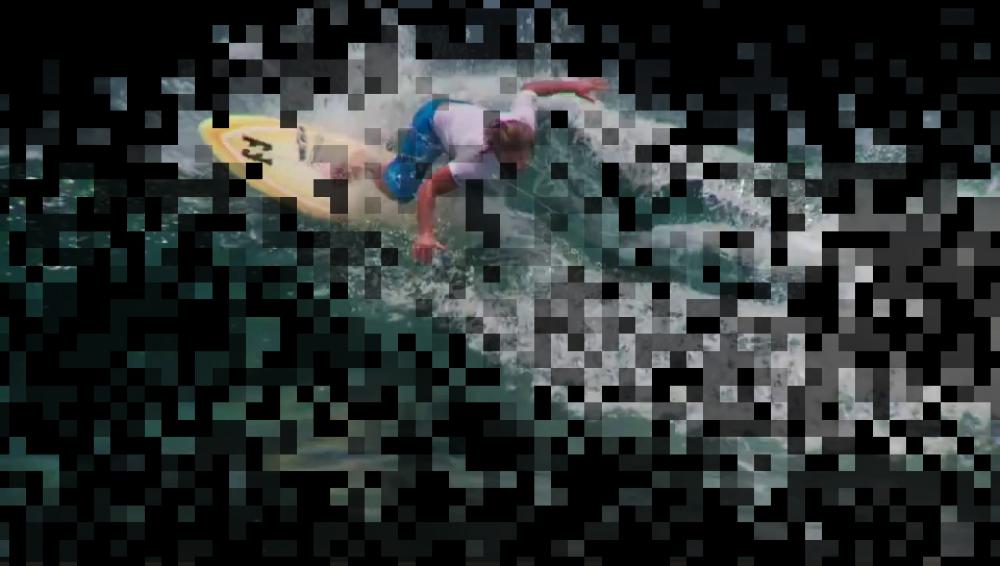} &
        \includegraphics[width=0.23\linewidth]{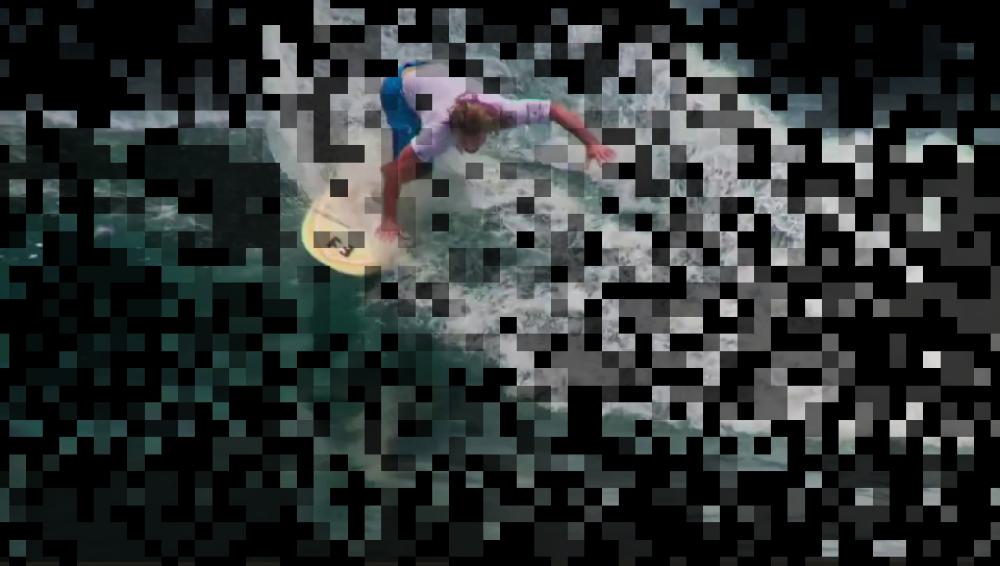} \\
        & \rotatebox{90}{\hspace{20pt}  t=5-8} &
        \includegraphics[width=0.23\linewidth]{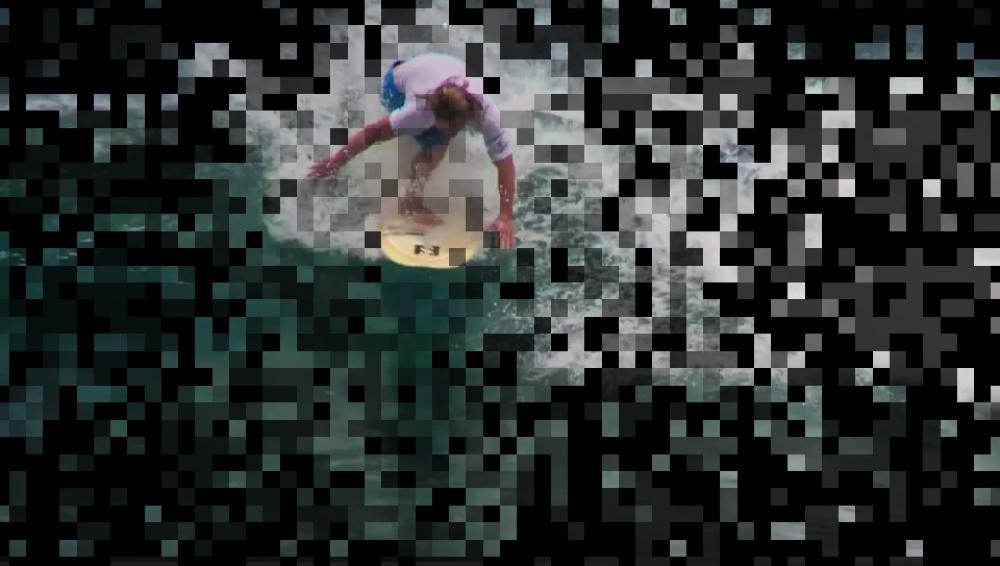} &
        \includegraphics[width=0.23\linewidth]{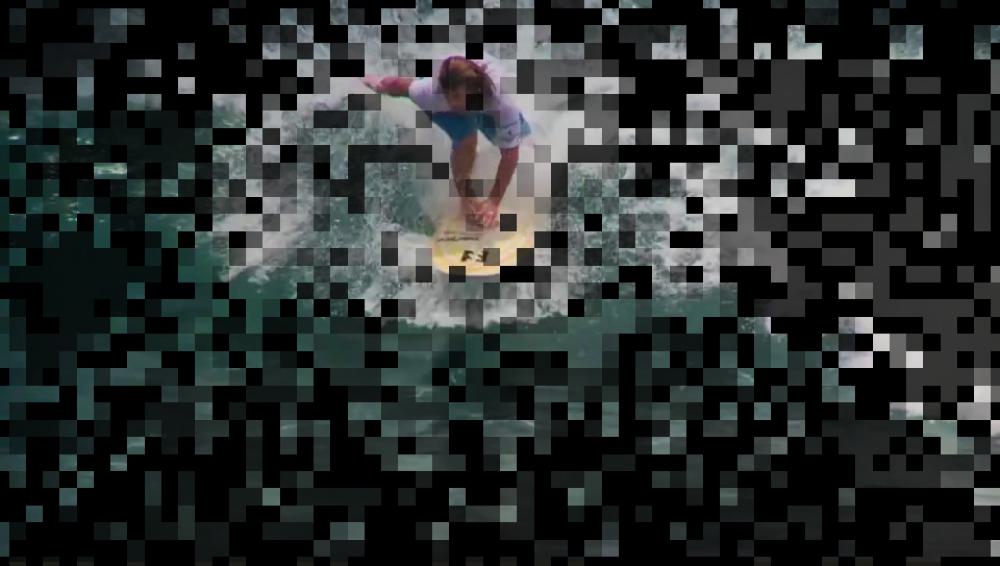} &
        \includegraphics[width=0.23\linewidth]{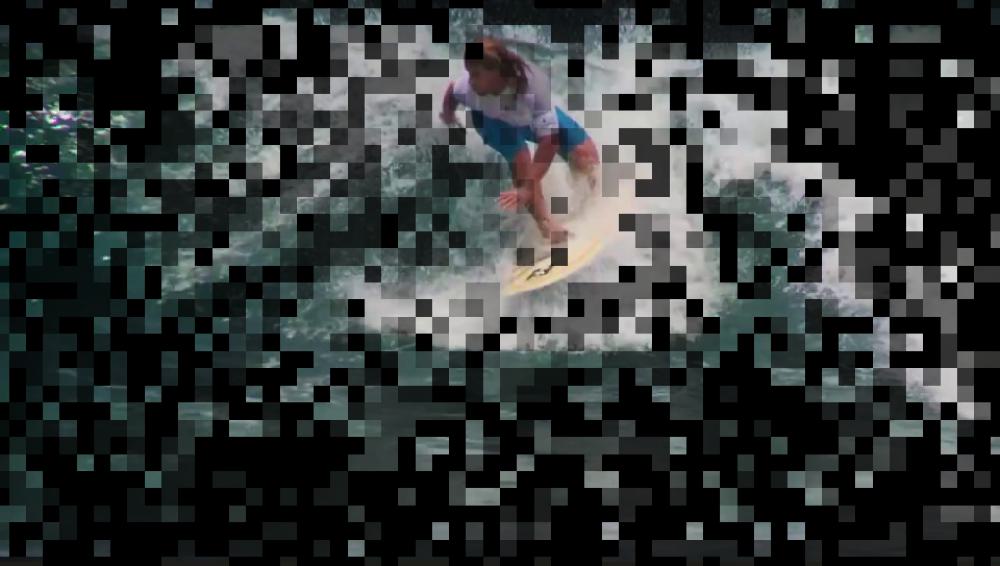} &
        \includegraphics[width=0.23\linewidth]{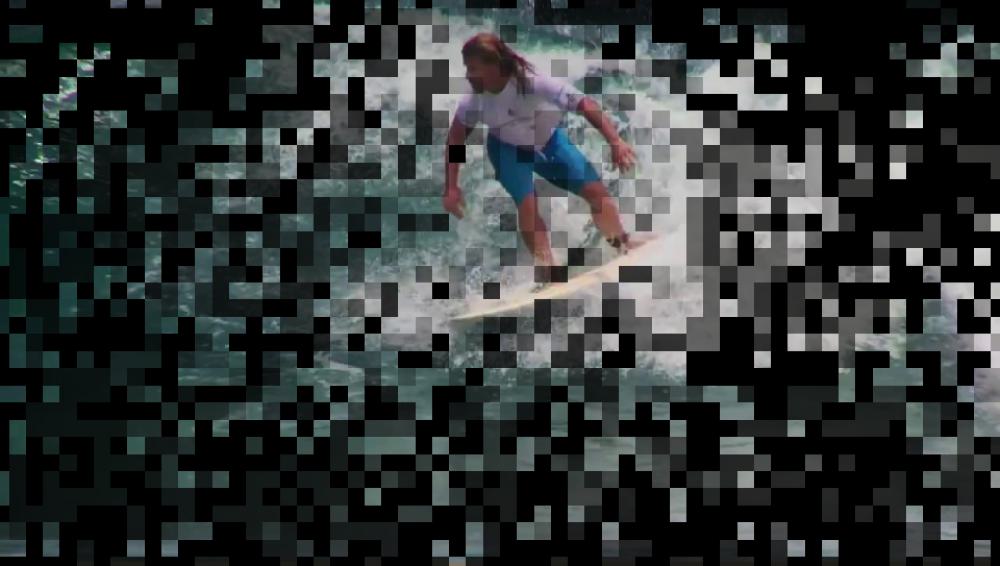} \\
        \midrule

        \multirow{2}{*}{\rotatebox{90}{Video 8}} & \rotatebox{90}{\hspace{20pt} t=1-4} &
        \includegraphics[width=0.23\linewidth]{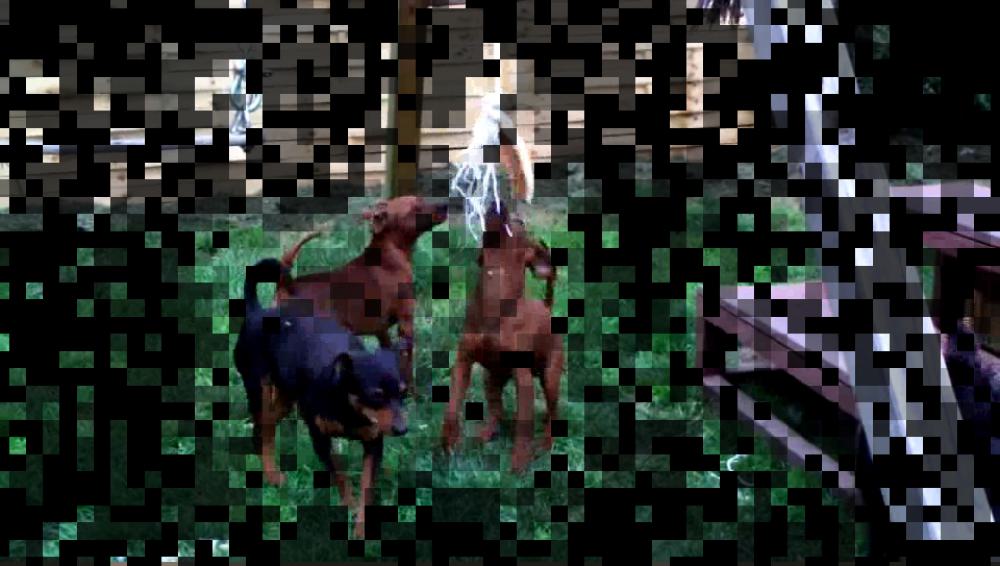} &
        \includegraphics[width=0.23\linewidth]{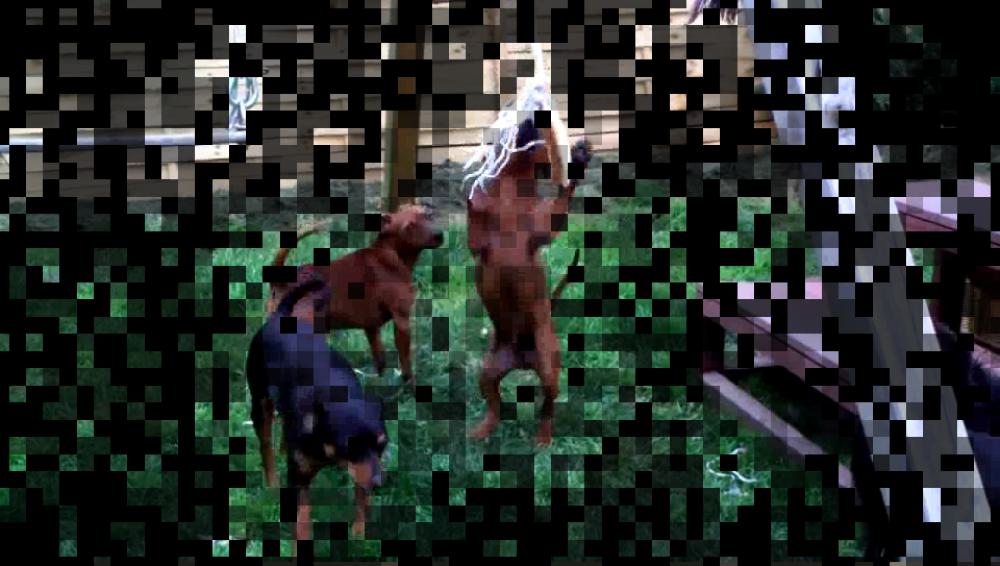} &
        \includegraphics[width=0.23\linewidth]{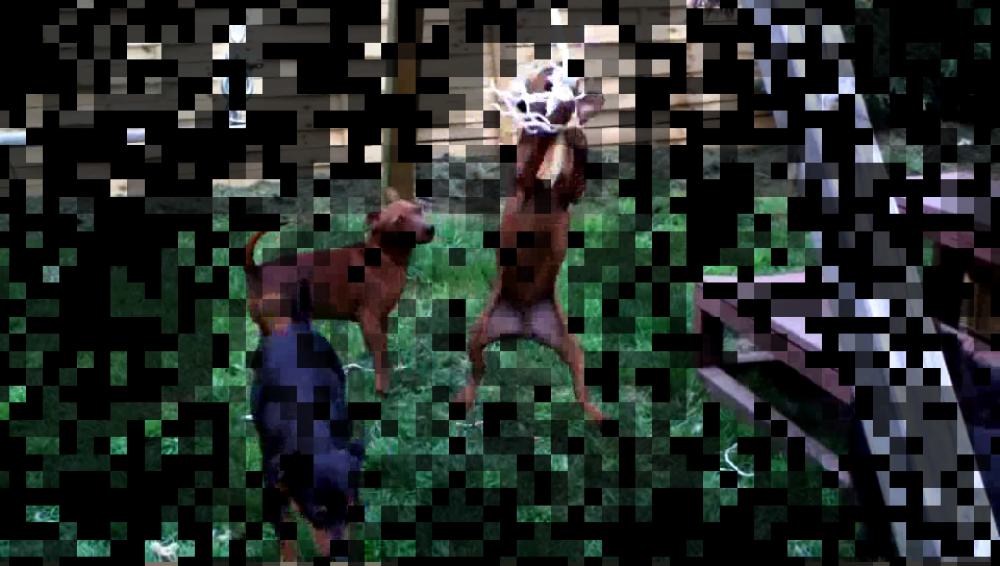} &
        \includegraphics[width=0.23\linewidth]{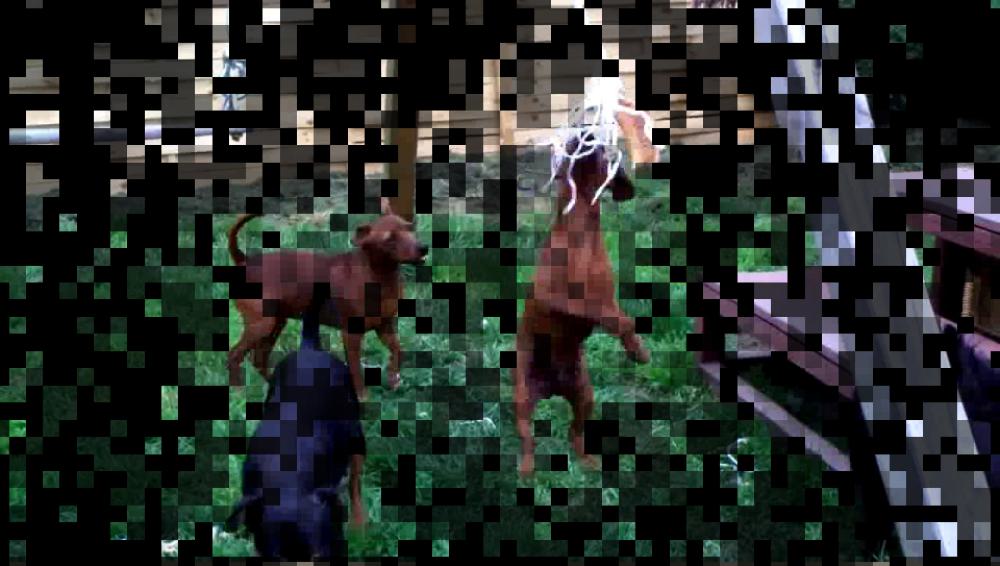} \\
        & \rotatebox{90}{\hspace{20pt}  t=5-8} &
        \includegraphics[width=0.23\linewidth]{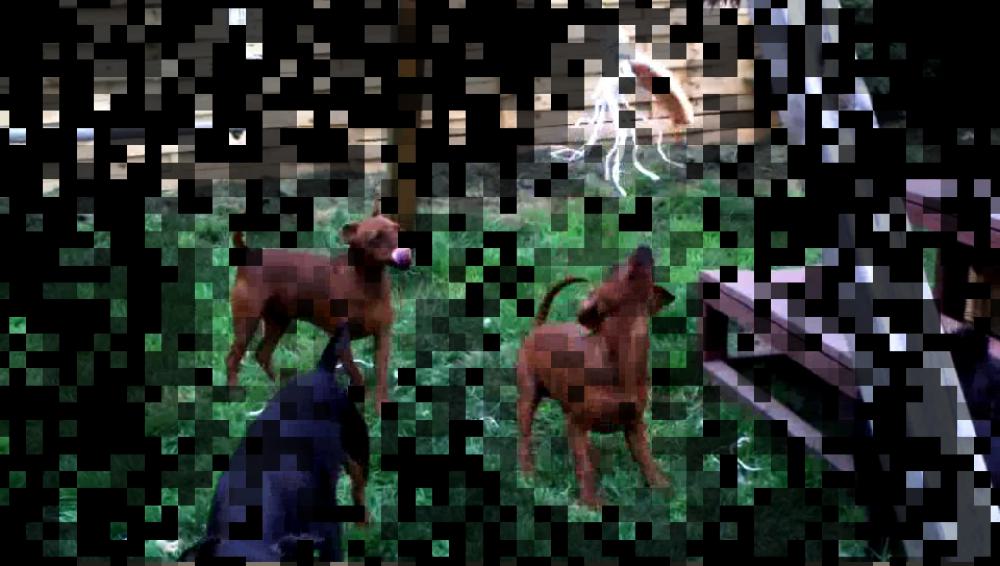} &
        \includegraphics[width=0.23\linewidth]{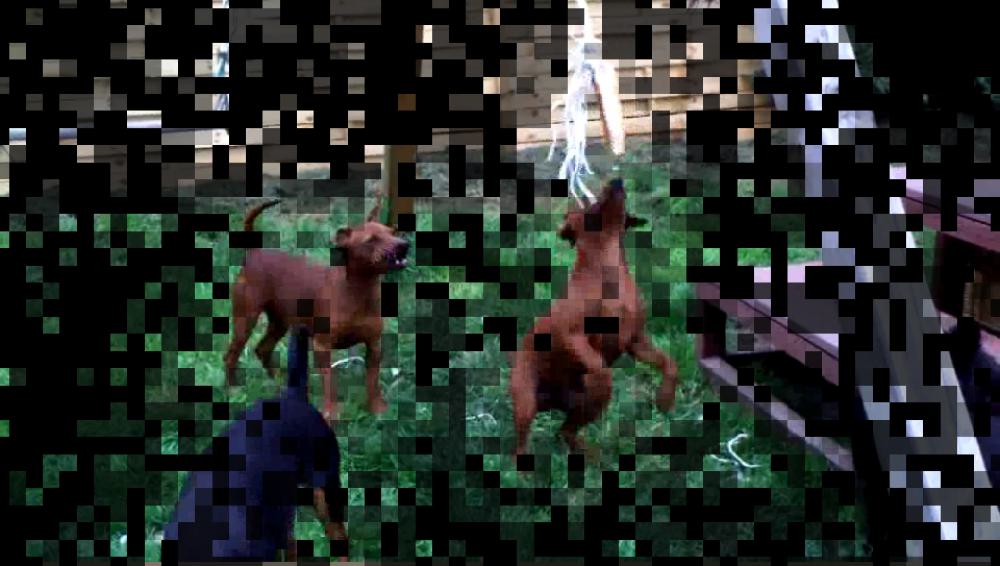} &
        \includegraphics[width=0.23\linewidth]{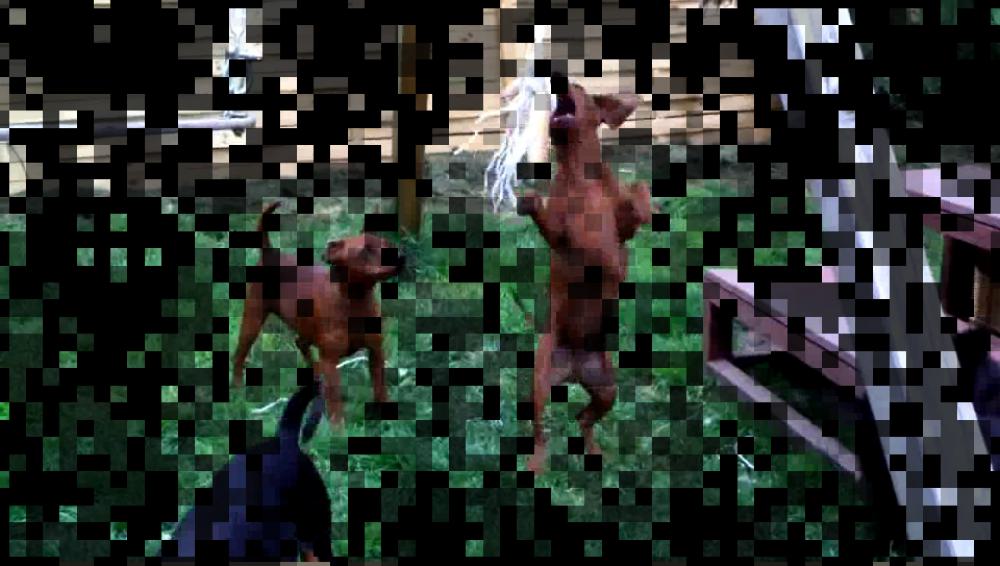} &
        \includegraphics[width=0.23\linewidth]{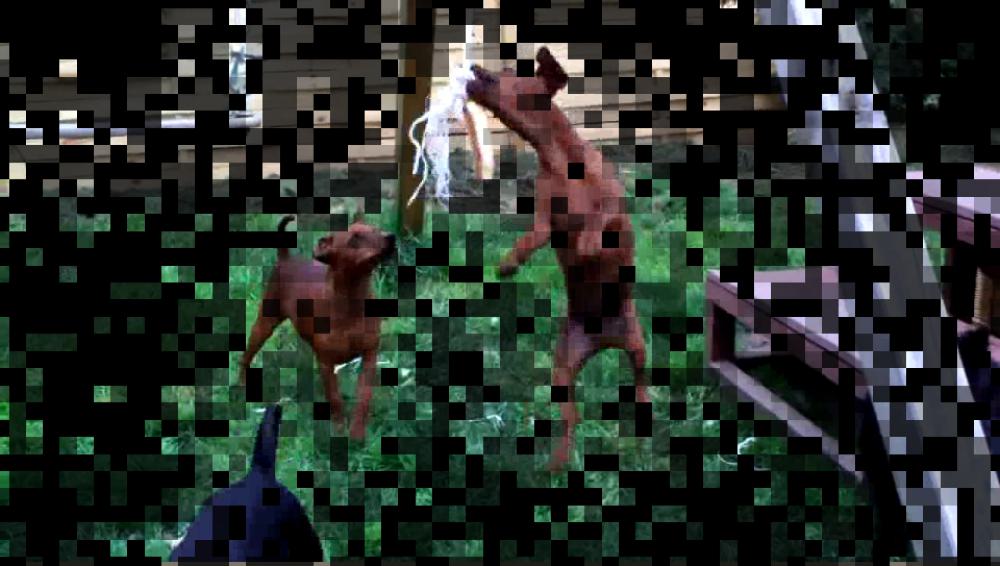} \\

    \end{tabular}
    \caption{Exemplary pruning masks for videos 5-8 at a 40\% Patch Keep Ratio (PKR). 
    Results are shown for timesteps $t\in[1,8]$.
    Patch activity is represented via grayscale shading, where darker patches are activated less.
    Fully black patches are removed by \textit{Map-SM} after the first layer.
    }
    \label{fig:video_examples02}
\end{figure*}

Moreover we show the patch activity of \textit{VPP} in 8 exemplarly video sequences in Fig.~\ref{fig:video_examples01} and~\ref{fig:video_examples02}.
\textit{VPP} highly activates foreground patches while removing background patches in early network layers.
Note that all fully black patches are removed after layer 1 by the introduced \textit{Map-SM}.
The results show that moving objects, such as the cyclist in video 2, exhibit high spatial overlap with the pruning masks across all four stages, including the initial mask.
This demonstrates that \textit{Map-SM} effectively accounts for the temporal displacement of foreground patches throughout the video sequence.


\end{document}